\documentclass{article}
\input{header_neurips} %%
\usepackage[final]{neurips_2020}

\usepackage{appendix}

\usepackage[subfigure]{tocloft}
\advance\cftsecnumwidth 0.5em\relax
\advance\cftsubsecindent 0.5em\relax
\advance\cftsubsecnumwidth 0.5em\relax

\usepackage[compact]{titlesec}  %slightly more compact vertical spacing before/after sections
\titlespacing*{\section}{0pt}{8pt}{0pt}
\titlespacing*{\subsection}{0pt}{4pt}{0pt}

\title{Model Fusion via Optimal Transport}

\author{%
	Sidak Pal Singh\thanks{Work done while at EPFL.} \\
	ETH Zurich, Switzerland\\
	\texttt{contact@sidakpal.com} \\
	\And
	Martin Jaggi \\
	EPFL, Switzerland \\
	\texttt{martin.jaggi@epfl.ch} \\
}

\begin{document}
\maketitle
\addtocontents{toc}{\protect\setcounter{tocdepth}{0}}
% !TEX root = supplementary.tex

\begin{abstract}
	 Combining different models is a widely used paradigm in machine learning applications. While the most common approach is to form an ensemble of models and average their individual predictions, this approach is often rendered infeasible by given resource constraints in terms of memory and computation, which grow linearly with the number of models. We present a layer-wise model fusion algorithm for neural networks that utilizes optimal transport to (soft-) align neurons across the models before averaging their associated parameters. 
	 
	 \hspace{0.5cm} We show that this can successfully yield ``one-shot'' knowledge transfer (i.e, without requiring any retraining) between neural networks trained on heterogeneous non-i.i.d. data. In both i.i.d. and non-i.i.d. settings , we illustrate that our approach significantly outperforms vanilla averaging, as well as how it can serve as an efficient replacement for the ensemble with moderate fine-tuning, for standard convolutional networks (like  \textsc{\textsc{VGG11}}), residual networks (like \textsc{ResNet18}), and multi-layer perceptrons on \textsc{CIFAR10}, \textsc{CIFAR100}, and \textsc{MNIST}. Finally, our approach also provides a principled way to combine the parameters of neural networks with different widths, and we explore its application for model compression. The code is available at the following link, \url{https://github.com/sidak/otfusion}. 
\end{abstract}

\section{Introduction}\label{sec:intro}

If two neural networks had a child, what would be its weights?
In this work, we study the fusion of two \emph{parent} neural networks---which were trained differently but have the same number of layers---into a single \emph{child} network. 
We further focus on performing this operation in a \textit{one-shot manner}, based on the network weights only, so as to minimize the need of any retraining.

This fundamental operation of merging several neural networks into one contrasts other widely used techniques for combining machine learning models:

\emph{Ensemble methods} have a very long history. They combine the outputs of several different models as a way to improve the prediction performance and robustness. However, this requires maintaining the $K$ trained models and running each of them at test time (say, in order to average their outputs). This approach thus quickly becomes infeasible for many applications with limited computational resources, especially in view of the ever-growing size of modern deep learning models.

The simplest way to fuse several parent networks into a single network of the same size is direct \emph{weight averaging}, which we refer to as vanilla averaging; here for simplicity, we assume that all network architectures are identical.
Unfortunately, neural networks are typically highly redundant in their parameterizations, so that there is no one-to-one correspondence between the weights of two different neural networks, even if they would describe the same function of the input.
In practice, vanilla averaging is known to perform very poorly on trained networks whose weights differ non-trivially.

Finally, a third way to combine two models is \emph{distillation}, where one network is retrained on its training data, while jointly using the output predictions of the other `teacher' network on those samples. Such a scenario is considered infeasible in our setting, as we aim for approaches not requiring the sharing of training data. 
This requirement is particularly crucial if the training data is to be kept private, like in federated learning applications, or is unavailable due to e.g. legal reasons.

\textbf{Contributions.}
We propose a novel layer-wise approach of aligning the neurons and weights of several differently trained models, for fusing them into a single model of the same architecture. 
Our method relies on optimal transport (OT) \citep{monge1781memoire, kantorovich1942translocation}, to minimize the transportation cost of neurons present in the layers of individual models, measured by the similarity of activations or incoming weights. 
The resulting layer-wise averaging scheme can be interpreted as computing the Wasserstein barycenter~\citep{agueh, cuturi_doucet14} of the probability measures defined at the corresponding layers of the parent models. 

We empirically demonstrate that our method succeeds in the one-shot merging of networks of different weights, and in all scenarios significantly outperforms vanilla averaging. More surprisingly, we also show that our method succeeds in merging two networks that were trained for slightly different tasks (such as using a different set of labels). The method is able to ``inherit'' abilities unique to one of the parent networks, while outperforming the same parent network on the task associated with the other network. Further, we illustrate how it can serve as a data-free and algorithm independent post-processing tool for structured pruning. Finally, we show that OT fusion, with mild fine-tuning, can act as efficient proxy for the ensemble, whereas vanilla averaging fails for more than two models.

\textbf{Extensions and Applications.}
The method serves as a new building block for enabling several use-cases: (1) The adaptation of a global model to personal training data. (2) Fusing the parameters of a bigger model into a smaller sized model and vice versa. 
 (3) Federated or decentralized learning applications, where training data can not be shared due to privacy reasons or simply due to its large size.  In general, improved model fusion techniques such as ours have strong potential towards encouraging model exchange as opposed to data exchange, to improve privacy $\&$ reduce communication costs.

\section{Related Work}\label{sec:related}

\textbf{Ensembling.} Ensemble methods \citep{Breiman1996, Wolpert, Schapire} have long been in use in deep learning and machine learning in general. However, given our goal is to obtain a single model, it is assumed infeasible to maintain and run several trained models as needed here.

\textbf{Distillation.}
Another line of work by \citet{hinton2015distilling, caruana,schmidhuber1992learning} proposes distillation techniques.  Here the key idea is to employ the knowledge of a pre-trained teacher network (typically larger and expensive to train) and transfer its abilities to a smaller model called the student network. During this transfer process, the goal is to use the relative probabilities of misclassification of the teacher as a more informative training signal.

While distillation also results in a single model, the main drawback is its computational complexity---the distillation process is essentially as expensive as training the student network from scratch, and also involves its own set of hyper-parameter tuning. In addition, distillation still requires sharing the training data with the teacher (as the teacher network can be too large to share), which we avoid here.

In a different line of work, \citet{shen2018meal} propose an approach where the student network is forced to produce outputs mimicking the teacher networks, by utilizing Generative Adversarial Network~\citep{goodfellow2014generative}. This still does not resolve the problem of high computational costs involved in this kind of knowledge transfer. Further, it does not provide a principled way to aggregate the parameters of different models.

\textbf{Relation to other network fusion methods.}
Several studies have investigated a method to merge two trained networks into a single network without the need for retraining \citep{smith2017periodic,utans1996weight,leontev2018non}.
\citet{leontev2018non}~propose Elastic Weight Consolidation, which formulates an assignment problem on top of  diagonal approximations to the Hessian matrices of each of the two parent neural networks. Their  method however only works when the weights of the parent models are already close, i.e. share a significant part of the training history \citep{smith2017periodic,utans1996weight}, by relying on SGD with periodic averaging, also called local SGD \citep{stich2019local}. Nevertheless, their empirical results~\citep{leontev2018non} do not improve over vanilla averaging.

\textbf{Alignment-based methods}. Alignment of neurons was considered in \citet{li2016convergent} to probe the representations learned by different networks. Recently, \citet{yurochkin2019bayesian} independently proposed a Bayesian non-parametric framework that considers matching the neurons of different MLPs in federated learning. In a concurrent work\footnote{An early version of our paper also appeared at NeurIPS 2019 workshop on OT, \href{https://arxiv.org/pdf/1910.05653.pdf}{arxiv:1910.05653}.}, \citet{Wang2020Federated} extend \cite{yurochkin2019bayesian}  to more realistic networks including CNNs, also with a specific focus on federated learning. In contrast, we develop our method from the lens of optimal transport (OT), which lends us a simpler approach by utilizing Wasserstein barycenters.  
The method of aligning neurons employed in both lines of work form  instances for the choice of ground metric in OT. Overall, we consider model fusion in general, beyond federated learning. For instance, we show applications of fusing different sized models (e.g., for structured pruning) as well as the compatibility of our method to serve as an initialization for distillation. From a practical side, our approach is $\#$ of layer times more efficient and also applies to ResNets.

To conclude, the application of Wasserstein barycenters for averaging the weights of neural networks has---to our knowledge---not been considered in the past. 

\vspace{-1mm}

\section{Background on Optimal Transport (OT)}\label{sec:backgroud_OT}
We present a short background on OT in the discrete case, and in this process set up the notation for the rest of the paper. 
OT gives a way to compare two probability distributions defined over a ground space~$\GS$, provided an underlying distance or more generally the cost of transporting one point to another in the ground space. 
Next, we describe the linear program (LP) which lies at the heart of OT.

\textbf{LP Formulation.} First, let us consider two empirical probability measures $\mu$ and $\nu$ denoted by a weighted sum of Diracs, i.e.,  $\mu=\sum_{i=1}^{n}  \vecind{\alpha}_i \; \delta(\vec{x}^{(i)})$ and $\nu=\sum_{i=1}^{m}  \vecind{\beta}_i \; \delta(\vec{y}^{(i)})$. Here $\delta(\vec{x})$ denotes the Dirac (unit mass) distribution at point $\vec{x}\in \GS$ and the set of points $\mX=(\vec{x}^{(1)},\dots,\vec{x}^{(n)})\in \GS^{n}$. The weight $\vec{\alpha} = (\alpha_1,\dots,\alpha_{n})$ lives in the probability simplex $\Sigma_{n} := \left\{\vec{a}\in\mathbb{R}_{_+}^{n}\;|\;\sum_{_{i=1}}^{n} a_i=1\right \}$ 
(and similarly $\vec{\beta}$).
Further, let $\mC_{ij}$ denote the ground cost of moving point $\vec{x}^{(i)}$ to $\vec{y}^{(j)}$. Then the optimal transport  between $\mu$ and~$\nu$ can be formulated as solving the following linear program,  

\begin{equation}\label{eq:wasserstein}
\text{OT}(\mu,\nu ; \mC) := \hspace{-1.5em} \displaystyle \min_{\substack{\mT \in \mathbb{R}_+^{(n \times m)} \text{s.t.}\  \ \mT \1_{m}=\,\vec{\alpha}, \; \mT^{\top} \1_{n}=\,\vec{\beta}  }} \langle \mT, \mC\rangle
\end{equation}

Here, \(\langle \mT, \mC\rangle := \operatorname{tr}\left(\mT^{\top} \mC\right)= \sum_{ij} T_{ij} C_{ij}\) is the Frobenius inner product of matrices.
The optimal $\mT\in\mathbb{R}_+^{(n \times m)}$ is called as the \emph{transportation matrix} or \emph{transport map}, and $T_{ij}$ represents the optimal amount of mass to be moved from point $\vec{x}^{(i)}$ to $\vec{y}^{(j)}$.   

\textbf{Wasserstein Distance.} In the case where $\GS = \mathbb{R}^d$ and the cost is defined with respect to a metric 
$ D_\GS $ over $ \GS $  \big(i.e.,~$C_{ij} = D_\GS(\vec{x}^{(i)}, \vec{y}^{(j)})^p$ for any $i,j\big)$,  OT establishes a distance between  probability distributions. This is called the $p$-Wasserstein distance and is defined as $\mathcal{W}_p(\mu,\nu) := \text{OT}(\mu, \nu; D_\GS^p)^{1/p}$. 

\textbf{Wasserstein Barycenters.} This represents the notion of averaging in the Wasserstein space. To be precise, the Wasserstein barycenter \cite{agueh}
is a probability measure that minimizes the weighted sum of ($p$-th power) Wasserstein distances to the given $K$ measures $\{\mu_{1},\dots,\mu_{K}\}$, with corresponding weights $\vec{\eta} = \{\eta_1, \dots, \eta_K\} \in \Sigma_K$. Hence,   it can be written as,  
$\mathcal{B}_p(\mu_{1},\dots,\mu_{K}) = \argmin_\mu \sum_{k=1}^K \eta_k \; \mathcal{W}_p(\mu_{k}, \nu)^p$. 

\vspace{-1mm}
\section{Proposed Algorithm}\label{sec:methodo}
\vspace{-1mm}

In this section, we discuss our proposed algorithm for model aggregation. First, we consider that we are averaging the parameters of only two neural networks, but later present the extension to the multiple model case. For now, we ignore the bias parameters and we only focus on the weights. This is to make the presentation succinct, and it can be easily extended to take care of these aspects.

\paragraph{Motivation.} As alluded to earlier in the introduction, the problem with vanilla averaging of parameters is the lack of one-to-one correspondence between the model parameters. In particular, for a given layer,  there is no direct matching between the neurons of the two models. For e.g., this means that the $p^{\text{th}}$ neuron of  model A might behave very differently (in terms of the feature it detects) from the $p^{\text{th}}$ neuron of the other model B, and instead might be quite similar in functionality to the ${p+1}^{\text{th}}$ neuron. 
Imagine, if we knew a perfect matching between the neurons, then we could simply align the neurons of model A with respect to B. Having done this, it would then make more sense to perform vanilla averaging of the neuron parameters. The matching or assignment could be formulated as a permutation matrix, and just multiplying the parameters by this matrix would align the parameters. 

\begin{figure*}[t!]
	\centering    
	
	\includegraphics[width=0.7\textwidth]{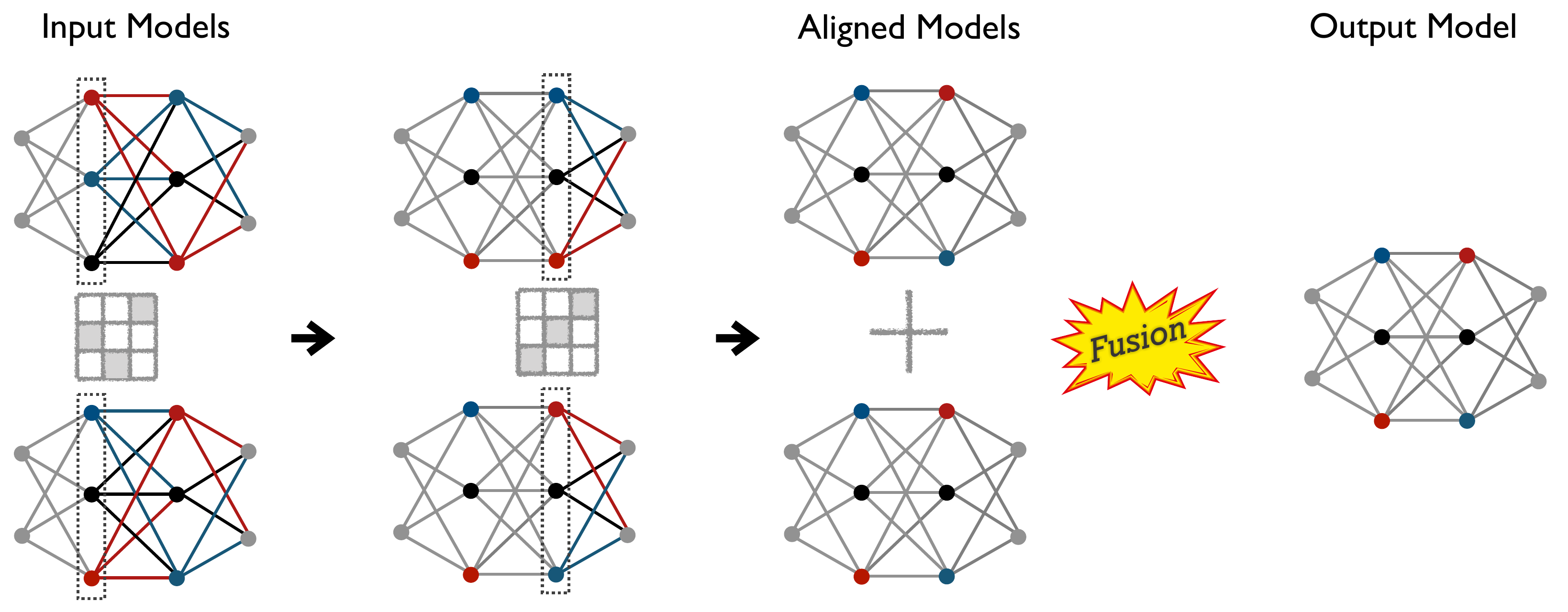}

	\caption{\textbf{Model Fusion procedure}: 
		The first two steps illustrate how the model A (top) gets aligned with respect to model B (bottom). The alignment here is reflected by the ordering of the node colors in a layer. Once each layer has been aligned, the model parameters get averaged (shown by the $+$) to yield a fused model at the end.\vspace{-2mm}}
	
	\label{fig:diagram}
\end{figure*}

But in practice, it is more likely to have soft correspondences between the neurons of the two models for a given layer, especially if their number is not the same across the two models. This is where optimal transport comes in and provides us a soft-alignment matrix in the form of the transport map~$\mT$. In other words, the alignment problem can be rephrased as optimally transporting the neurons in a given layer of model A to the neurons in the same layer of model B. 

\paragraph{General procedure.} 
Let us assume we are at some layer $\ell$ and that neurons in the previous layers have already been aligned. Then, we define probability measures over neurons in this layer for the two models as, $\mu^{(\ell)}=\big(\vec{\alpha}^{(\ell)}, \mX[\ell]\big)$ and $\nu^{(\ell)}=\big(\vec{\beta}^{(\ell)}, \mY[{\ell}]\big)$, where $\mX, \mY$ are the measure supports. 

Next, we use uniform distributions to initialize the histogram (or probability mass values) for each layer.  Although we note that it is possible to additionally use other measures of neuron importance~\citep{dhamdhere2018how, sundararajan2017axiomatic}, but we leave it for a future work. In particular, if the size of layer $\ell$ of models A and B is denoted by $n^{(\ell)}$, $m^{(\ell)}$ respectively, we get $\vec{\alpha}^{(\ell)} \leftarrow \1_{n^{(\ell)}}/n^{(\ell)}, \;  \vec{\beta}^{(\ell)} \leftarrow  \1_{m^{(\ell)}}/m^{(\ell)}$.

Now, in terms of the alignment procedure, we first align the incoming edge weights for the current layer~$\ell$. This can be done by post-multiplying with the previous layer transport matrix $\mT^{(\ell-1)}$, normalized appropriately via the inverse of the corresponding column marginals~$\vec{\beta}^{(\ell-1)}$: 

\begin{equation}\label{eq:ot-upd}
\widehat{\mW}^{(\ell, \, \ell-1)}_A   \leftarrow \mW^{(\ell, \,\ell-1)}_A \mT^{(\ell-1)} \text{diag}\big(1/{\vec{\beta}^{(\ell-1)}}\big).
\end{equation}

This update can be interpreted as follows: the matrix \(\mT^{(\ell-1)} \text{diag}\left(\vec{\beta}^{-(\ell-1)}\right)\) has \(m^{(\ell-1)}\) columns in the simplex
\(\Sigma_{n^{(\ell-1)}} \), thus post-multiplying $\mW^{(\ell, \,\ell-1)}_A$ with it will produce a convex combination of the points in~\(\mW^{(\ell, \,\ell-1)}_A\) with weights defined
by the optimal transport map~\(\mT^{(\ell-1)}\). \vspace{1mm}

Once this has been done, we focus on aligning the neurons in this layer $\ell$ of the two models. Let us assume, we have a suitable ground metric $D_\GS$ (which we discuss in the sections ahead). Then we compute the optimal transport map $\mT^{(\ell)}$ between the measures $\mu^{(\ell)}, \nu^{(\ell)}$ for layer $\ell$, i.e., $\mT^{(\ell)}, \; \mathcal{W}_{2} \ \leftarrow \text{OT}(\mu^{(\ell)}, \nu^{(\ell)}, D_\GS)$,
where $\mathcal{W}_{2}$ denotes the obtained Wasserstein-distance. Now, we use this transport map $\mT^{(\ell)}$ to align the neurons (more precisely the weights) of the first model (A) with respect to the second (B), 

\begin{equation}\label{eq:align}
\widetilde{\mW}^{(\ell, \,\ell-1)}_A   \leftarrow \text{diag}\bigg(\frac{1}{{\vec{\beta}^{(\ell)}}}\bigg) {\mT^{(\ell)}}^{\top} \widehat{\mW}^{(\ell,\, \ell-1)}_A.
\end{equation}

We will refer to model A's weights, $\widetilde{\mW}^{(\ell, \,\ell-1)}_A$, as those aligned with respect to model B. Hence, with this alignment in place, we can average the weights of two layers to obtain the fused weight matrix $\mW^{(\ell, \,\ell-1)}_\mathcal{F}$, as in Eq.~\eqref{eq:avg}. We carry out this procedure over all the layers sequentially.% 
\begin{equation}\label{eq:avg}\textstyle
\mW^{(\ell, \,\ell-1)}_\mathcal{F}\leftarrow \frac{1}{2} \big(\widetilde{\mW}^{(\ell,\, \ell-1)}_A + \mW^{(\ell, \,\ell-1)}_B\big).
\end{equation}
Note that, since the input layer is ordered identically for both  models, we start the alignment from second layer onwards. Additionally, the order of neurons for the very last layer, i.e., in the output layer, again is identical. Thus, the (scaled) transport map at the last layer will be equal to the identity.

\paragraph{Extension to multiple models.}
The key idea is to begin with an estimate $\widehat{M}_\mathcal{F}$ of the fused model, then align all the given models with respect to it, and finally return the average of these aligned weights as the final weights for the fused model. For the two model case, this is equivalent to the procedure we discussed above when the fused model is initialized to model B, i.e., $\widehat{M}_\mathcal{F} \leftarrow M_B$. Because, aligning model B with this estimate of the fused model will yield a (scaled) transport map equal to the identity. And then, Eq.~\eqref{eq:avg} will amount to returning the average of the aligned weights. 

\paragraph{Alignment strategies.} Now, we discuss how to design the ground metric $D_\GS$ between the inter-model neurons. Hence, we branch out into the following two strategies to \textsc{GetSupport}: 

\textit{~~(a) Activation-based alignment ($\psi = \text{`acts'}$):} In this variant, we run inference over a set of $m$ samples, $S = \lbrace\xx\rbrace_{i=1}^m$ and store the activations for all neurons in the model. Thus, we consider the neuron activations, concatenated over the samples into a vector, as the support of the measures, and we denote it as $\mX_k \leftarrow \textsc{acts}\big( M_k(S)\big), \, \mY \leftarrow \textsc{acts}\big(M_{\mathcal{F}}(S)\big)$. Then the neurons across the two models are considered to be similar if they produce similar activation outputs for the given set of samples. We measure this by computing the Euclidean distance between the resulting vector of activations. This serves as the ground metric for optimal transport computations. 
In practice, we use the pre-activations.

\textit{~~(b) Weight-based alignment $\psi = \text{`wts'}$): } Here, we consider that the support of each neuron is given by the weights  of the incoming edges (stacked in a vector). Thus, a neuron can be thought as being represented by the row corresponding to it in the weight matrix. So, the support of the measures in such an alignment type is given by, $\mX_k{\lbrack\ell\rbrack} \leftarrow \widehat{\mW}^{(\ell, \, \ell-1)}_k,  \; \mY{\lbrack\ell\rbrack} \leftarrow  \widehat{\mW}^{(\ell, \,\ell-1)}_{\mathcal{F}}$. The reasoning for such a choice stems from the  neuron activation at a particular layer being calculated as the inner product between this weight vector and the previous layer output. 
The ground metric then used for OT computations is again the Euclidean distance between weight vectors corresponding to the neurons $p$ of \ma $\,$ and $q$ of \mb$\,$  (see \textsc{line 12} of Algorithm \ref{alg:act_algo}). Besides this difference of employing the actual weights in the ground metric (\textsc{line 6, 10}), rest of the procedure is identical.

Lastly, the overall procedure is summarized in Algorithm \ref{alg:act_algo} ahead, where the $\textsc{GetSupport}$ selects between the  above strategies based on the value of $\psi$.

% !TEX root = text_neurips.tex
\begin{algorithm}[h]
	\caption{Model Fusion (with $\psi=\{\text{`acts'}, \text{`wts'}\}-$alignment)}\label{alg:act_algo}
\begin{algorithmic}[1]
	\vspace{1mm}
    \Inp{Trained models $\lbrace M_k\rbrace_{k=1}^{K}$ and initial estimate of the fused model $\widehat{M}_{\mathcal{F}}$
    } 
	\vspace{1mm}
    \out{Fused model $M_{\mathcal{F}}$ with weights $\bm{W}_\mathcal{F}$}
    \vspace{1mm}
	\Notation{For model $M_k$, size of the layer $\ell$ is written as $n^{(\ell)}_k$, and the weight matrix between the layer~$\ell$ and $\ell-1$ is denoted as $\mW^{(\ell, \,\ell-1)}_k$. Neuron support tensors are given by $\mX_k, \mY$.}
	    \vspace{1mm}
    \Initialize{The size of input layer $n^{(1)}_k \leftarrow m^{(1)}$ for all $k \in [K]$;  so $\vec{\alpha}^{(1)}_k = \vec{\beta}^{(1)} \leftarrow \1_{m^{(1)}}/m^{(1)}$} and  the transport map is defined as $\mT_k^{(1)} \leftarrow \text{diag}({\vec{\beta}^{(1)}}) \;\; \mathcal{I}_{m^{(1)}  \times m^{(1)}}$. 
    \vspace{1mm}
   	\vspace{1mm}
\For{each layer $\ell=2,\ldots, L$}
		\vspace{2mm}
		 \State \makebox[20mm][l]{$\vec{\beta}^{(\ell)}, \;\mY[\ell] $} $\leftarrow  \1_{m^{(\ell)}}/m^{(\ell)}, \; \Call{GetSupport}{\widehat{M}_\mathcal{F}, \psi, \ell }$  
		\vspace{1mm}
		 \State \makebox[20mm][l]{$\nu^{(\ell)}  $} $\leftarrow  \big(\vec{\beta}^{(\ell)}, \;\mY[\ell]\big)$  
		 \Comment{{\small\textcolor{blue}{ Define probability measure for initial fused model $\widehat{M}_\mathcal{F}$}}}		   
		\vspace{2mm}

		 \For{each model $k=1,\ldots, K$}
			
			\vspace{1mm}
			
		    \State \makebox[20mm][l]{$\widehat{\mW}^{(\ell, \, \ell-1)}_k$}  $ \leftarrow \mW^{(\ell, \,\ell-1)}_k \mT_k^{(\ell-1)} \text{diag}\big(\frac{1}{{\vec{\beta}^{(\ell-1)}}}\big)$
		    \vspace{1mm} 
		    \Comment{\textcolor{blue}{Align incoming edges for $M_k$ }}
			
			 \State \makebox[20mm][l]{$\vec{\alpha}_k^{(\ell)}, \; \mX_k[\ell]       $} $\leftarrow \1_{n^{(\ell)}_k}/n^{(\ell)}_k, \; \Call{GetSupport}{M_k, \psi, \ell }$  
		    \vspace{1mm}
		     \State \makebox[20mm][l]{$\mu_k^{(\ell)}  $} $\leftarrow \big(\vec{\alpha}_k^{(\ell)}, \; \mX_k[\ell] \big)$  
		    \Comment{\textcolor{blue}{Define probability measure for model $M_k$ }}		   
		        \vspace{1mm}
			\vspace{1mm}
		    \label{eq:gm-algo}\State \makebox[20mm][l]{$D^{(\ell)}_\GS[p, q]$} $\leftarrow \| \mX_k[\ell][p] - \mY[\ell][q] \|_2, \; {\scriptstyle\forall \, p\,\in [n_k^{(\ell)}], \, q\,\in [m^{(\ell)}]}$ 
               \Comment{\textcolor{blue}{Form ground metric}}
               		\vspace{2mm}
		    \State \makebox[20mm][l]{$\mT_k^{(\ell)}, \; \mathcal{W}_2^{(\ell)} \ $} $\leftarrow \text{OT}\big(\mu_k^{(\ell)}, \nu^{(\ell)}, D^{(\ell)}_\GS\big)$
 			     \Comment{\textcolor{blue}{Compute OT map and distance}}			
		               	\vspace{1mm}

		    \State \makebox[20mm][l]{
		    	$\widetilde{\mW}^{(\ell, \,\ell-1)}_k$}   $\leftarrow \text{diag}\big(\frac{1}{{\vec{\beta}^{(\ell)}}}\big) {\mT^{(\ell)}}^{\top}  \widehat{\mW}^{(\ell,\, \ell-1)}_k$
	         	\Comment{\textcolor{blue}{Align model $M_k$ neurons}}         
	         	\vspace{1mm}
     	\EndFor
     	\vspace{2mm}
		
		\State \makebox[20mm][l]{$
	    	\mW^{(\ell, \,\ell-1)}_\mathcal{F}$} $\leftarrow \frac{1}{K} \sum_{k=1}^K \widetilde{\mW}^{(\ell,\, \ell-1)}_k$
		\vspace{1mm}
		\Comment{\textcolor{blue}{Average model weights}}
               	\vspace{1mm}
    \EndFor
    \end{algorithmic}
\end{algorithm}

\subsection{Discussion}
\textbf{Pros and cons of alignment type.} An advantage of the weight-based alignment is that it is independent of the dataset samples, making it useful in privacy-constrained scenarios. On the flip side, the activation-based alignment only needs unlabeled data, and an interesting prospect for a future study would be to utilize synthetic data. But, activation-based alignment may help tailor the fusion to certain desired kinds of classes or domains.
Fusion results for both are nevertheless quite similar (c.f. Table \ref{tab:act-wt-mnist}).

\textbf{Combinatorial hardness of the ideal procedure.} In principle, we should actually search over the space of permutation matrices, jointly across all the layers. But this would be  computationally intractable for models such as deep neural networks, and thus we fuse in a layer-wise manner and in a way have a greedy procedure. 

\textbf{\# of samples used for activation-based alignment.} 
We typically consider a mini-batch of  $\sim 100$ to $400$ samples for these experiments. Table  \ref{tab:act-wt-mnist} in the Appendix, shows that effect of increasing this mini-batch size on the fusion performance and we find that  
even as few as $25$ samples are enough to outperform vanilla averaging.

\textbf{Exact OT and runtime efficiency:} Our fusion procedure is efficient enough for the deep neural networks considered here (\textsc{VGG11}, \textsc{ResNet18}), so we primarily utilize exact OT solvers. While the runtime of exact OT is roughly cubic in the cardinality of the measure supports, it is not an issue for us as this cardinality (which amounts to the network width)  is $\le 600 $ for these networks.  In general, modern-day neural networks are typically deeper than wide. To give a concrete estimate, the \textit{time taken to fuse six \textsc{VGG11} models is $\approx 15$ seconds} on 1 Nvidia V100 GPU (c.f. Section \ref{sec:app-time} for more details). It is possible to further improve the runtime by adopting the entropy-regularized OT~\cite{cuturi2013sinkhorn}, but this looses slightly in terms of test accuracy compared to exact OT (c.f. Table \ref{tab:exact-ot}).

\vspace{-1mm}
\section{Experiments}\label{sec:empirical}

\paragraph{Outline.} We first present our results for one-shot fusion when the models are trained on \textit{different data distributions}. Next, in Section~\ref{sec:compression}, we consider (one-shot) fusion in the case when model sizes are different (i.e., unequal layer widths to be precise). In fact, this aspect \textit{facilitates a new tool} that can be applied in ways not possible with vanilla averaging.  Further on, we focus on the use-case of obtaining an \textit{efficient} replacement for ensembling models in Section~\ref{sec:ensemble}. Lastly, in Section~\ref{sec:teacher-student} we present fusion in the teacher-student setting, and compare OT fusion and distillation in that context.
\vspace{-1mm}
\paragraph{Empirical Details.}  We test our model fusion approach on standard image classification datasets, like  CIFAR10 with commonly used convolutional neural networks (CNNs) such as \textsc{\textsc{\textsc{VGG11}}} \citep{simonyan2014deep}  and residual networks like ResNet18  \cite{He_2016}; and on MNIST, we use a fully connected network with 3 hidden layers of size $400, 200, 100$, which we refer to as \textsc{MLPNet}.
As baselines, we mention the performance of `prediction' ensembling and `vanilla' averaging, besides that of individual models.  Prediction ensembling refers to keeping all the models and averaging their predictions (output layer scores), and thus reflects in a way the ideal (but unrealistic) performance that we can hope to achieve when fusing into a single model. 
Vanilla averaging denotes the direct averaging of parameters. All the performance scores are test accuracies. Full experimental details are provided in Appendix \ref{sec:app-exp-details}.

\subsection{Fusion in the setting of heterogeneous data and tasks
}\label{sec:one-shot}
We first consider the setting of merging two models A and B, but assume that model A has some special skill or knowledge (say, recognizing an object) which B does not possess. However, B is overall more powerful across the remaining set of skills in comparison to A. The goal of fusion now is to obtain a single model that can gain from the strength of B on overall skills and also acquire the specialized skill possessed by A. 
Such a scenario can arise e.g. in reinforcement learning where these models are agents that have had different training episodes so far. Another possible use case lies in federated learning~\citep{mcmahan2016communicationefficient}, where model A is a client application that has been trained to perform well on certain tasks (like personalized keyword prediction) and model B is the server that typically has a strong skill set for a range of tasks (general language model).

The natural constraints in such scenarios are (a) ensuring privacy and (b) minimization communication frequency. This implies that the training examples can not be shared between A and B to respect privacy and a one-shot knowledge transfer is ideally desired, which eliminates e.g., joint training. 

\begin{figure}[t]\vspace{-5mm}
	\centering    
	\subfigure[Different initialization]{\label{fig:diff_init}
		\includegraphics[width=0.38\textwidth]{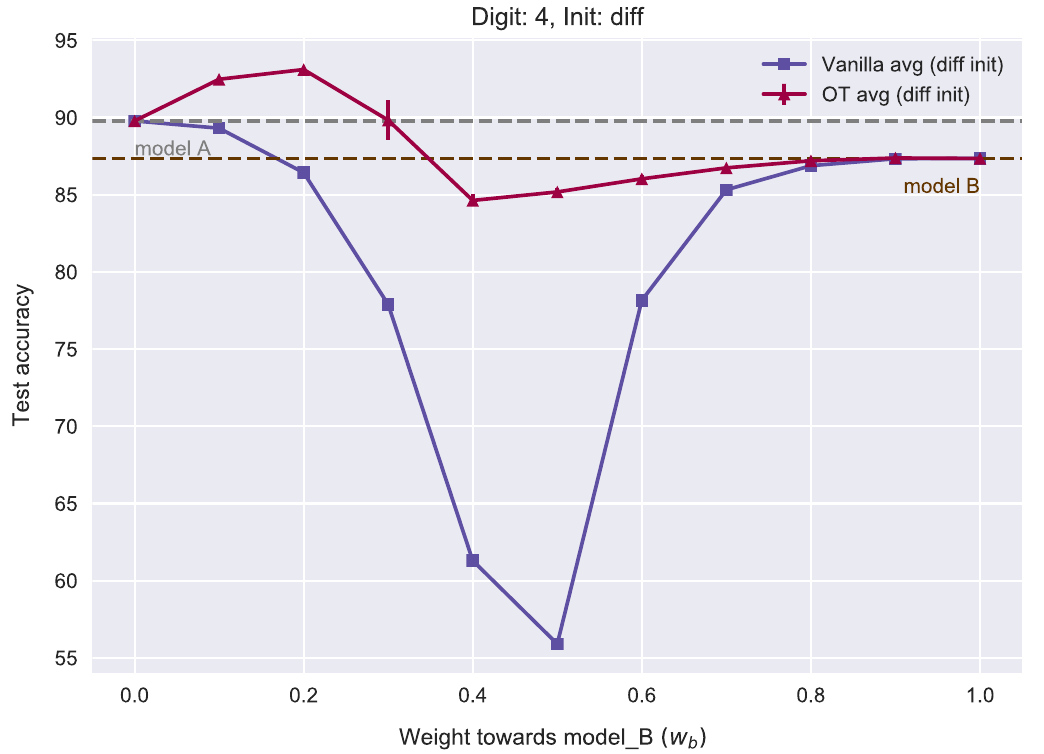}
	\vspace{-2mm}}
	\subfigure[Same initialization ]{\label{fig:same_init}\includegraphics[width=0.38\textwidth]{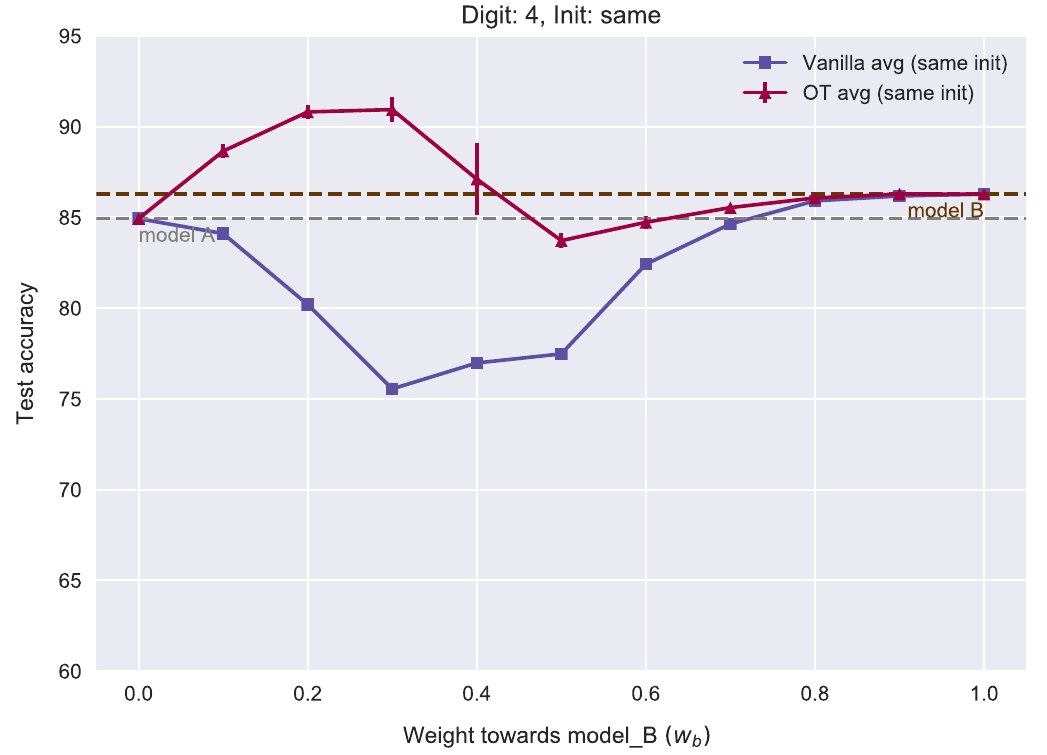}
	\vspace{-2mm}}
	\caption{\textbf{One-shot skill transfer performance} when the specialist model A and the generalist model B are fused in varying proportions ($w_B$), for different and same initializations.  
		The OT avg. (fusion) curve (in magenta) is obtained by activation-based alignment and we plot the mean performance over 5 seeds along with the error bars for standard deviation. \textit{No retraining is done here.}\vspace{-4mm}}
\label{fig:skill-transfer}	
\end{figure}

At a very abstract level, these scenarios are representative of aggregating models that have been trained on non-i.i.d data distributions. To simulate a heterogeneous data-split, we consider the MNIST digit classification task with \mlp$\,$ models, where the unique skill possessed by model A corresponds to recognizing one particular `personalized' label (say $4$), which is unknown to B. Model B contains $90\%$ of the remaining training set (i.e., excluding the label $4$), while A has the other $10\%$. Both are trained on their portions of the data for 10 epochs , and other training settings are identical.  

Figure~\ref{fig:skill-transfer}	illustrates the results for fusing models A and B (in different proportions), both when they have different parameter initializations or when they share the same initialization. OT fusion \footnote{Only the receiver A's own examples are used for computing the activations, avoiding the sharing of data.} significantly outperforms the vanilla averaging of their parameters in terms of the overall test accuracy in both the cases, and also improves over the individual models. E.g., in Figure~\ref{fig:diff_init}, where
the individual models obtain $89.78\%$ and $87.35\%$ accuracy respectively on the overall (global) test set, OT avg. achieves the best overall test set accuracy of $93.11\%$. Thus, confirming the successful skill transfer from both parent models, without the need for any retraining.

Our obtained results are robust to other scenarios when (i) some other label (say $6$) serves as the special skill and (ii) the $\%$ of remaining data split is different. These results are collected in the  Appendix \ref{sec:app-skill}, where in addition we also present results without the special label as well.

\paragraph{The case of multiple models.} In the above example of two models, one might also consider maintaining an ensemble, however the associated costs for ensembling become prohibitive as soon as the numbers of models increases. Take for instance, four models: A, B, C and D, with the same initialization and assume that A again possessing the knowledge of a special digit (say, 4). Consider that the rest of the data is divided as $10\%, 30\%, 50\%, 10\%$. Now training in the similar setting as before, these models end up getting (global) test accuracies of $87.7\%, 86.5\%, 87.0\%,83.5\%$ respectively. Ensembling the predictions yields $95.0\%$ while vanilla averaging obtains $80.6\%$. In contrast, OT averaging results in $\textbf{93.6\%}$ test accuracy ($\approx6\%$ gain over the best individual model), while being $4\times$ more efficient than ensembling. Further details can be found in the Appendix \ref{sec:app-mult-skill}.

\subsection{Fusing different sized models}\label{sec:compression}
An advantage of our OT-based fusion is that it allows the layer widths to be different for each input model. Here, our procedure first identifies which weights of the bigger model should be mapped to the smaller model (via the transport map), and then averages the aligned models (now both of the size of the smaller one). We can thus combine the parameters of a bigger network into a smaller one, and vice versa, allowing new use-cases in (a) model compression and (b) federated learning.

\begin{wrapfigure}{R}{0.5\textwidth}
	\centering   
\includegraphics[width=\textwidth]{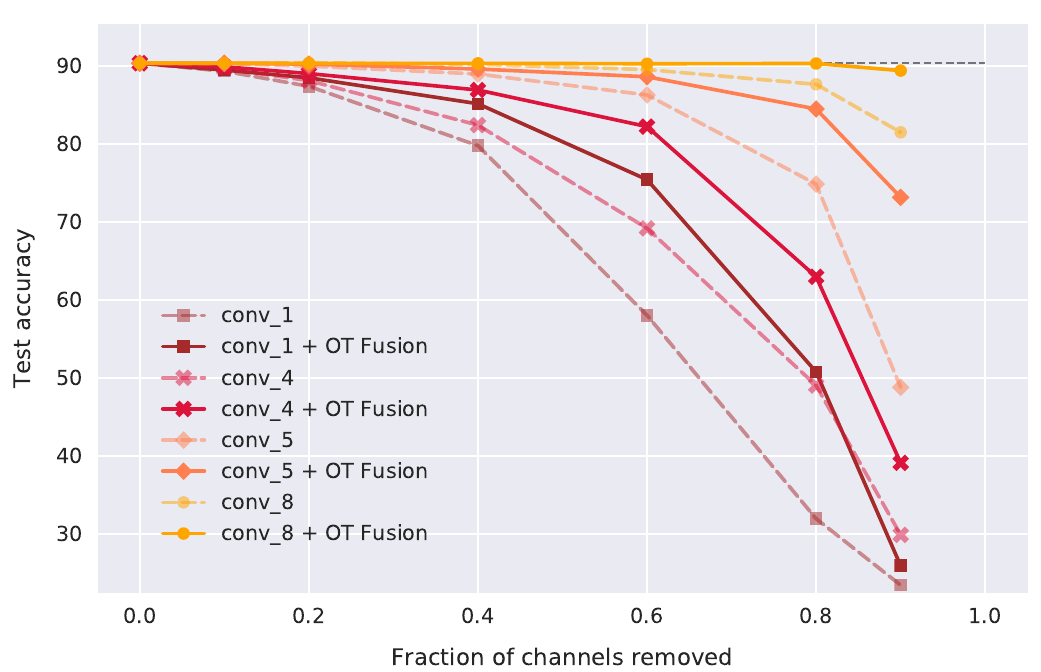}
	\caption{\textbf{Post-processing for structured pruning}: 
		Fusing the initial dense \textsc{VGG11} model into the pruned model helps test accuracy of the pruned model on \textsc{CIFAR10}.}
	\label{fig:pruning_l1}
\end{wrapfigure}

\paragraph{(a) Post-processing tool for structured pruning.} Structured pruning~\citep{li2016pruning,Molchanov_2019_CVPR,10.1145/3005348} is an approach to model compression that aims to remove entire neurons or channels, resulting in an out-of-the-box reduction in inference costs, while affecting the performance minimally. A widely effective method for CNNs is to remove the filters with smallest $\ell_1$ norm ~\cite{li2016pruning}. \textit{Our key idea in this context is to fuse the original dense network into the pruned network, instead of just throwing it away. }

Figure~\ref{fig:pruning_l1} shows the gain in test accuracy on \textsc{CIFAR10} by carrying out  OT fusion procedure (with weight-based alignment) when different convolutional layers of \textsc{VGG11} are pruned to increasing amounts. For all the layers, we consistently obtain a significant improvement in performance, and $\approx10\%$ or more gain in the high sparsity regime. We also observe similar improvements other layers as well as when multiple (or all) layers are pruned simultaneously (c.f. Appendix~\ref{sec:app-pruning}). 

Further, these gains are also significant when measured with respect to the overall sparsity obtained in the model. E.g., structured pruning the \textsc{conv\_8} to $90\%$ results in a net sparsity of $23\%$ in the model. 
After this pruning, the accuracy of the model drops from $90.3\%$ to $81.5\%$, and on applying OT fusion, the performances recovers to $89.4\%$. As  another example take \textsc{conv\_7}, where after structured pruning to $80\%$, OT fusion improves the performance of the pruned model from $87.6\%$ to $90.1\%$ while achieving an overall sparsity of $41\%$ in the network (see ~\ref{sec:app-pruning}).

Our goal  here is not to propose a method for structured pruning, but rather a post-processing tool that can help regain the drop in performance due to pruning. These results are thus independent of the pruning algorithm used, and e.g., Appendix~\ref{sec:app-pruning} shows similar gains when the filters are pruned based on $\ell_2$ norm (Figure~\ref{fig:l2_all}) or even randomly (Figure~\ref{fig:random_all}). Further, Figure~\ref{fig:l1_all_cifar100} in the appendix also shows the results when applied to VGG11 trained on CIFAR100 (instead of CIFAR10). Overall, OT fusion offers a\textit{ completely data-free approach} to improving the performance of the pruned model, which can be handy in the limited data regime or when retraining is prohibitive.

\paragraph{(b) Adapting the size of client and server-side models in federated learning.}\label{sec:client_server}  
\begin{wrapfigure}{R}{0.43\textwidth}
	\vspace{-3mm}
	\centering
	\includegraphics[width=0.95\textwidth]{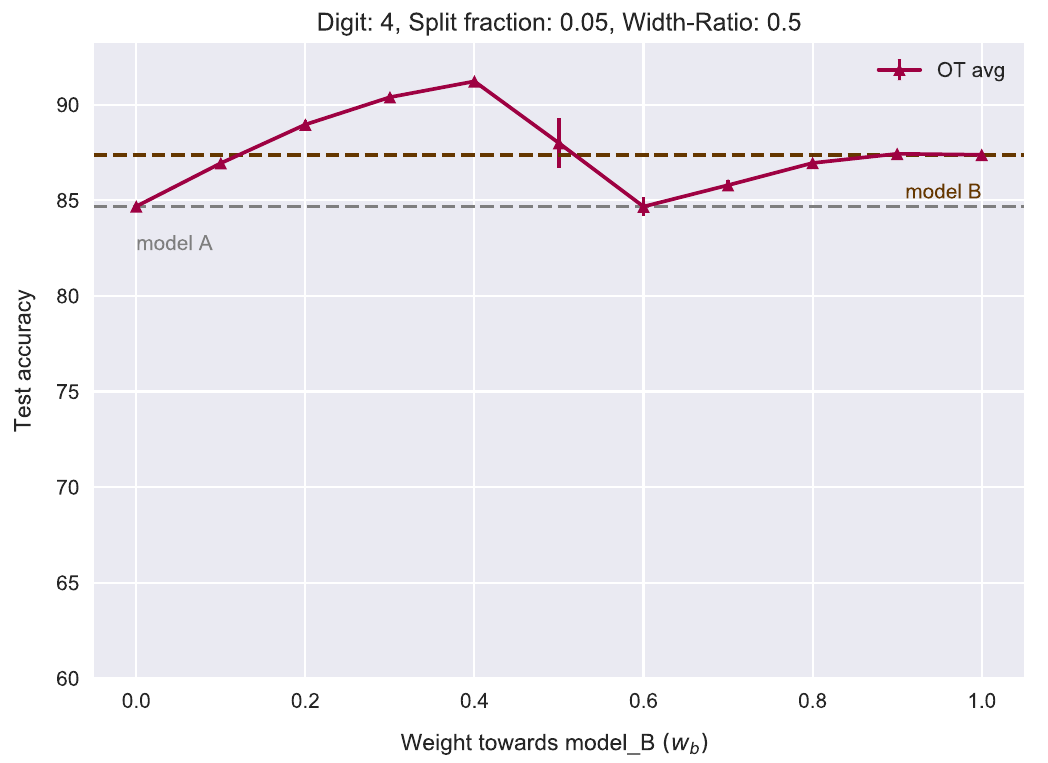}
	\caption{\textbf{One-shot skill transfer for different sized models}: Results of fusing the small client model A into the larger server model B, for varying proportions $w_B$ in which they are fused. See Appendix \ref{app:skill_transfer_size} for more details. 
		\vspace{-2mm}
	}
	\label{fig:skill-transfer-widthratio}
	
\end{wrapfigure}

Given the huge sizes of contemporary neural networks, it is evident that we will not able to fit the same sized model on a client device as would be possible on the server. However, this might come at the cost of reduced performance. Further, the resource constraints might be fairly varied even amongst the clients devices, thus necessitating the flexibility to adapt the model sizes.

We consider a similar formulation, as in the one-shot knowledge transfer setting from Section~\ref{sec:one-shot}, except that now the model B has twice the layer widths as compared to the corresponding layers of model A. Vanilla averaging of parameters, a core component of the widely prevalent FedAvg algorithm \cite{mcmahan2016communicationefficient}, gets ruled out in such a setting. Figure~\ref{fig:skill-transfer-widthratio} shows how OT fusion/average can still lead to a successful knowledge transfer between the given models. 

\vspace{2mm}
\subsection{Fusion for  Efficient Ensembling}\label{sec:ensemble}
\vspace{2mm}
\subsubsection{The case of two Models}
In this section, our goal is to obtain a single model which can serve as a proxy for an ensemble of models, even if it comes at a slight decrease in performance relative to the ensemble, \textit{for future efficiency}. Specifically, here we investigate how much can be gained by fusing multiple models that differ only in their parameter initializations (i.e., seeds). This means that models are trained on the same data, so unlike in Section~\ref{sec:one-shot} with a heterogeneous data-split, the gain here might be limited.

\begin{wraptable}{R}{0.6\textwidth}\ra{1.2}
	\centering
	\resizebox{\textwidth}{!}{
		\begin{tabular}{@{}c|cc|ccc|cc@{}}
			\toprule
			\multirow{1}{*}{\textsc{Dataset +}} & \multirow{2}{*}{\textsc{\ma}} & \multirow{2}{*}{\textsc{\mb}} & \multirow{1}{*}{\textsc{Prediction}} & \multirow{1}{*}{\textsc{Vanilla} } & \multirow{1}{*}{\textsc{OT}} &  \multicolumn{2}{c}{\textsc{Finetuning}}\\
			
			\multirow{1}{*}{\textsc{Model}}	& & & \multirow{1}{*}{\textsc{avg.}} & \multirow{1}{*}{\textsc{avg.} } & \multirow{1}{*}{\textsc{avg. }} &  \multirow{1}{*}{\textsc{Vanilla}} & \multirow{1}{*}{\textsc{OT}}\\
			\midrule
			
			\cifar  $\,$+ &	\multirow{1}{*}{90.31} & \multirow{1}{*}{90.50}  & \multirow{1}{*}{{91.34}} & \multirow{1}{*}{17.02}  &  \multirow{1}{*}{85.98} & \multirow{1}{*}{90.39} & \multirow{1}{*}{{\textbf{90.73}}} \\
			\vgg &	\multicolumn{2}{c|}{1 $\times$}  &1 $\times$ & 2 $\times$ & 2 $\times$ & 2 $\times$  & \textbf{2 $\times$} \\
			\midrule
			\cifar  $\,$+ &	\multirow{1}{*}{93.11} & \multirow{1}{*}{93.20}  & \multirow{1}{*}{{93.89}} & \multirow{1}{*}{18.49}  &  \multirow{1}{*}{77.00} &  \multirow{1}{*}{93.49} & \multirow{1}{*}{{\textbf{93.78}}} \\
			\resnet &	\multicolumn{2}{c|}{1 $\times$}  &1 $\times$ & 2 $\times$ & 2 $\times$ & 2 $\times$  & \textbf{2 $\times$} \\
			
			\bottomrule
		\end{tabular}
	}
	\caption{Results for fusing convolutional \& residual networks, along with the effect of finetuning the fused models, on \cifar. The number below the test accuracies indicate the factor by which a fusion technique is efficient over maintaining all the given models. 
	}
	\label{tab:act_retrain_cifar}
\end{wraptable}

 We study this in context of deep networks such as \textsc{VGG11} and \textsc{ResNet18} which have been trained to convergence on \cifar. As a first step, we consider the setting when we are given just two models, the results for which are present in Table \ref{tab:act_retrain_cifar}. We observe that vanilla averaging absolutely fails in this case, and is 3-5$\times$ worse than OT averaging, in case of \resnet $\,$ and \vgg $\,$ respectively. OT average, however, does not yet improve over the individual models. This can be attributed to the combinatorial hardness of the underlying alignment problem, and the greedy nature of our algorithm as mentioned before. As a simple but effective remedy, we consider finetuning (i.e., retraining) from the fused or averaged models. Retraining helps for both vanilla and OT averaging, but in comparison, the OT averaging results in a better score for both the cases as shown in Table \ref{tab:act_retrain_cifar}. E.g., for \resnet, OT avg. + finetuning gets almost as good as prediction ensembling on test accuracy.

The finetuning scores for vanilla and OT averaging correspond to their best obtained results, when retrained with several finetuning learning rate schedules for a total of 100 and 120 epochs in case of \vgg and \resnet $\,$ respectively. We also considered finetuning the individual models across these various hyperparameter settings (which of course will be infeasible in practice), but the best accuracy mustered via this attempt for \textsc{ResNet18} was 93.51, in comparison to 93.78 for OT avg. + finetuning. See Appendix \ref{sec:app-retrain-all} and \ref{sec:app-retrain-curve} for detailed results and typical retraining curves.

\subsubsection{The Multiple Models  ($>2$) case}
Now, we discuss the case of more than two models, where the savings in efficiency relative to the ensemble are even higher. As before, we take the case of \textsc{VGG11} on \cifar~ and additionally \textsc{CIFAR100}~\footnote{We simply adapt the \textsc{VGG11} architecture used for \textsc{CIFAR10} and train it on \textsc{CIFAR100} for 300 epochs. Since our focus here was not to obtain best individual models, but rather to investigate the efficacy of fusion.}, but now consider $\lbrace4, 6, 8\rbrace -$ such models that have been trained to convergence, each from a different parameter initialization. Table~\ref{tab:act_retrain_cifar100_mult} shows the results for this in case of \textsc{CIFAR100} (results for \textsc{CIFAR10} are similar and can be found in Table~\ref{tab:act_retrain_cifar_mult}). 

 We find that the performance of vanilla averaging degrades to close-to-random performance, and interestingly even fails to retrain, despite trying numerous settings of optimization hyperparameters (like learning rate and schedules, c.f. Section~\ref{app:cifar_mult_opt_try}). In contrast, OT average performs significantly better even without fine-tuning, and results in a mean test accuracy gain $\sim \lbrace1.4\%, 1.7\%, 2\%\rbrace$ over the best individual models after fine-tuning, in the case of $\lbrace4, 6, 8\rbrace -$ base models respectively.

\begin{table}[h!] \centering\ra{1.2}
	\centering
	\resizebox{\textwidth}{!}{
		\begin{tabular}{@{}c|c|ccc|cc@{}}
			\toprule
			\multirow{1}{*}{\textit{\cifar +}} & \multirow{2}{*}{\textsc{Individual Models}}  & \multirow{1}{*}{\textsc{Prediction}} & \multirow{1}{*}{\textsc{Vanilla} } & \multirow{1}{*}{\textsc{OT}} &  \multicolumn{2}{c}{\textsc{Finetuning}}\\
			
			\multirow{1}{*}{\textsc{\vgg }} &   & \multirow{1}{*}{\textsc{avg.}} & \multirow{1}{*}{\textsc{avg.} } & \multirow{1}{*}{\textsc{avg. }} &  \multirow{1}{*}{\textsc{Vanilla}} & \multirow{1}{*}{\textsc{OT}}\\
			\midrule
			Accuracy  &	\multirow{1}{*}{[90.31, 90.50, 90.43, 90.51]}  & \multirow{1}{*}{91.77} & \multirow{1}{*}{10.00}  &  \multirow{1}{*}{73.31} & \multirow{1}{*}{12.40} & \multirow{1}{*}{{90.91}} \\
			
			Efficiency &	1 $\times$& 1 $\times$ & 4 $\times$ & 4 $\times$  & 4 $\times$  & 4 $\times$    \\
			\midrule
			Accuracy  &	\multirow{1}{*}{[90.31, 90.50, 90.43, 90.51, 90.49, 90.40]} & \multirow{1}{*}{91.85}  & \multirow{1}{*}{10.00}  &  \multirow{1}{*}{72.16} &  \multirow{1}{*}{11.01} & \multirow{1}{*}{{91.06}} \\
			Efficiency &	1 $\times$& 1 $\times$ & 6 $\times$ & 6 $\times$  & 6 $\times$  & 6 $\times$    \\
			
			\bottomrule
		\end{tabular}
	}
	\caption{\small Results of our OT average + finetuning based efficient alternative for ensembling in contrast to vanilla average + finetuning, for more than two input models (\vgg) with different initializations. 
	}
	\label{tab:act_retrain_cifar_mult}
\end{table}

Overall, Tables~\ref{tab:act_retrain_cifar} and \ref{tab:act_retrain_cifar_mult} show the importance of aligning the networks via OT before averaging. Further finetuning of the OT fused model, always results in an improvement over the individual models while being number of models times more efficient than the ensemble.

\begin{table}[h!] \centering\ra{1.2}

			\resizebox{\textwidth}{!}{
				\vspace{-2.5em}
				\begin{tabular}{@{}c|c|c|cc@{}}
					\toprule
					\multirow{1}{*}{\textit{\textsc{CIFAR100} +}} & \multirow{2}{*}{\textsc{Individual Models}}  & \multirow{1}{*}{\textsc{Prediction}}  &  \multicolumn{2}{c}{\textsc{Finetuning}}\\
					
					\multirow{1}{*}{\textsc{\vgg }} &   & \multirow{1}{*}{\textsc{avg. }} &  \multirow{1}{*}{\textsc{Vanilla}} & \multirow{1}{*}{\textsc{OT}}\\
					\midrule
					Accuracy  &	\multirow{1}{*}{[62.70, 62.57, 62.50, 62.92]}  & \multirow{1}{*}{66.32} & \multirow{1}{*}{4.02} & \multirow{1}{*}{{\textbf{64.29}$\pm$ \textbf{0.26}}} \\
					
					Efficiency &	1 $\times$& 1 $\times$ & 4 $\times$  & \textbf{4 $\times$  }  \\
					\midrule
					Accuracy  &	\multirow{1}{*}{[62.70, 62.57, 62.50, 62.92, 62.53, 62.70]}  & \multirow{1}{*}{66.99} & \multirow{1}{*}{0.85} & \multirow{1}{*}{{\textbf{64.55} $\pm$ \textbf{0.30}}} \\
					
					Efficiency &	1 $\times$& 1 $\times$ & 6 $\times$  & \textbf{6 $\times$ }   \\
					\midrule
					
					Accuracy  &	\multirow{1}{*}{[62.70, 62.57, 62.50, 62.92, 62.53, 62.70, 61.60, 63.20]}  & \multirow{1}{*}{67.28} & \multirow{1}{*}{1.00} & \multirow{1}{*}{{\textbf{65.05}$\pm$ \textbf{0.53}}} \\
					
					Efficiency &	1 $\times$& 1 $\times$ & 8 $\times$  & \textbf{8 $\times$ }   \\
					
					\bottomrule
				\end{tabular}
			}
			\caption{\small Efficient alternative to ensembling via OT fusion on \textsc{\textbf{CIFAR100}} for \textsc{VGG11}. Vanilla average fails to retrain. Results shown are mean $\pm$ std. deviation over \textbf{5 seeds}.}
		\label{tab:act_retrain_cifar100_mult}
		
\end{table}

\subsubsection{Remarks}

\paragraph{Handling ResNets.} The presence of shortcut connections in the ResNet architecture~\cite{he2016deep} creates two branches: one from the residual block and the other arising from the shortcut side. These branches get accumulated (i.e., added) before going to the outgoing layer. As a result, in the case of ResNets, we will have transport maps flowing from both the branches. Let's call them $\mT_{\text{short}}$ and $\mT_{\text{res}}$ respectively. But for the outgoing layer, we can only pre-multiply by just one of these matrices. While it is possible to enforce the the transport map flowing out form the residual block is the same as $\mT_{\text{short}}$, i.e., the residual block does not introduce further permutations or does not impact the alignment. But instead of presuming this, we employ a simple heuristic : we seek an outgoing map $\mT_{\text{out}}$ that minimizes the distance from both the shortcut side and the residual block. In other words, 
$$
\mT_{\text{out}} := \argmin_{\mT} \quad \beta \, \|\mT - \mT_{\text{short}}\|^2_F \, + \, (1-\beta)\, \|\mT - \mT_{\text{res}}\|^2_F\,.
$$

Moreover, we are constrained to search over the space of transport maps. For simplicity, here we employ the simple choice of $\beta=0.5$, but is likely that a more informed choice (potentially separately for each residual block) could additionally help. Hence, this boils down  our choice to $\mT_{\text{out}} = 0.5 \,  \mT_{\text{short}} + 0.5  \mT_{\text{res}}$. 

\paragraph{Iterative version of Fusion Algorithm.} For non-residual networks, our Algorithm~\ref{alg:act_algo} converges in a single step for activations-based alignment. But, we noticed that in case of residual networks, multiple iterations (in practice, generally $2-3$) can help. The iterative version of the algorithm is nothing but just feeding in the output of Algorithm~\ref{alg:act_algo} as input to itself, in the form of new guess of the fused model estimate. For instance, in the reported one-shot fusion results for \resnet on \cifar, we actually used this iterative version and the accuracy improved by an additional~$10\%$ (when running just one additiional iteration, after which the transport maps converged).

\paragraph{Hard vs Soft Alignment.} This point goes more or less without saying, but to spell it out explicitly: we mainly employ EMD for optimal transport computation, and not the regularized Sinkhorn variant. Hence solutions found with EMD, when the model widths are identical, are in fact based on \textit{permutation matrices}. Therefore, our work can already be seen as pointing towards the potential of linear mode connectivity after correcting for inter-network symmetries with permutations. 

\subsection{Teacher-Student Fusion}\label{sec:teacher-student}

We present the results for a setting where we have pre-trained teacher and student networks, and we would like to transfer the knowledge of the larger teacher network into the smaller student network. This is essentially reverse of the client-server setting described in Section~\ref{sec:client_server}, where we fused the knowledge acquired at the (smaller) client model into the bigger server model. We consider that all the hidden layers of the teacher model \ma $,$ are a constant $\rho \times$ wider than all the hidden layers of student model \mb. Vanilla averaging can not be used due to different sizes of the networks. However, OT fusion is still applicable, and as a baseline we consider finetuning the model \mb.

 We experiment with two instances of this (a) on \mnist $\,$+ \mlp, with $\rho \in \{2, 10\}$ and (b) on \cifar$\,$ + \vgg, with $\rho \in \{2, 8\}$, and the results are presented in the Table \ref{tab:model_compression} (results for \textsc{MNIST} are present in the Table~\ref{tab:model_compression_app}).  We observe that across all the settings, OT avg. + finetuning improves over the original model \mb, as well as outperforms the finetuning of the model \mb, thus resulting in the desired knowledge transfer from the teacher network. 

\begin{table}[h!] \centering\ra{1.3}
\centering
\resizebox{0.6\textwidth}{!}{
	\begin{tabular}{@{}cc|c|cc|cc@{}}
		\toprule
		\multirow{1}{*}{\textit{\textsc{Dataset + }}} & \multirow{1}{*}{\textsc{$\#$ params}}  &\multirow{1}{*}{\textsc{Teacher}} &  \multicolumn{2}{c|}{\textsc{Students}} &  \multicolumn{2}{c}{\textsc{Finetuning}} \\
		\multirow{1}{*}{\textsc{Model}} & \multirow{1}{*}{ (\ma, \mb) }  & \multirow{1}{*}{\textit{\ma}}   &  \multirow{1}{*}{\textit{\mb}}  &\multirow{1}{*}{\textsc{OT avg.}} &  \multirow{1}{*}{\textit{\mb}}& \multirow{1}{*}{\textsc{OT avg.}}\\
		\midrule
		\multirow{1}{*}{\cifar $\,$+} & \multirow{1}{*}{(118 M, 32 M)} &	91.22 &	90.66 & 86.73  & 90.67 & \textbf{90.89}\\
		\multirow{1}{*}{ \vgg} &	\multirow{1}{*}{(118 M, 3 M )} &	91.22 & {89.38} & 88.40	  & 89.64 & \textbf{89.85}\\
		\bottomrule
\end{tabular}}
\caption{\textit{Knowledge transfer from teacher \ma $\,$ into (smaller) student models. }The finetuning results of each method are at their best scores across different finetuning hyperparameters (like, learning rate schedules). OT avg. has the same number of parameters as \mb. Also, here we use activation-based alignment. Further details can be found in Appendix~\ref{app:fusion_size}.}\label{tab:model_compression}
\end{table}

\paragraph{Fusion and Distillation.} Now, we compare OT fusion, distillation, and their combination, in context of transferring the knowledge of a large pre-trained teacher network into a smaller student network. We consider three possibilities for the student model in distillation: (a) randomly initialized network, (b) smaller pre-trained model \mb, and (c) OT fusion (avg.) of the teacher into model \mb. 

We focus on \mnist $\,$ + \mlp $\,$, as it allows us to perform an extensive sweep over the distillation-based hyperparameters (temperature, loss-weighting factor) for each method.   Further, we contrast these distillation approaches with the baselines of simply finetuning the student models, i.e., finetuning \mb$\,$ as well as OT avg. model. Results of these experiments are reported in Table \ref{tab:distill_mnist_main}.
 
 We find that distilling with  OT fused model as the student model yields better performance than initializing randomly or with the pre-trained \mb. Further, when averaged across the considered temperature values $=\{20, 10, 8, 4, 1\}$, we observe that distillation of the teacher into random or \mb$\,$ performs worse than simple OT avg. + finetuning (which also does not require doing such a sweep that would be prohibitive in case of larger models or datasets). These experiments are discussed in detail in Appendix \ref{app:distill_more}. An interesting direction for future work would be to use intermediate OT distances computed during fusion as a means for regularizing or distilling with hidden layers.

\begin{table}[h!]\ra{1.4}
	\centering
	\resizebox{0.6\textwidth}{!}{
		\begin{tabular}{@{}c|cc|cc|ccc@{}}
			\toprule
			\multirow{1}{*}{\textsc{Teacher}} &  \multicolumn{2}{c|}{\textsc{Students}}
 &  \multicolumn{2}{c|}{\textsc{Finetuning}} &\multicolumn{3}{c}{\textsc{Distillation}} \\

		\multirow{1}{*}{\textsc{\ma}} & \multirow{1}{*}{\textsc{\mb}} &   \multirow{1}{*}{\textsc{OT avg.}} & \multirow{1}{*}{\textsc{\mb}} & \multirow{1}{*}{\textsc{OT avg.}}  &\multirow{1}{*}{\textsc{Random}}  &  \multirow{1}{*}{\textsc{\mb}} & \multirow{1}{*}{\textsc{OT avg.}}\\
			
			\midrule

			\multirow{1}{*}{98.11} &	\multirow{1}{*}{97.84}	 & \multirow{1}{*}{95.49}  &  	\multirow{1}{*}{98.04} & 	\multirow{1}{*}{98.19} & 98.18 & 98.22  & \textbf{98.30}\\
			\cmidrule{1-5}
			\multicolumn{5}{c|}{\footnotesize Mean across distillation temperatures}& 98.13  & 98.17 & \textbf{98.26}\\
			\bottomrule
		\end{tabular}
	}
	\caption{\textit{Fusing the bigger teacher model \ma $\,$  to half its size ($\rho=2$). }
		Both finetuning and distillation were run for 60 epochs using SGD with the same hyperparameters. Each entry has been averaged across 4 seeds. 
	}
	\label{tab:distill_mnist_main}
\end{table}

Hence, this suggests that OT fusion + finetuning can go a long way in an efficient knowledge transfer from a bigger model into a smaller one, and can be used alongside when distillation is feasible.

\section{Conclusion}
We show that averaging the weights of models, by first doing a layer-wise (soft) alignment of the neurons via optimal transport, can serve as a versatile tool for fusing models in various settings. This results in (a) successful one-shot transfer of knowledge between models without sharing training data, (b) data free and algorithm independent post-processing tool for structured pruning, (c) and more generally, combining parameters of different sized models. Lastly, the OT average when further finetuned, allows for just keeping one model rather than a complete ensemble of models at inference. Future avenues include application in distributed optimization and continual learning, besides extending our current toolkit to fuse models with different number of layers, as well as, fusing generative models like GANs~\citep{goodfellow2014generative} (where ensembling does not make as much sense). The promising empirical results of the presented algorithm, thus warrant attention for further use-cases.

\section*{Broader Impact} 

Model fusion is a fundamental building block in machine learning, as a way of direct knowledge transfer between trained neural networks. Beyond theoretical interest, it can serve a wide range of concrete applications.
For instance, collaborative learning schemes such as federated learning are of increasing importance for enabling privacy-preserving training of ML models, as well as a better alignment of each individual's data ownership with the resulting utility from jointly trained machine learning models, especially in applications where data is user-provided and privacy sensitive~\citep{kairouz2019advances}.
Here fusion of several models is a key building block to allow several agents to participate in joint training and knowledge exchange. We propose that a reliable fusion technique can serve as a step towards more broadly enabling privacy-preserving and efficient collaborative learning.

\subsection*{Acknowledgments}
We would like to thank R\'emi Flamary, Boris Muzellec, Sebastian Stich and other members of MLO, as well as the anonymous reviewers for their comments and feedback.

\begin{small}
\bibliographystyle{unsrtnat}
\bibliography{references}
\end{small}

\onecolumn
\clearpage
\newpage{}
\appendix
\appendixpage

\renewcommand{\thesection}{S\arabic{section}}
\renewcommand{\thetable}{S\arabic{table}}
\renewcommand{\thefigure}{S\arabic{figure}}
\renewcommand{\thefootnote}{S\arabic{footnote}}
\setcounter{figure}{0}
\setcounter{table}{0}
\setcounter{footnote}{0}

\setlength\cftparskip{2pt}
\setlength\cftbeforesecskip{2pt}
\setlength\cftaftertoctitleskip{3pt}
\addtocontents{toc}{\protect\setcounter{tocdepth}{2}}
\setcounter{tocdepth}{1}
\hypersetup{linkcolor=black}

\tableofcontents

\clearpage

\hypersetup{linkcolor=red}

%\section{Model Fusion Algorithm}\label{sec:app_algo}
%\input{act_algo}

%\vspace{10mm}

%Before going further, note that our code can be found under the following link \href{https://github.com/sidak/otfusion}{https://github.com/sidak/otfusion}.

\section{Technical specifications}\label{sec:app}

\subsection{Experimental Details}\label{sec:app-exp-details}

\paragraph{VGG11 training details.} It is trained by SGD for $300$ epochs with an initial learning rate of $0.05$, which gets decayed by a factor of $2$ after every $30$ epochs. Momentum $=0.9$ and weight decay $=0.0005$. The batch size used is $128$. Checkpointing is done after every epoch and the best performing checkpoint in terms of test accuracy is used as the individual model.  The block diagram of VGG11 architecture is shown below for reference. 

\begin{figure}[h]
	\centering
	\includegraphics[width=0.8\textwidth]{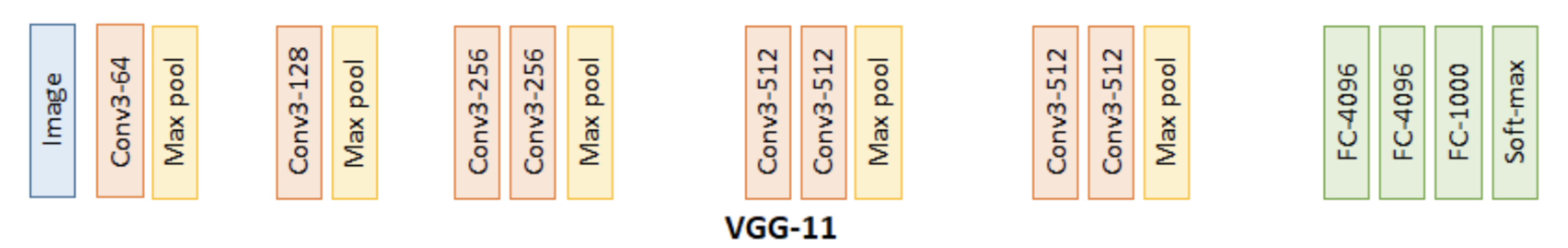}
	\caption{Block diagram of the \textsc{VGG11} architecture. Adapted from \url{https://bit.ly/2ksX5Eq}.}
	\label{fig:vgg}
\end{figure}

\paragraph{\textsc{MLPNet}  training details.} This is also trained by SGD at a constant learning rate of $0.01$ and momentum $=0.5$. The batch size used is $64$. 

\paragraph{\resnet $\,$ training details.} Again, we use SGD as the optimizer, with an initial learning rate of 0.1, which gets decayed by a factor of 10 at epochs $\{150, 250\}$. In total, we train for 300 epochs and similar to the VGG11 setting we use the best performing checkpoint as the individual model. Other than that, momentum $=0.9$, weight decay $=0.0001$, and batch size $=256$. We skip the batch normalization for the current experiments, however, it can possibly be handled by simply multiplying the batch normalization parameters in a layer by the obtained transport map while aligning the neurons. 

\paragraph{Other details.}

\textit{Pre-activations.} The results for the activation-based alignment experiments are based on pre-activation values, which were generally found to perform slightly better than post-activation values. 

\textit{Regularization.} The regularization constant used for the activation-based alignment results in Table \ref{tab:act-wt-mnist} is $0.05$.

\textit{Common details.} The bias of a neuron is set to zero in all of the experiments. It is possible to handle it as a regular weight by keeping the corresponding input as~$1$, but we leave that for future work. 

\subsection{Combining weights and activations for alignment}\label{sec:app-combine}
The output activation of a neuron over input examples gives a good signal about the presence of features in which the neuron gets activated. Hence, one way to combine this information in the above variant with weight-based alignment is to use them in the probability mass values. 

In particular, we can take a mini-batch of samples and store the activations of all the neurons. Then we can use the mean activation as a measure of a neuron's significance. But it might be that some neurons produce very high activations (in absolute terms) irrespective of the kind of input examples. Hence, it might make sense to also look at the standard deviation of activations. Thus, one can combine both these factors into an importance weight for the neuron as follows:

\begin{equation}
\text{importance}_k[2, \cdots, L ] =  \overline{M_k}([x_1, \cdots, x_d]) \odot \sigma(M_k([x_1, \cdots, x_d]))
\end{equation}

Here, $M_k$ denotes the $k^{\text{th}}$ model into which we pass the inputs $[x_1, \cdots, x_d]$, $\overline{M}$ denotes the mean, $\sigma(.)$ denotes the standard deviation and $\odot$ denotes the elementwise product.  Thus, we can now set the probability mass values $b^{(l)}_k \propto \text{importance}_k[l]$, and the rest of the algorithm remains the same.  

\subsection{Optimal Transport }
\label{app:subsec:ot}

We make use of the Python Optimal Transport (POT)\footnote{\textcolor{blue}{\url{http://pot.readthedocs.io/en/stable/}}} for performing the computation of Wasserstein distances and barycenters on CPU. These can also be implemented on the GPU to further boost the efficiency, although it suffices to run on CPU for now, as evident from the timings below. 

\subsection{Timing information}\label{sec:app-time}
The following timing benchmarks are done on 1 Nvidia V100 GPU. The time taken to average two \mlp $\,$  models for \mnist $\,$ is $\approx$ 3 seconds. For averaging \vgg $\,$ models on \cifar, it takes about $\approx$ 5 seconds. While in case of \resnet $\,$ on \cifar, it takes $\approx$ 7 seconds. These numbers are for the activation-based alignment, and also include the time taken to compute the activations over the mini-batch of examples. 

The weight-based alignment can be faster as it does not need to compute the activations. For instance, when weight-based alignment is employed to average two \vgg $\,$ models on \cifar, it takes $\approx$ 2.5 seconds.

\section{Ablation studies}\label{sec:app-ablation}

\subsection{Aggregation performance as training progresses}\label{sec:app-grad}
We compare the performance of averaged models at various points during the course of training the individual models (for the setting of MLPNet on MNIST). We notice that in the early stages of training, vanilla averaging performs even worse, which is not the case for OT averaging. The corresponding Figure \ref{fig:epoch} and Table \ref{tab:act_epoch} can be found in Section \ref{sec:app-grad} of the Appendix. 
Overall, we see OT averaging outperforms vanilla averaging by a large margin, thus pointing towards the benefit of aligning the neurons via optimal transport. 

\begin{figure}[h]
	\centering\vspace{-1em}
	\includegraphics[width=0.7\textwidth]{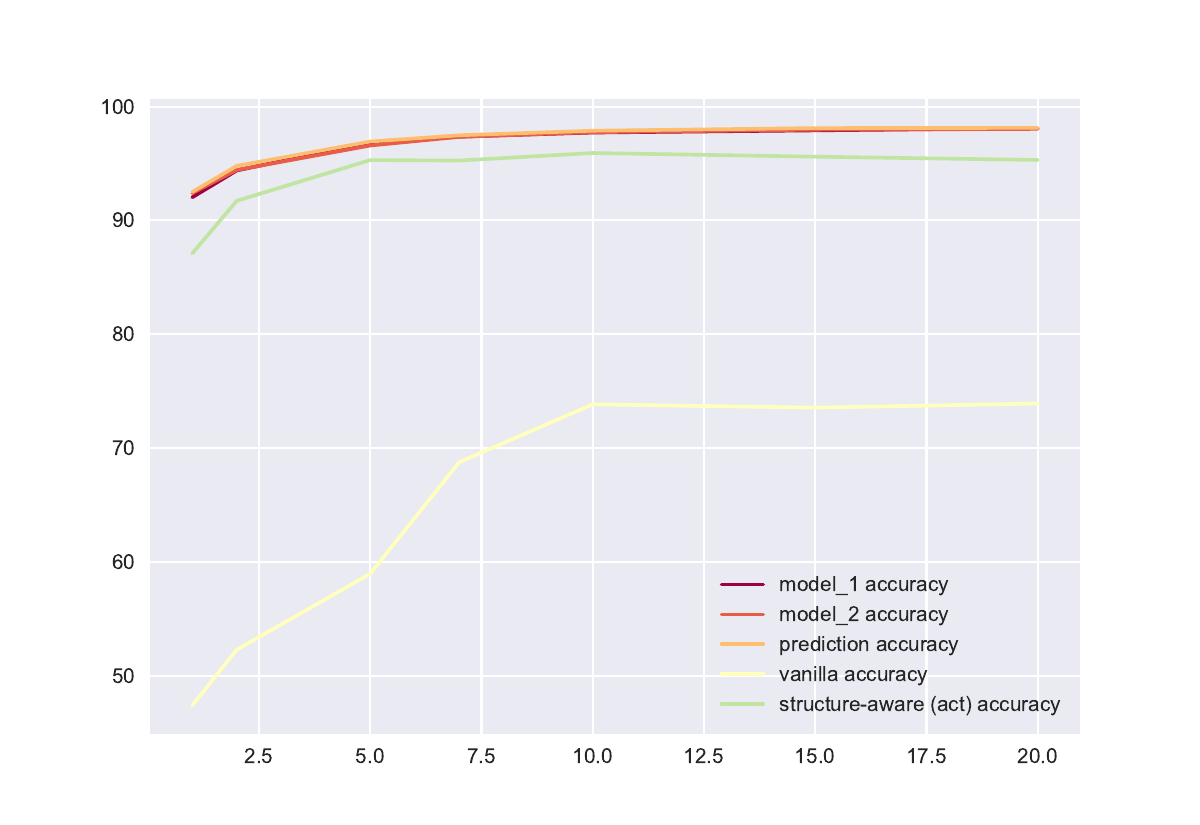}
	\caption{Illustrates the performance of various aggregation methods as training proceeds, for (\mnist, \mlp). The plots correspond to the results reported in Table \ref{tab:act_epoch}. The activation-based alignment of the OT average (labelled as structure-aware accuracy in the figure) is used based on $m=200$ samples.
	}\label{fig:epoch}
	
\end{figure}

\begin{table}[h] \centering\ra{1.2}
	\centering
		\resizebox{0.7\textwidth}{!}{
	\begin{tabular}{@{}c|cc|cc|c@{}}
		\toprule
		\multirow{1}{*}{\textsc{Epoch}} & \multirow{1}{*}{\textsc{Model A}} & \multirow{1}{*}{\textsc{Model B }} & \multirow{1}{*}{\textsc{Prediction avg.}} & \multirow{1}{*}{\textsc{Vanilla avg.}} & \multicolumn{1}{c}{\textsc{OT avg.}} \\
		
		\midrule
		01 & 92.03 &  92.40 & 92.50 &  47.39 & 87.10\\
		02 & 94.39 &  94.43 & 94.79 &  52.28 & 91.72\\
		05 & 96.83 &  96.58 & 96.93 &  58.96 & 95.30 \\
		07 & 97.36 &  97.34 & 97.48 &  68.76 &  95.26 \\
		10 & 97.72 &  97.75 & 97.88 &  73.84 & 95.92 \\
		15 & 97.91 &  97.97 & 98.11 &  73.55 &  95.60 \\
		20 & 98.11 &  98.04 & 98.13 &  73.91 & 95.31 \\
		
		\bottomrule
	\end{tabular}
	\caption{\textbf{Activation-based alignment (MNIST, MLPNet): }Comparison of performance when ensembled after different training epochs. The \# samples used for activation-based alignment, $m=50$. The corresponding plot for this table is illustrated in Figure \ref{fig:epoch}.}
	\label{tab:act_epoch}
}
\end{table}

\subsection{Transport map for the output layer.} 
Since our algorithm runs until the output layer, we inspect the alignment computed for the last output layer. We find that the ratio of the trace to the sum for this last  transport map is $ \approx 1$, indicating accurate alignment as the ordering of output units is the same across models.

\subsection{Effect of mini-batch size needed for activation-based mode}\label{sec:bsz_effect}

Here, the individual models used are \textsc{MLPNet}'s which have been trained for~10 epochs on MNIST. They differ only in their seeds and thus in the initialization of the parameters alone. We ensemble the final checkpoint of these models via OT averaging and the baseline methods.  

\begin{table}[h] \centering\ra{1.2}
	\centering
	\vspace{-0.1em}
	\resizebox{0.7\textwidth}{!}{
		\setlength{\tabcolsep}{5pt}
		\begin{tabular}{@{}lc|cc|c|c|c@{}}
			\toprule
			\multirow{2}{*}{\textit{\ma}} & \multirow{2}{*}{\textsc{\mb}} & \multirow{1}{*}{\textsc{Prediction}} & \multirow{1}{*}{\textsc{Vanilla}} & \multirow{2}{*}{$m$}& \multirow{1}{*}{\textsc{OT avg. (Sinkhorn) }} & \multirow{1}{*}{\textsc{\ma $\,$ aligned}}\\
			&  			&  	\textsc{avg.}		& 		\textsc{avg.}		&  &  \multicolumn{2}{c}{Accuracy (mean $\pm$ stdev)} \\ \midrule
			\multicolumn{7}{l}{\footnotesize \textit{(a) Activation-based Alignment}} \\
			\midrule
			\multirow{6}{*}{97.72} & \multirow{6}{*}{97.75} & \multirow{6}{*}{97.88} & \multirow{6}{*}{73.84}  & 2 & 24.80 $\pm$ 6.93 & 20.08 $\pm$ 2.42 \\
			& & & & 10 & 75.04 $\pm$ 11.35  & 88.18 $\pm$ 8.45\\
			& & & & 25 & 90.95 $\pm$  3.98 & 95.36 $\pm$ 0.96 \\
			& & & & 50 & 93.47 $\pm$ 1.69 &  96.04 $\pm$ 0.59 \\
			& & & & 100 & 95.40 $\pm$ 0.52 & \textbf{97.05 $\pm$ 0.17} \\
			& & & & 200 & \textbf{95.78 $\pm$  0.52 } & 97.01 $\pm$ 0.16 \\
			\midrule
			\multicolumn{7}{l}{\footnotesize  \textit{(b) Weight-based Alignment}} \\
			\midrule
			97.72 & 97.75 &97.88 & 73.84 & --- & 95.66  & 96.32 \\
			
			\bottomrule
	\end{tabular}}\vspace{-0.1em}
	\caption{One-shot averaging for (MNIST, MLPNet) \textbf{with Sinkhorn and regularization $=0.05$}: Results showing the performance (i.e., test classification accuracy (in $\%$)) of the OT averaging in contrast to the baseline methods.	The last column refers to the aligned model A which gets (vanilla) averaged with model B, giving rise to our OT averaged model.	$m$ is the size of mini-batch over which activations are computed. 
	} 
	\label{tab:act-wt-mnist}
\end{table}

\subsection{Effect of regularization}

The results for activation-based alignment presented in the Table~\ref{tab:act-wt-mnist} above use the regularization constant $\lambda=0.05$. Below, we also show the results with a higher regularization constant $\lambda = 0.1$. As expected, we find that using a lower value of regularization constant leads to better results in general, since it better approximates OT. 

\begin{table}[h!] \centering\ra{1.2}
	\centering
\resizebox{0.7\textwidth}{!}{
	\begin{tabular}{@{}lc|cc|c|c|cc@{}}
		\toprule
		\multirow{2}{*}{\textsc{\ma}} & \multirow{2}{*}{\textsc{\mb }} & \multirow{2}{*}{\textsc{Prediction}} & \multirow{2}{*}{\textsc{Vanilla} } & \multirow{2}{*}{$m$}& \multirow{1}{*}{\textsc{OT avg.}} & \multirow{1}{*}{\textit{\ma aligned}}\\
		&  			&  			& 				&  &  \multicolumn{2}{c}{Accuracy (mean $\pm$ stdev)} & \\ \midrule
		\multirow{6}{*}{97.72} & \multirow{6}{*}{97.75} & \multirow{6}{*}{97.88} & \multirow{6}{*}{73.84}  & 2 & 25.05 $\pm$ 7.22 & 19.42 $\pm$ 2.28\\
		& & & & 10 & 72.86 $\pm$ 11.93  & 74.35 $\pm$ 14.40\\
		& & & & 25 & 89.49 $\pm$  5.21 & 90.88 $\pm$ 4.91 \\
		& & & & 50 & 92.88 $\pm$ 2.03 &  94.54 $\pm$ 1.36 \\
		& & & & 100 & 95.14 $\pm$ 0.49 & 96.42 $\pm$ 0.39 \\
		& & & & 200 & \textbf{95.70 $\pm$  0.54} & \textbf{96.63 $\pm$ 0.23 }\\
		\bottomrule
	\end{tabular}
}
	\caption{Activation-based alignment (MNIST, MLPNet) \textbf{with Sinkhorn and regularization $=0.1$}: Results showing the performance (i.e., test classification accuracy ) of the averaged and aligned models of OT based averaging in contrast to vanilla averaging of weights as well as the prediction based ensembling. $m$ denotes the number of samples over which activations are computed, i.e., the mini-batch size. }
	\label{tab:act-mnist-highreg}
\end{table}

\subsection{Exact vs regularized variant}
In Table~\ref{tab:exact-ot}, we contrast the results obtained when no regularization is used and exact optimal transport is considered. Since using the exact optimal transport is fast enough, we default to using it hereafter. 
\begin{table}[h!] \centering\ra{1.2}
	\centering
	\resizebox{0.7\textwidth}{!}{
		\setlength{\tabcolsep}{5pt}
		\begin{tabular}{@{}cc|cc|c|cc@{}}
			\toprule
			\multirow{2}{*}{\textit{\ma}} & \multirow{2}{*}{\textsc{\mb}} & \multirow{1}{*}{\textsc{Prediction}} & \multirow{1}{*}{\textsc{Vanilla}} & 			\multirow{1}{*}{\textsc{Alignment}}  &\multirow{1}{*}{\textsc{OT avg. }} & \multirow{1}{*}{\textsc{\ma $\, \,$ aligned}}\\
			&		&  	\textsc{avg.}		& 		\textsc{avg.}	& \textsc{Type}	&   \multicolumn{2}{c}{Accuracy (mean)} \\
			\midrule
			\multicolumn{4}{c|}{\footnotesize \textit{Regularized OT (via Sinkhorn)}}  & Activation & 95.78  & 97.01  \\
			\cmidrule{1-4}
			\multirow{1}{*}{97.72} &\multirow{1}{*}{ 97.75} & \multirow{1}{*}{97.78} & \multirow{1}{*}{73.84} & Weight & 95.66  & 96.32 \\
			\midrule
			\multicolumn{4}{c|}{\footnotesize \textit{Exact OT}} &Activation & 96.21 & 97.72\\
			\cmidrule{1-4}
			\multirow{1}{*}{97.72} &\multirow{1}{*}{ 97.75} & \multirow{1}{*}{97.78} & \multirow{1}{*}{73.84} & Weight &  96.63 & 97.72 \\
			\bottomrule
	\end{tabular}}
	\vspace{-1mm}
	\caption{\textbf{Exact vs Regularized OT: } Results showing the performance gain with exact OT 
		for activation/weight based alignment. Here, regularization $\lambda=0.05$.}
	\label{tab:exact-ot}
\end{table}

\subsection{Layer-wise Optimal Transport distances}\label{sec:app-layerwise-analysis}

\begin{figure}[h]
	\centering
	
	\includegraphics[width=0.7\textwidth]{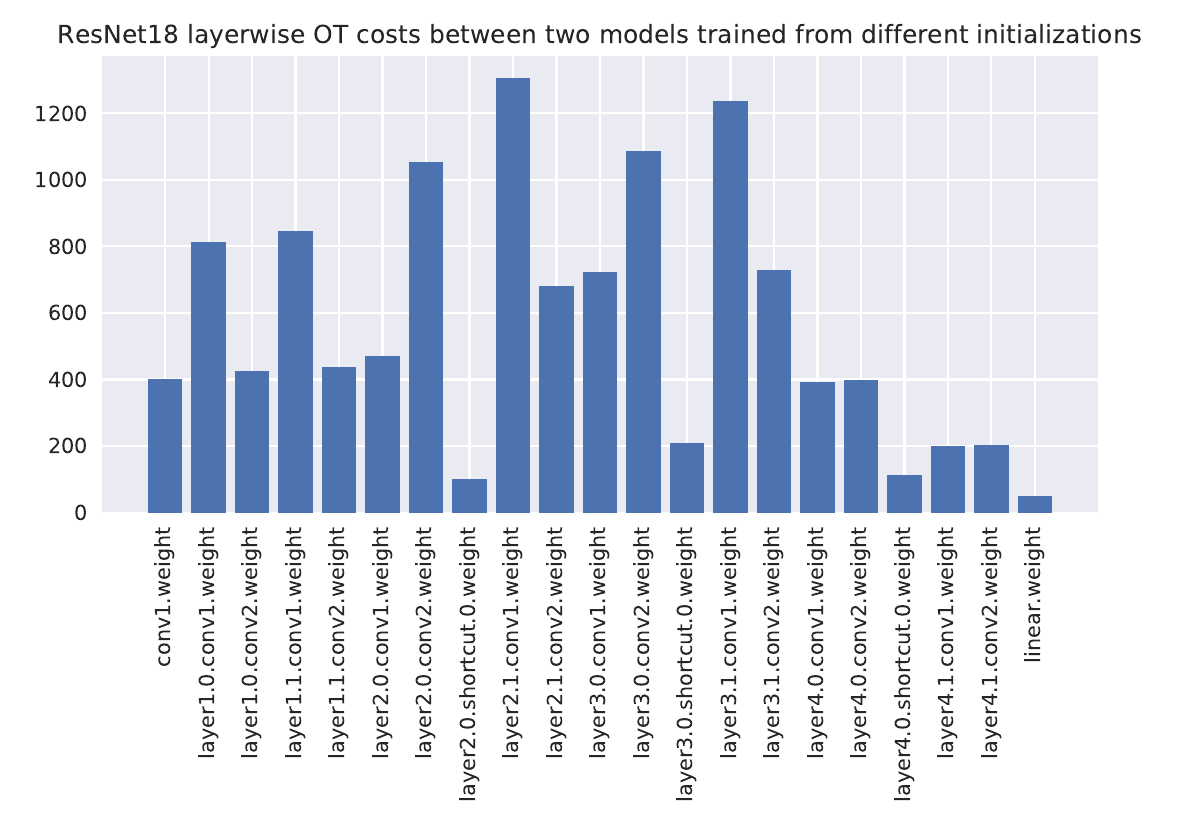}
	\caption{Illustrates the layerwise Optimal Transport costs between the corresponding layers of two ResNet18 models trained from different initializations, when using activation-based alignment with mini-batch size $m=200$. 
	}\label{fig:ot-costs-layerwise}
	
\end{figure}
A possible application of our model fusion approach can be for inspecting the similarity of representations at various layers across different neural networks. Thus, it could provide an alternative perspective for this problem of understanding the similarity of representations, besides the canonical correlation analysis (CCA) based methods used in the past \citep{morcos2018insights}. Figure~\ref{fig:ot-costs-layerwise} gives an example of this for two ResNet18 models trained from different initializations. Here, we used activation-based alignment with mini-batch size $m=200$. An extensive study, however, remains beyond the scope of this paper. 
\clearpage
\section{Detailed finetuning results}\label{sec:app-retrain-all}

In Tables \ref{tab:act_retrain_mnist_exact_detailed}, \ref{tab:act_retrain_cifar_detailed}, and \ref{tab:act_retrain_resnet_detailed}, we report the results of finetuning (i.e. retraining) the averaged models for (\mnist, \mlp) and (\cifar, \vgg).  For comparison, we also show the performance of individual models when further finetuned in this setting. Although in general,\textbf{ the individual model finetuning is not realistic}, since it is not known which one will lead to an improvement and this incurs $\#$ models $\times$ the finetuning cost.

\subsection{Two model scenario}

\subsubsection{For MNIST + \textsc{MLPNet}}\label{sec:app-exact-retrain}

The finetuning is carried out for 60 epochs at the following set of constant learning rates $\{0.01, 0.002, 0.001, 	0.00067, 0.0005\}$. Note that the original models were trained for 10 epochs at a learning rate of $0.01$. For OT average, we use the activation-based alignment with mini-batch size $m=200$. 

Table \ref{tab:act_retrain_mnist_exact_detailed} shows the results for each method at their best respective finetuning runs.  
\begin{table}[h] \centering\ra{1.2}
	\centering
	\resizebox{0.7\textwidth}{!}{
		\begin{tabular}{@{}c|cc|cc@{}}
			\toprule
			\multirow{1}{*}{\textsc{Finetuning LR}} & \multirow{1}{*}{\textsc{Model A}} & \multirow{1}{*}{\textsc{Model B }} &  \multirow{1}{*}{\textsc{Vanilla avg.} } & \multicolumn{1}{c}{\textsc{OT avg. (exact)}} \\
			\midrule
			\multicolumn{5}{l}{\textit{Baseline Results}} \\
			\midrule
			--- &97.72 &	\cellcolor{orange!40}{97.75} &	73.84 & 96.54	\\
			\midrule
			\multicolumn{5}{l}{\textit{Results for the \textbf{best} finetuning run} (reported at the best checkpoint)} \\
			\midrule
			0.01 & 98.21 & 98.13 & 98.23 & \cellcolor{orange!40}{\textbf{98.35}} \\
			0.002 & 98.13 & 98.03 & 98.13 & \textbf{98.21 }\\
			0.001 & 98.09 & 98.03 & 97.98 & \textbf{98.14} \\
			0.00067 & \textbf{98.11} & 98.00 & 97.83 & 98.07 \\
			0.0005 & \textbf{98.09} & 98.01 & 97.70 & 98.05 \\
			\bottomrule
		\end{tabular}
	}
	\caption{\textbf{Effect of finetuning the individual and averaged models for (MNIST, MLPNet):}  Best finetuning runs have been reported for each method. Cells in orange highlight the best scores in each regime.}
	\label{tab:act_retrain_mnist_exact_detailed}
\end{table}

We also show in Table \ref{tab:act_retrain_mnist_exact_detailed_averaged} the results when averaged across 5 finetuning runs for each of the finetuning LR, as the cost of finetuning here is not as prohibitive in comparison to  finetuning VGG11 and ResNet18 models. We see that performance trend remains in accordance with the previous Table~\ref{tab:act_retrain_mnist_exact_detailed}.
\begin{table}[h] \centering\ra{1.3}
	\centering
	\resizebox{0.7\textwidth}{!}{
		\begin{tabular}{@{}c|cc|cc@{}}
			\toprule
			\multirow{1}{*}{\textsc{Finetuning LR}} & \multirow{1}{*}{\textsc{Model A}} & \multirow{1}{*}{\textsc{Model B }} &  \multirow{1}{*}{\textsc{Vanilla avg.} } & \multicolumn{1}{c}{\textsc{OT avg. (exact)}} \\
			\midrule
			\multicolumn{5}{l}{\textit{Baseline Results}} \\
			\midrule
			--- &97.72 &	\cellcolor{orange!40}{97.75} &	73.84 & 96.21 $\pm$ 0.36	\\
			\midrule
			\multicolumn{5}{l}{\textit{\textbf{Averaged} results across the finetuning runs} (reported at the best checkpoint)} \\
			\midrule
			0.01 & 98.19 $\pm$ 0.02 & 98.11 $\pm$ 0.02 & 98.22 $\pm$ 0.02 & \cellcolor{orange!40}{\textbf{98.28 $\pm$ 0.05}} \\
			0.002 & 98.13 $\pm$ 0.01 & 98.03 $\pm$ 0.01 & 98.13 $\pm$ 0.01 & \textbf{98.15 $\pm$ 0.07}\\
			0.001 & \textbf{98.11 $\pm$ 0.02 } & 98.01 $\pm$  0.01 & 97.99 $\pm$ 0.01 & 98.08 $\pm$ 0.05  \\
			0.00067 & \textbf{98.11 $\pm$ 0.02}& 98.00 $\pm$ 0.01 & 97.83 $\pm$ 0.02 & 98.05 $\pm$ 0.04 \\
			0.0005 & \textbf{98.09 $\pm$ 0.01} & 98.01 $\pm$ 0.00 & 97.68 $\pm$ 0.01 & 98.03 $\pm$ 0.03 \\
			\bottomrule
		\end{tabular}
	}
	\caption{\textbf{Effect of finetuning the individual and averaged models for (MNIST, MLPNet):}  Average of the results across 5 finetuning runs as well as their standard deviation are reported for each method. Cells in orange highlight the best scores in each regime. }
	\label{tab:act_retrain_mnist_exact_detailed_averaged}
\end{table}
\clearpage

\subsubsection{For \cifar $\,$+ \vgg}\label{sec:cifar_vgg_app}

 As a recall, the original models were trained for 300 epochs at an initial learning rate of $0.05$, which was decayed by a factor of 2 after every 30 epochs. The finetuning is carried out for 100 epochs at the following set of initial learning rates $\{0.01, 0.05, 0.0033, 0.0025\}$. Also, similar to training, the learning rate is decayed in the finetuning process. Note that, here finetuning at the initial learning rate of $0.01$ causes model B to diverge and hence we skip the results for this setting. 

For OT average, we use the weight-based alignment.  Table \ref{tab:act_retrain_cifar_detailed} shows the best results for each method during their finetuning run.

\begin{table}[h] \centering\ra{1.3}
	\centering
	\resizebox{0.7\textwidth}{!}{
		\begin{tabular}{@{}c|cc|cc@{}}
			\toprule
			\multirow{1}{*}{\textsc{Finetuning LR}} & \multirow{1}{*}{\textsc{Model A}} & \multirow{1}{*}{\textsc{Model B }} &  \multirow{1}{*}{\textsc{Vanilla avg.} } & \multicolumn{1}{c}{\textsc{OT avg. (exact)}} \\
			
			\midrule
			\multicolumn{5}{l}{\textit{Baseline Results}} \\
			\midrule
			--- &	\multirow{1}{*}{90.31} & \cellcolor{orange!40}{\multirow{1}{*}{90.50}} & \multirow{1}{*}{17.02}  &  85.98\\
			\midrule
			\multicolumn{5}{l}{\textit{Results after finetuning} (reported scores are at best checkpoint)} \\
			\midrule
			0.01 & 90.29 & 90.53 & 90.39 & \cellcolor{orange!40}{\textbf{90.73}} \\
			0.005 & 90.36 & 90.47 & 90.16 & \textbf{90.64} \\
			0.0033 & 90.28 & 90.39 & 90.13 & \textbf{90.39} \\
			0.0025 & 90.45 & \textbf{90.50} & 89.88 & 90.30 \\
			\bottomrule
		\end{tabular}
	}
	\caption{\textbf{Effect of finetuning the individual and averaged models  for (CIFAR10, \textsc{VGG11}):}  Model A $\&$ Model B baseline accuracies correspond to best checkpoints when originally trained for 300 epochs. Cells in orange highlight the best scores in each regime.
	}
	\label{tab:act_retrain_cifar_detailed}
\end{table}

\subsubsection{For \cifar $\,$+ \resnet}

As a recall, the original models were trained for 300 epochs at an initial learning rate of $0.1$, which was decayed by a factor of 10 at the epochs $\{150, 250\}$. The finetuning is carried out for 120 epochs at the following set of initial learning rates $\{0.1, 0.04, 0.02\}$. For OT average, we use the activation-based alignment, with mini-batch size $m=200$.

\begin{table}[h] \centering\ra{1.3}
	\centering
	\resizebox{0.7\textwidth}{!}{
		\begin{tabular}{@{}c|cc|cc@{}}
			\toprule
\textsc{Finetuning LR} & \textsc{Model A} & \textsc{Model B } &  \textsc{Vanilla avg.}  & \textsc{OT avg. (exact)} \\
			\midrule
			\multicolumn{5}{l}{\textit{Baseline Results}} \\
			\midrule 
			--- &	\multirow{1}{*}{93.11} & \cellcolor{orange!40}{\multirow{1}{*}{93.20}} & \multirow{1}{*}{18.49}  &  67.46\\
			\midrule
			\multicolumn{5}{l}{\textit{Results after finetuning }(reported at the best checkpoint)}\\
			\midrule
			\multicolumn{5}{l}{\footnotesize(a) \textit{LR decay epochs$\,=[20, 40, 60, 80, 100]$}} \\
 			 \multicolumn{5}{l}{\makebox[0.6\textwidth]{\dashrule[black]}} \\
			0.1 & 93.51 & 93.43 & 93.29 & \cellcolor{orange!40}{\textbf{93.78}} \\
			0.04 & 93.35 & 93.34 & 93.28 & \textbf{93.35} \\
			0.02 & 93.28 & \textbf{93.28} & 93.09 & 92.97 \\
			\midrule
			\multicolumn{5}{l}{\footnotesize(b) \textit{LR decay epochs$ \,=[40, 80]$}}  \\
 			 \multicolumn{5}{l}{\makebox[0.6\textwidth]{\dashrule[black]}} \\
			0.1 & 93.49 & 93.32 & 93.34 & \cellcolor{orange!40}{\textbf{93.59}} \\
			0.04 & 93.27 & 93.34 & \textbf{93.49} & 93.38 \\
			0.02 & 93.21 & \textbf{93.33} & 93.17 & 93.15 \\
			\bottomrule
		\end{tabular}
	}
	\caption{\textbf{Effect of finetuning the individual and averaged models  for (CIFAR10, \textsc{ResNet18}):}  Model A and Model B baseline accuracies correspond to best checkpoints when originally trained for 300 epochs. Cells in orange highlight the best scores in each regime.
	}
	\label{tab:act_retrain_resnet_detailed}
\end{table}

Table \ref{tab:act_retrain_resnet_detailed} shows the best results for each method during their finetuning run. The learning rate is decayed by a factor of 2 in the finetuning process as per two schedules: (a) after every 20 epochs, and (b) after every 40 epochs. These are indicated in the respective sections of  the Table \ref{tab:act_retrain_resnet_detailed}.

\clearpage
\subsection{Multiple model scenario: CIFAR10}\label{app:cifar_mult_opt_try}
Now, we discuss in detail, the experiments performed for the multiple model setting on \textsc{CIFAR10}. Namely, when we have $4$ and $6$ \textsc{VGG11} models, that have different initializations, but are trained identically on the entire data, as mentioned in Table~\ref{tab:act_retrain_cifar_mult}. 

\begin{table}[h!] \centering\ra{1.2}
	\centering
	\vspace{-0.3em}
	\resizebox{\textwidth}{!}{
		\begin{tabular}{@{}c|c|ccc|cc@{}}
			\toprule
			\multirow{1}{*}{\textit{\cifar +}} & \multirow{2}{*}{\textsc{Individual Models}}  & \multirow{1}{*}{\textsc{Prediction}} & \multirow{1}{*}{\textsc{Vanilla} } & \multirow{1}{*}{\textsc{OT}} &  \multicolumn{2}{c}{\textsc{Finetuning}}\\
			
			\multirow{1}{*}{\textsc{\vgg }} &   & \multirow{1}{*}{\textsc{avg.}} & \multirow{1}{*}{\textsc{avg.} } & \multirow{1}{*}{\textsc{avg. }} &  \multirow{1}{*}{\textsc{Vanilla}} & \multirow{1}{*}{\textsc{OT}}\\
			\midrule
			Accuracy  &	\multirow{1}{*}{[90.31, 90.50, 90.43, 90.51]}  & \multirow{1}{*}{91.77} & \multirow{1}{*}{10.00}  &  \multirow{1}{*}{73.31} & \multirow{1}{*}{12.40} & \multirow{1}{*}{{90.91}} \\
			
			Efficiency &	1 $\times$& 1 $\times$ & 4 $\times$ & 4 $\times$  & 4 $\times$  & 4 $\times$    \\
			\midrule
			Accuracy  &	\multirow{1}{*}{[90.31, 90.50, 90.43, 90.51, 90.49, 90.40]} & \multirow{1}{*}{91.85}  & \multirow{1}{*}{10.00}  &  \multirow{1}{*}{72.16} &  \multirow{1}{*}{11.01} & \multirow{1}{*}{{91.06}} \\
			Efficiency &	1 $\times$& 1 $\times$ & 6 $\times$ & 6 $\times$  & 6 $\times$  & 6 $\times$    \\
			
			\bottomrule
		\end{tabular}
	}
	\caption{Results of our OT average + finetuning based efficient alternative for ensembling in contrast to vanilla average + finetuning, for more than two input models (\vgg) with different initializations trained on \textsc{CIFAR10}. 
	}
	\label{tab:act_retrain_cifar_mult}
\end{table}

We consider finetuning the averaged models, with many different optimization hyperparameters, however vanilla average fails to finetune or retrain. In particular, we finetune for 150 epochs with learning rate obtained by dividing the original learning rate (with which models were trained) by factors of $\{1, 2, 4, 8, 16\}$ (called `initial decay'). Further, similar to learning rate schedule followed in the training, we try decaying the learning rate by a factor of $\{1.1, 1.5, 2.0\}$ after every 20 epochs. We also tried adjusting the interval after which the learning rate was decayed (like 40 epochs), but this was again to no avail in being able to finetune the vanilla average. So for simplicity, in the rest of discussion, we consider that the interval after which the learning rate gets decayed is 20 epochs.

Across all the settings OT average is able to successfully retrain, except when the learning rate is set to the original learning rate of $0.05$, with which models were trained (i.e., initial decay of $1$). This is to be expected as the OT average without retraining itself already performs fairly well, and setting such a high learning rate is bound to cause this. In contrast, vanilla average fails to retrain at all, with the best accuracy of $12.40$ and $11.01$ for the case of 4 and 6 models, when the initial decay is $1$, and the learning rate decay is $1.1$.

Finetuning from OT average results, in a significant improvement for numerous settings of the above hyperparameters, and below, we show the top 5 such settings in Table~\ref{tab:fourmodel} for both 4 and 6 models. (For OT average, we use the activation-based alignment.)

\begin{table}[h] \centering\ra{1.2}
	\centering
	\resizebox{\textwidth}{!}{
		\begin{tabular}{@{}ccccc@{}}
			\toprule
			\multirow{2}{*}{\textsc{Initial decay factor}} & \multirow{2}{*}{\textsc{Scheduled LR decay factor}} &\multirow{2}{*}{\textsc{Decay interval}} &   \multicolumn{2}{c}{\textsc{Finetuning}} \\
			\cmidrule{4-5}
			&  & &  \multirow{1}{*}{\textsc{Vanilla avg.} } & \multicolumn{1}{c}{\textsc{OT avg.}} \\
			
			\midrule
			\multicolumn{5}{l}{\textit{(i) Number of models $= 4$}} \\
			\midrule
			2 & 2.0 & 20& 10.34 & 90.91 \\
			4 & 2.0 & 20 & 10.32 & 90.80 \\
			2 & 2.0 & 40 & 10.34 & 90.74 \\
			2 & 1.5 & 20 & 10.34 & 90.67 \\
			4 & 2.0 & 40& 10.32 & 90.66 \\

			\midrule
			\multicolumn{5}{l}{\textit{(ii) Number of models $= 6$}} \\
			\midrule
			2 & 2.0 & 20& 10.00 & 91.06 \\
			2 & 1.5 & 20& 10.00 & 90.97 \\
			4 & 2.0 & 20& 10.00 & 90.88 \\
			4 & 2.0 & 40& 10.00 & 90.81 \\ 
			8 & 2.0 & 40& 10.00 & 90.69 \\ 
			
			\bottomrule
		\end{tabular}
	}
	\caption{Different finetuning settings which show how OT fusion can improve over the individual models after finetuning, while the vanilla average fails to do so. As a result, we obtain one single improved model that can be used as an efficient replacement for the ensemble.}
	\label{tab:fourmodel}
\end{table}

\clearpage
\section{Finetuning curves}\label{sec:app-retrain-curve}

\begin{figure}[h]
	\centering
	
	\includegraphics[width=0.6\textwidth]{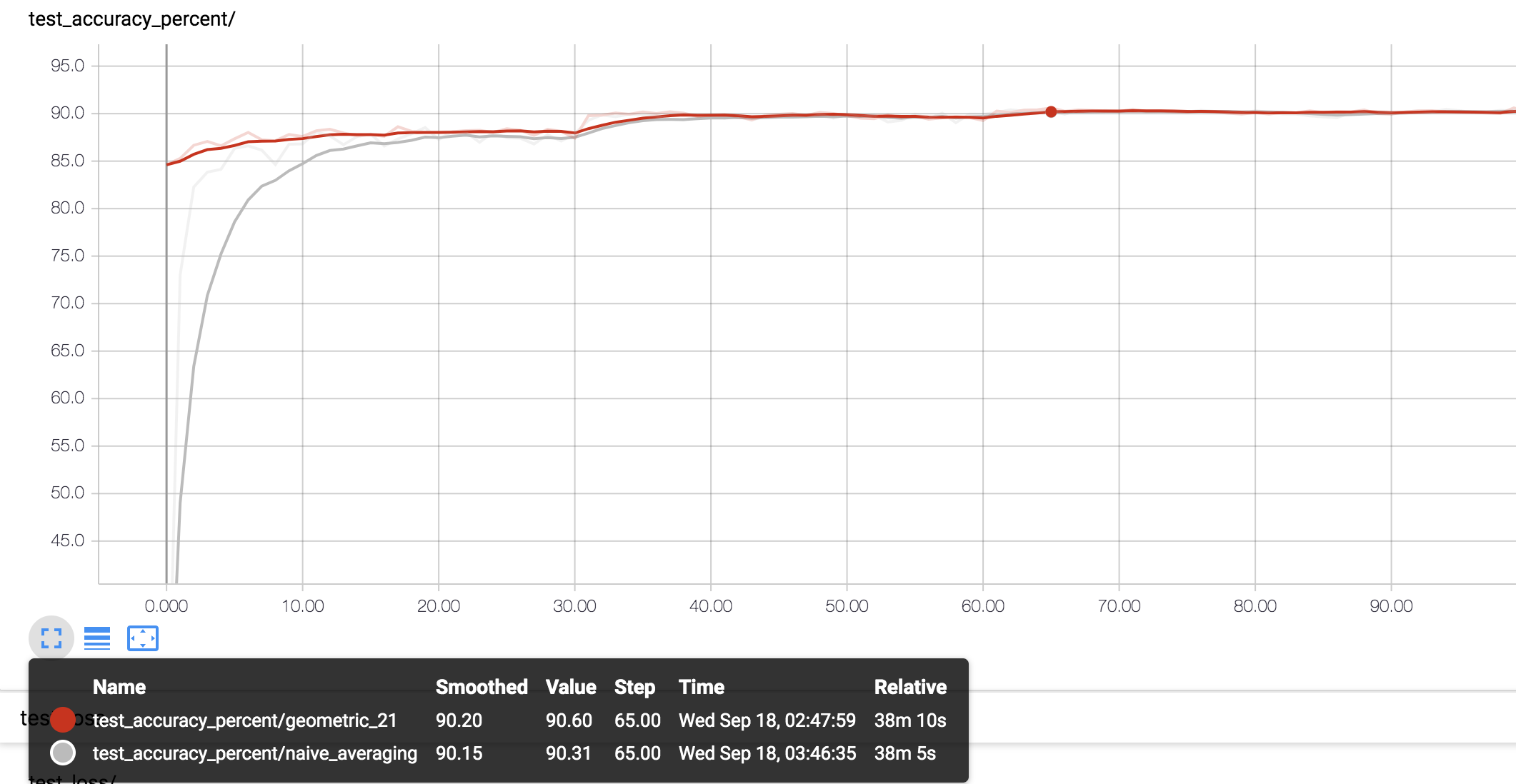}
	\caption{Illustrates the performance of OT averaging (referred to as geometric in the figure legend) and vanilla averaging during the process of retraining for \cifar  $\,$ with \vgg.
	}\label{fig:retrain-cifar}
	
\end{figure}

\begin{figure}[h]
	\centering
	
	\includegraphics[width=0.6\textwidth]{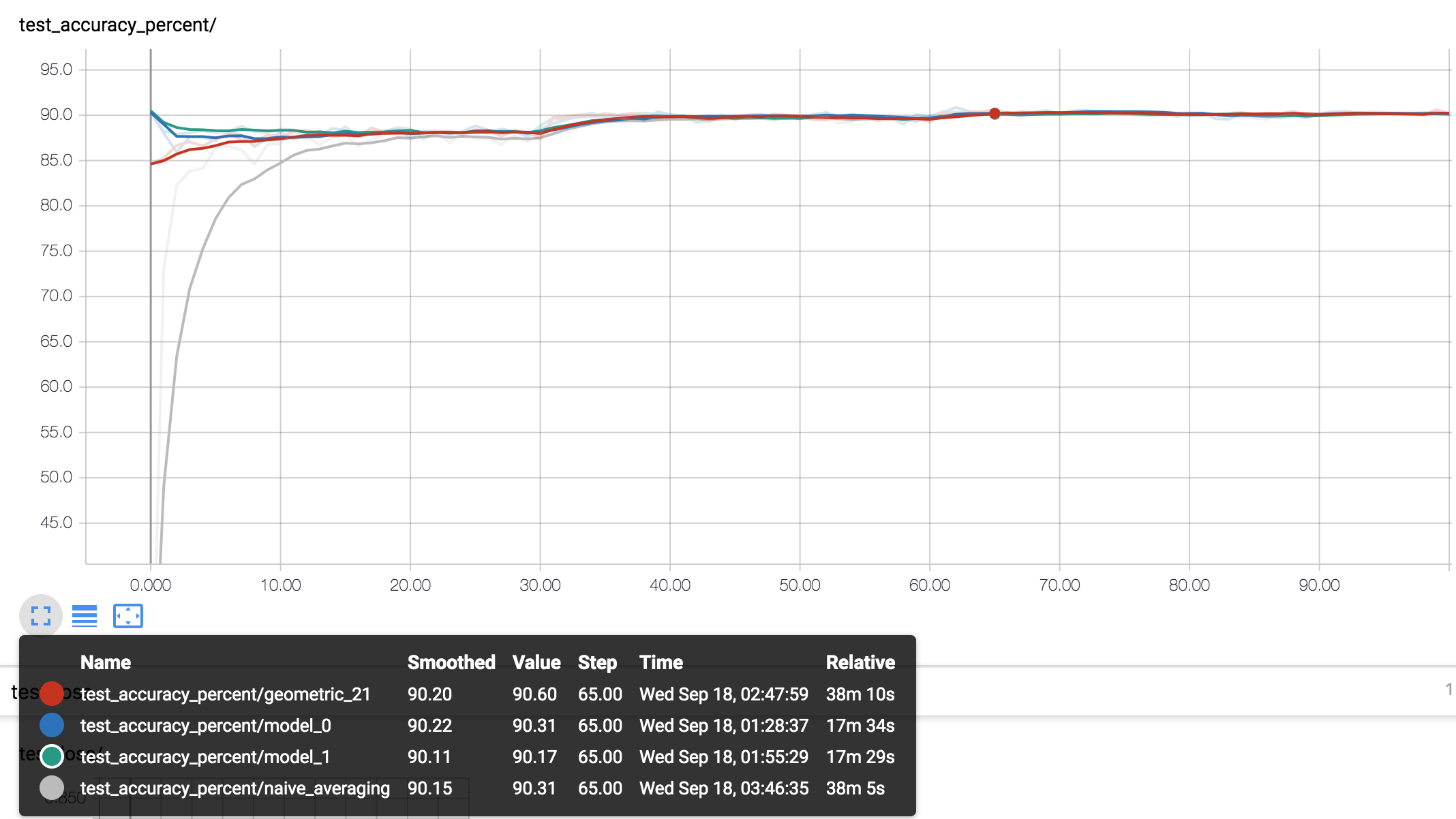}
	\caption{Retraining with reference plots of individual models. Other than that same as above. 
	}\label{fig:retrain-cifar-indv}
	
\end{figure}

\clearpage
\section{Skill Transfer: Additional Results}\label{sec:app-skill}
\subsection{Remaining Data Split: $10\%$}
\begin{figure}[h]
	
	\centering    
	\subfigure[Special digit 4, same init avg]{\label{fig:b}\includegraphics[width=0.4\textwidth]{figures/final_3_personalized/auto_personalized_4_frac_0.1_same_init.pdf}}
	\subfigure[Special  6, same init avg]{\label{fig:b}\includegraphics[width=0.4\textwidth]{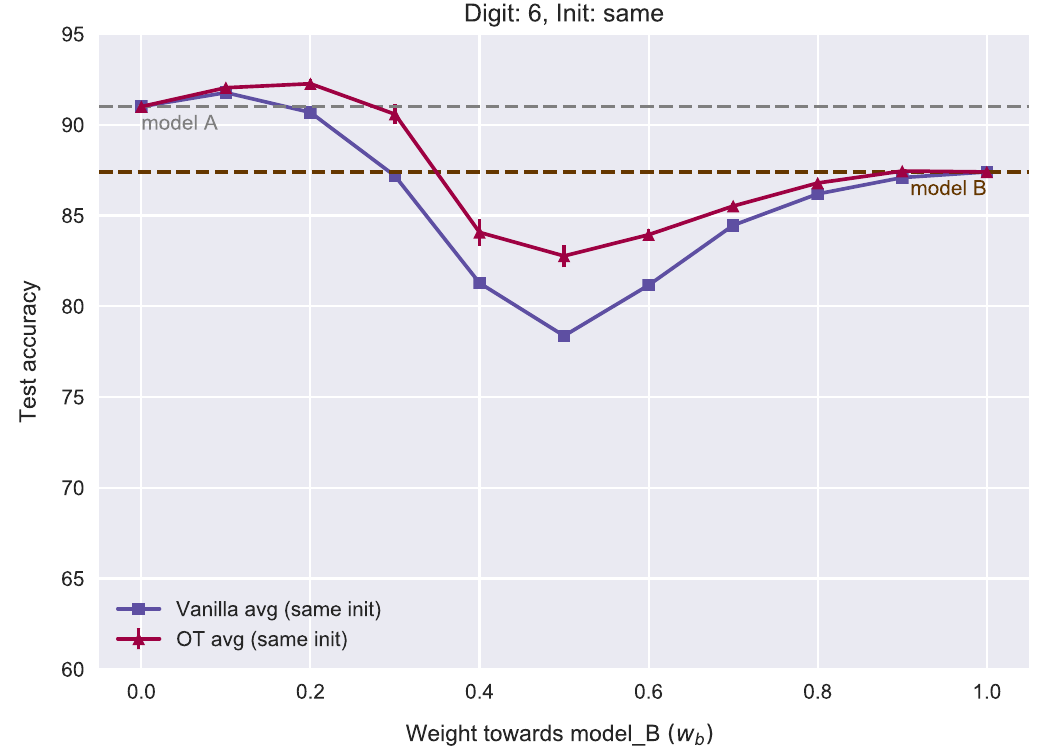}}
	\subfigure[Special digit 6, different init avg]{\label{fig:b}\includegraphics[width=0.4\textwidth]{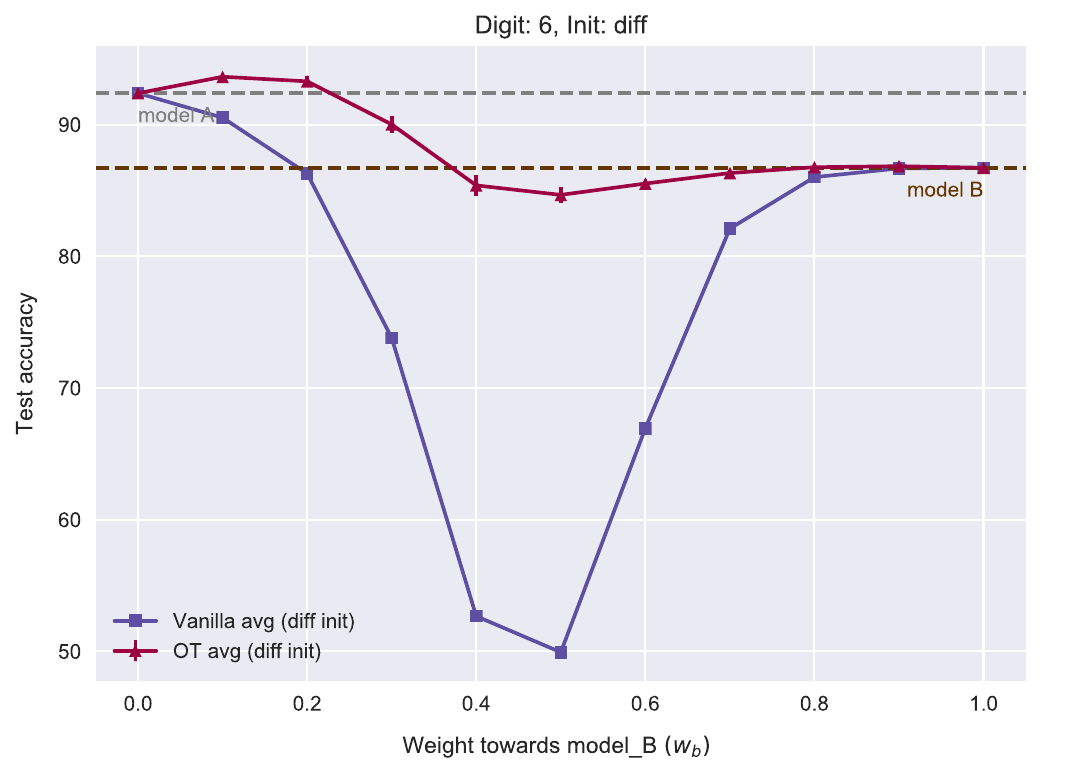}}
	\caption{\textbf{Skill Transfer performance}: Comparison results of OT based model fusion (OT avg) with vanilla averaging for different $w_B$. Each point for OT avg. curve (magenta colored) is obtained by activation-based alignment with a batch size $m=400$, and we plot the mean performance over 5 seeds along with the error bars, which show the corresponding standard deviation. Here the remaining data besides the special digit, is split as $90\%$ for model B and the other $10\%$ for model A.}
	\label{fig:skill-transfer-more1}

\end{figure}
\clearpage
\subsection{Remaining Data Split: $5\%$}
\begin{figure}[h]
	
	\centering    
	\subfigure[Special digit 4, same init avg]{\label{fig:b}\includegraphics[width=0.4\textwidth]{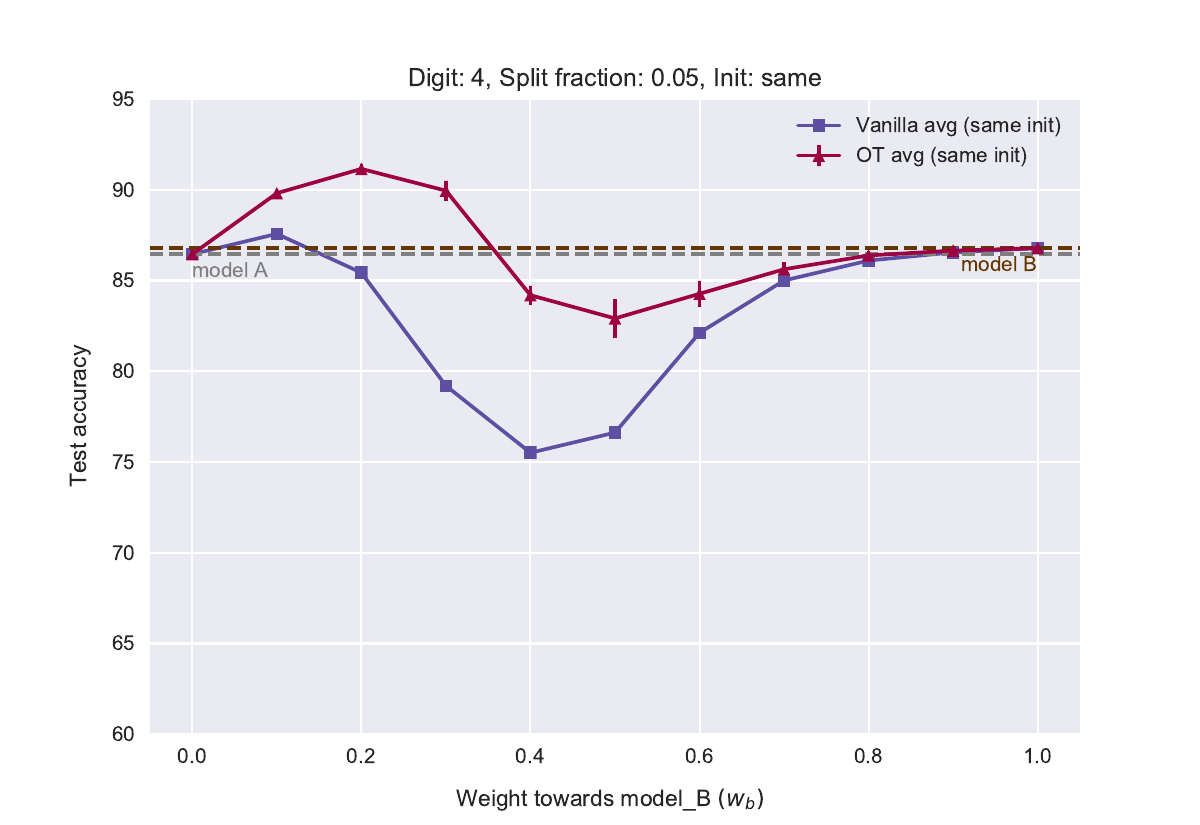}}
	\subfigure[Special digit  4, different init avg]{\label{fig:b}\includegraphics[width=0.4\textwidth]{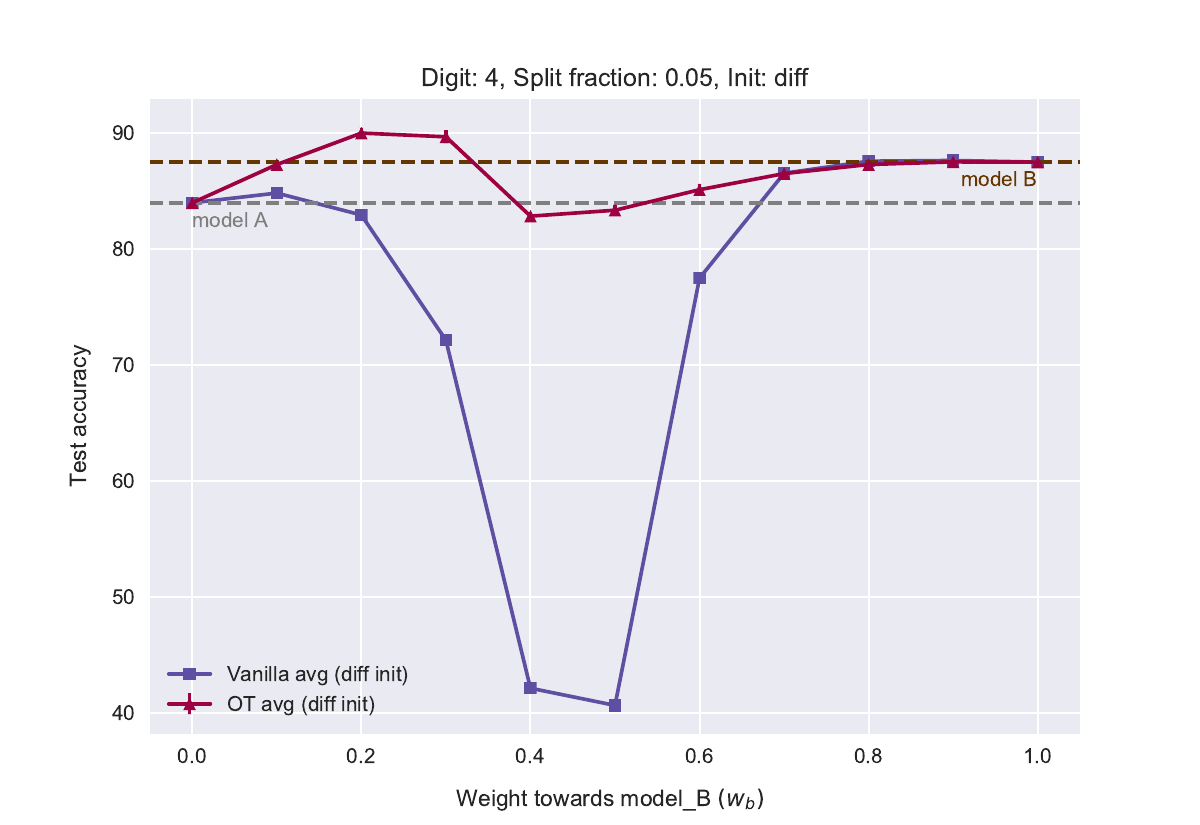}}
	\subfigure[Special digit  6, same init avg]{\label{fig:b}\includegraphics[width=0.4\textwidth]{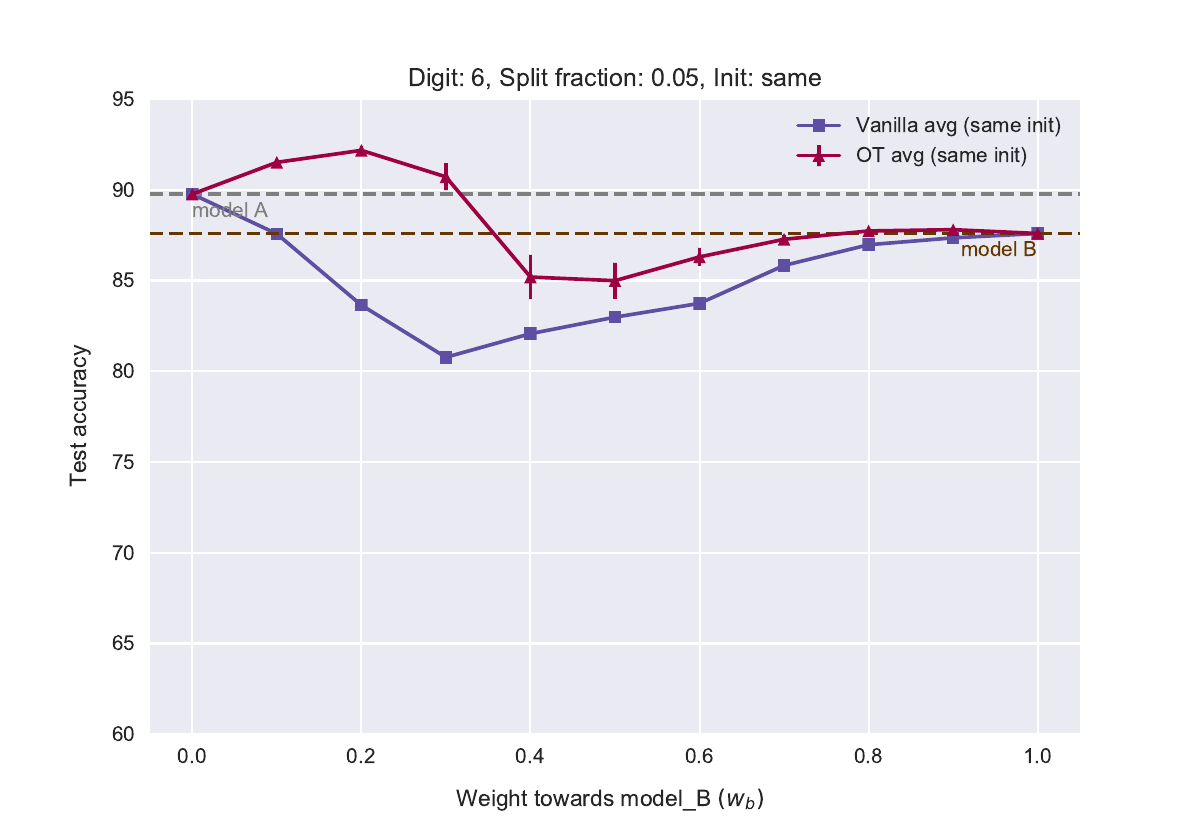}}
	\subfigure[Special digit  6, different init avg]{\label{fig:b}\includegraphics[width=0.4\textwidth]{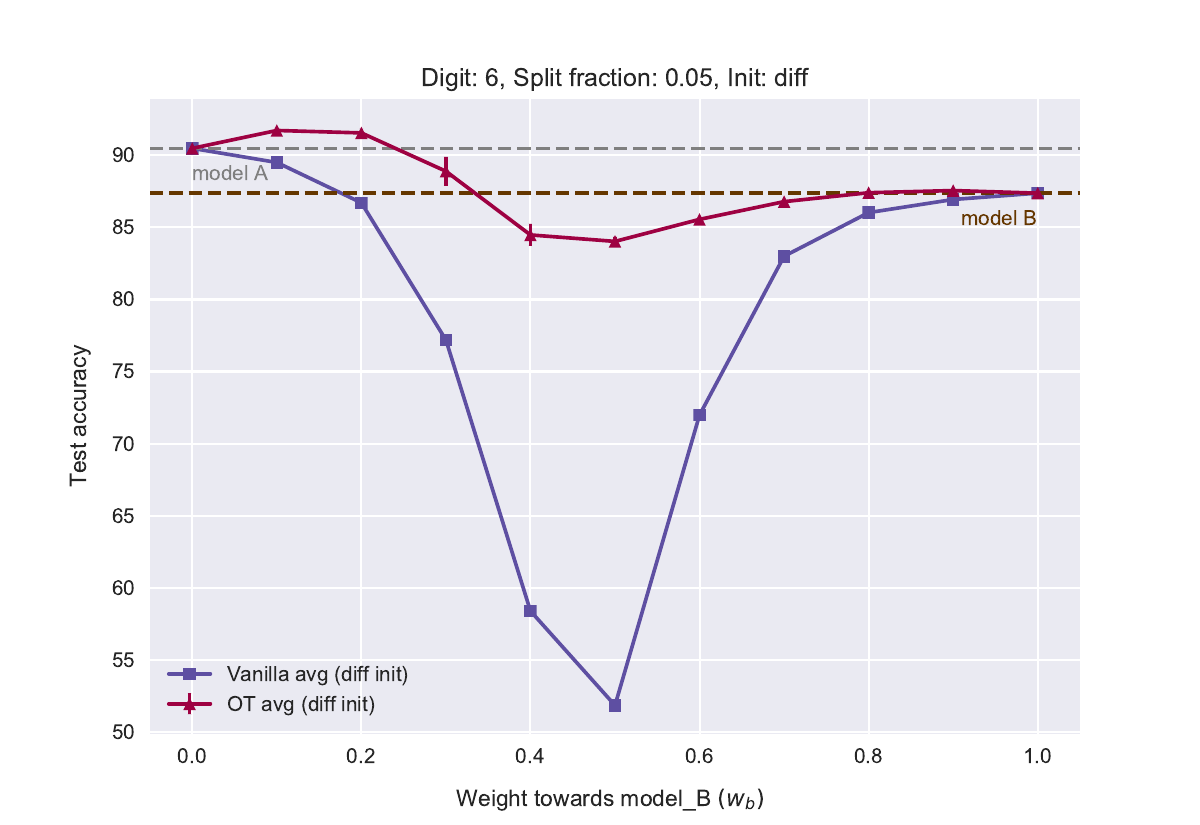}}
	\caption{\textbf{Skill Transfer performance}: Comparison results of OT based model fusion (OT avg) with vanilla averaging for different $w_B$. Each point for OT avg. curve (magenta colored) is obtained by activation-based alignment with a batch size $m=400$, and we plot the mean performance over 5 seeds along with the error bars, which show the corresponding standard deviation. Here the remaining data besides the special digit, is split as $95\%$ for model B and the other $5\%$ for model A.}
	\label{fig:skill-transfer-more2}

\end{figure}
\clearpage
\subsection{Scenarios without specialized labels}
 Even if we don't exclude a digit and just alter the fraction of data between A and B, results are similar. E.g., take \textsc{MLPNets} A and B with \textit{same} initialization (\textit{to help vanilla averaging}), but A has $30\%$ and B has $70\%$ of the data. This results in (global) test accuracy $\%$ of $94.2$ and $95.0$ for A and B resp. OT fusion is better than vanilla averaging when combining A and B for all proportions, with best results as, OT: mean $\textbf{95.3}$ (stdev=0.1), vanilla avg: $95.1$ at proportions $0.1, 0.9$ respectively. Ensembling is better than both ($95.5$), but requires 2x more memory and inference time.  
 
 Likewise, for other data splits (such as $10\%$ vs $90\%$, $50\%$ vs $50\%$, etc), OT fusion outperforms the individual models as well as vanilla averaging. For, further settings, also see Section~\ref{sec:app-mult-skill}.

\section{Results for one-shot skill-transfer under size constraints}\label{app:skill_transfer_size}

\begin{figure}[h]
	
	\centering    
	\subfigure[Special digit 4, data split\% = 10 ]{\label{fig:b}\includegraphics[width=0.4\textwidth]{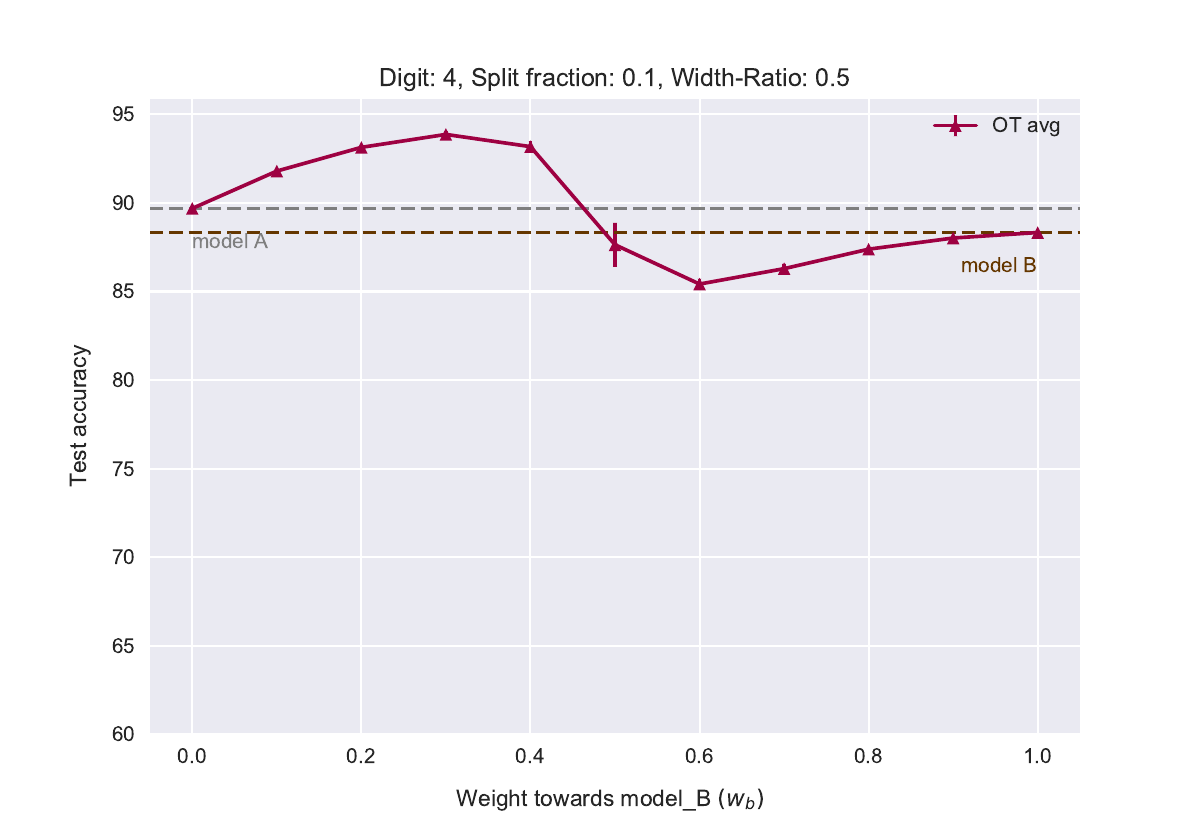}}
	\subfigure[Special digit  4, data split\% = 5 ]{\label{fig:b}\includegraphics[width=0.4\textwidth]{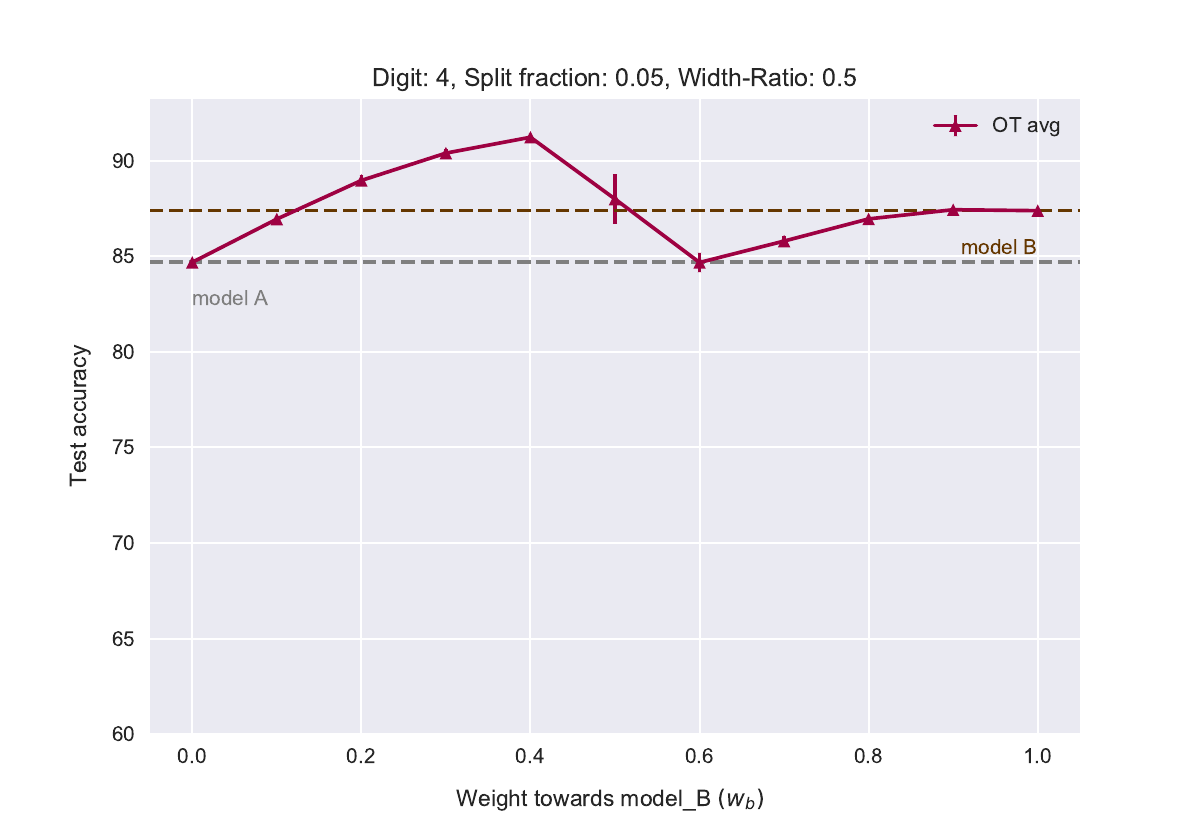}}
	\subfigure[Special digit  6, data split\% = 10 ]{\label{fig:b}\includegraphics[width=0.4\textwidth]{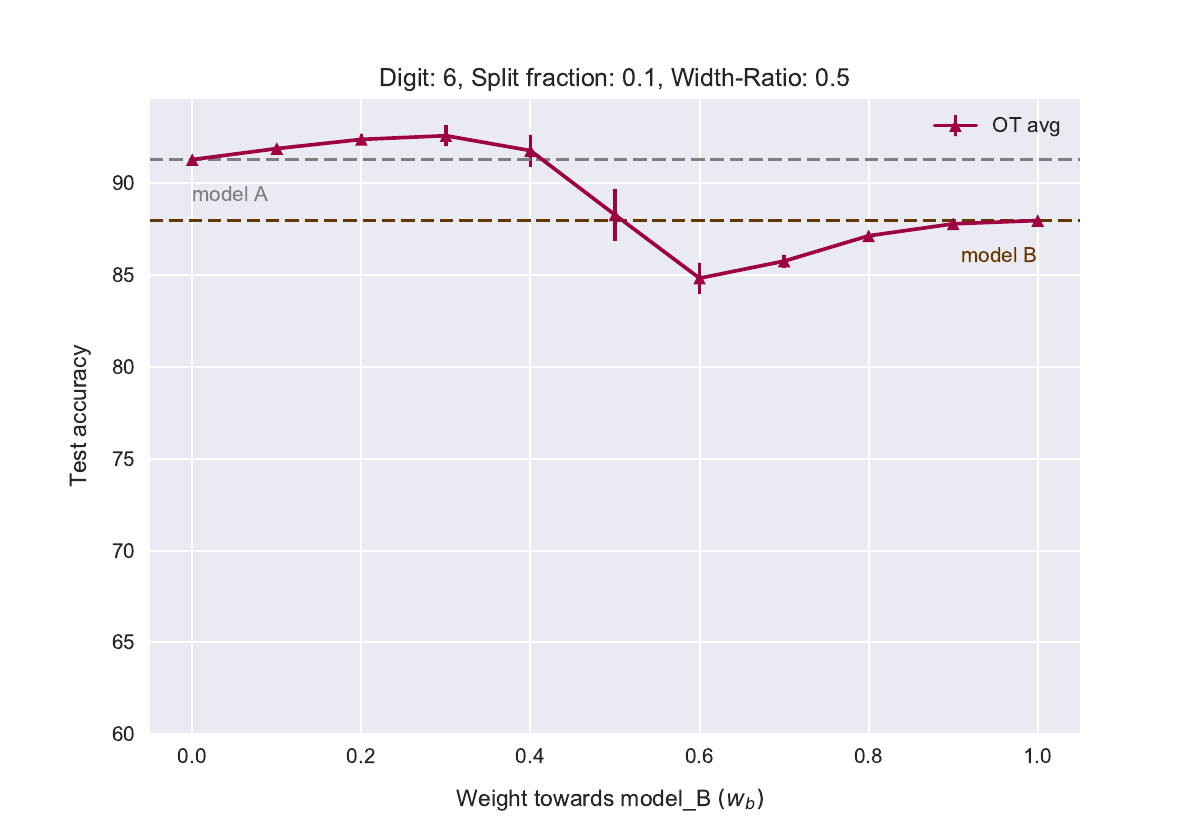}}
	\subfigure[Special digit  6, data split\% = 5 ]{\label{fig:b}\includegraphics[width=0.4\textwidth]{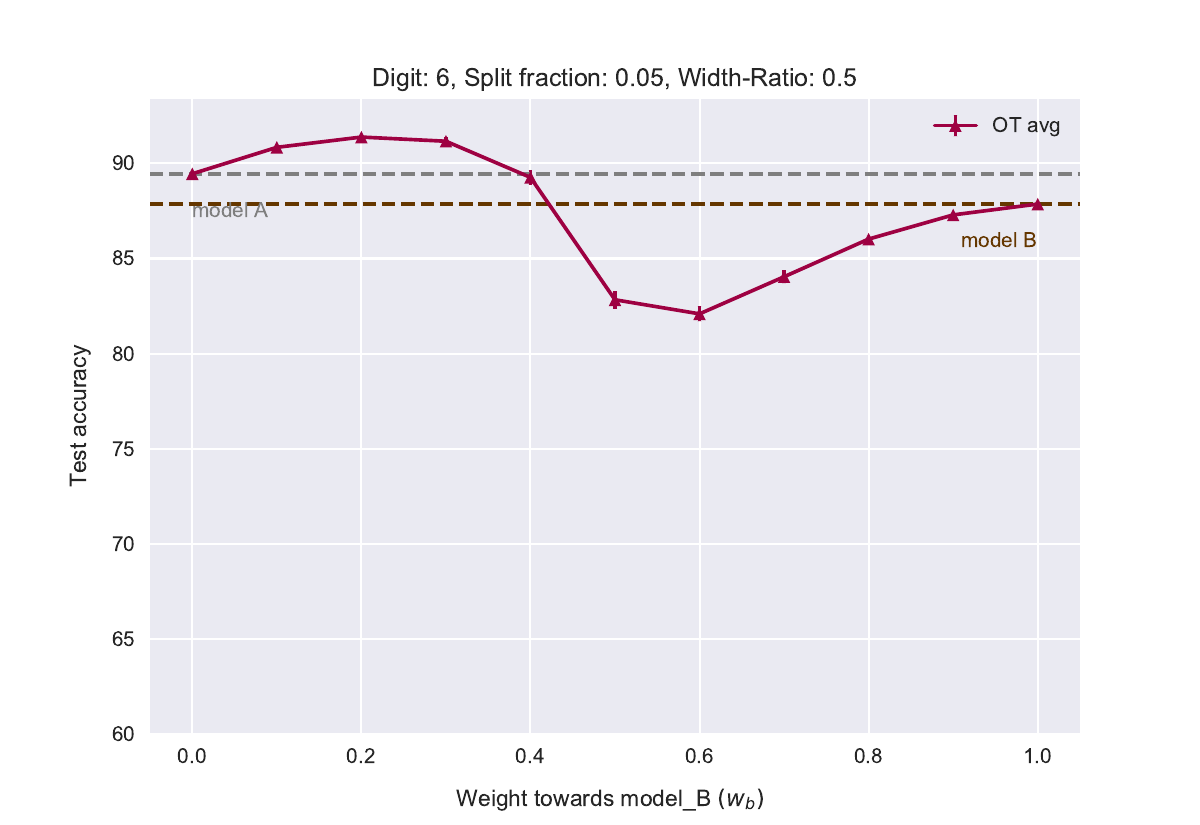}}
	\caption{\textbf{Skill Transfer performance for different sized models}: Results of OT-based model fusion (OT avg) for different $w_B$. Unlike the results in the previous section, vanilla averaging is not possible here as the models are of unequal sizes. `Width-Ratio 0.5' in the figure title denotes the ratio of the hidden layers sizes of model A and B. Each point for OT avg. curve (magenta colored) is obtained by activation-based alignment with a batch size $m=400$, and we plot the mean performance over 5 seeds along with the error bars, which show the corresponding standard deviation. The data split \% indicates the amount of remaining data besides the special digit which is present with model A. Model B contains 100 - data split\% of this remaining data.}
	\label{fig:skill-transfer-more-widthratio}
	
\end{figure}

Here, we present results for one-shot skill-transfer when the two models are of unequal sizes. More concretely, as an example, we consider that the hidden layers of the generalist model B are twice as wide as that of the specialist model A. Figure~\ref{fig:skill-transfer-more-widthratio} illustrates the results for OT-based model fusion (OT average) in such a setting. Note that, here vanilla averaging can not be applied as the models are of different sizes. To the best of our knowledge, we are unaware of any other method that can allow for such one-shot skill transfer (i.e., fuse the given different size models into a single model in one-shot). 

Rest of the technical details are identical as in the setup of Sections~\ref{sec:one-shot} in the main text and \ref{sec:app-skill} in the supplementary. 

\section{Multi-model one-shot skill transfer}\label{sec:app-mult-skill}

To recap, here we take four \textsc{MLPNet} models: A, B, C and D, with the same initialization and assume that A again possessing the knowledge of a special digit (say, 4). Consider that the rest of the data is divided as $10\%, 30\%, 50\%, 10\%$. 

Now they are trained in a similar setting for 10 epochs, by the end of which these models obtain (global) test accuracies of $87.7\%, 86.5\%, 87.0\%,83.5\%$ respectively. Since A is the only model which has seen the special digit `4', we assign it a larger proportion in the final fused model. In particular, we consider fusing the models in proportions of $0.7, 0.1, 0.2, 0.1$ respectively \textit{(later normalized to sum to 1).} Then, ensembling the predictions yields $95.0\%$ while vanilla averaging obtains $80.6\%$. In contrast, OT averaging results in $\textbf{93.6\%}$ test accuracy ($\approx6\%$ gain over the best individual model), while being $4\times$ more efficient than ensembling. 

This is also robust to many other proportions in which the models are combined. For example, decreasing the weight of model A so that the proportions are $0.6, 0.1, 0.2, 0.1$, gives: Prediction ensembling $95.03\%$, vanilla average $78.44\%$, OT average $\textbf{92.72\%}$. Or increasing the proportion of B and D, i.e., let the proportions be instead $0.7,0.15,0.2,0.15$. The results for such a case are as follows, Prediction ensembling $94.91\%$, vanilla average $76.14\%$, OT average $91.67\%$. Take another example, say we increase the proportion of model C now, so as to have the  proportions $0.7,0.1,0.3,0.1$. In this case, we get Prediction ensembling $95.15\%$, vanilla average $77.93\%$, OT average $\textbf{92.21}\%$. We can go on for many other examples, but the results remain similar. 

Overall, we find that OT average leads to a significant across all these examples, and outperforms vanilla average by a large margin. In comparison to prediction ensembling, it is slightly worse in terms of accuracy, but it enjoys $4\times$ efficiency, with respect to future usage and maintenance. 

\section{Post-processing for structured pruning}\label{sec:app-pruning}
\begin{figure}[t]\vspace{-5mm}
	\centering    
	\subfigure[conv\_1]{
		\includegraphics[width=0.4\textwidth]{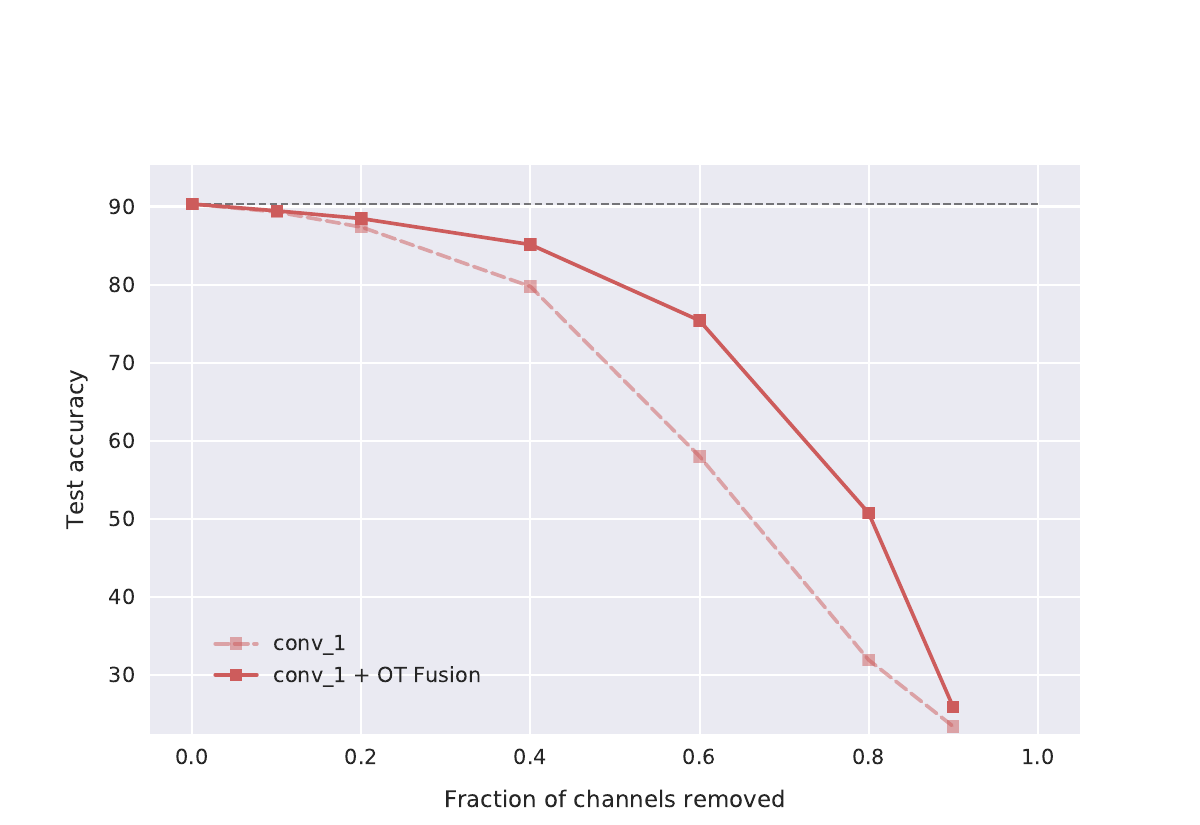}
	}
	\subfigure[conv\_2]{\includegraphics[width=0.4\textwidth]{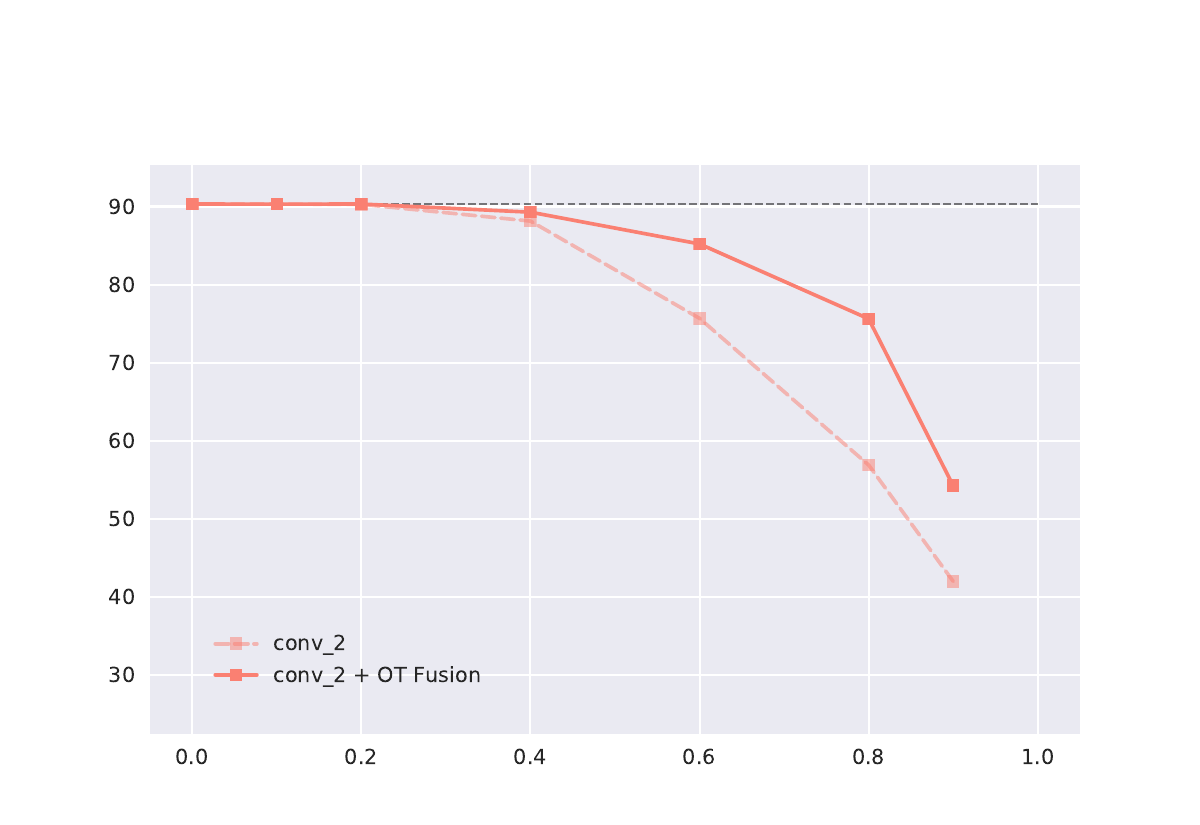}
	}\vspace{-3mm}
	\subfigure[conv\_3]{\includegraphics[width=0.4\textwidth]{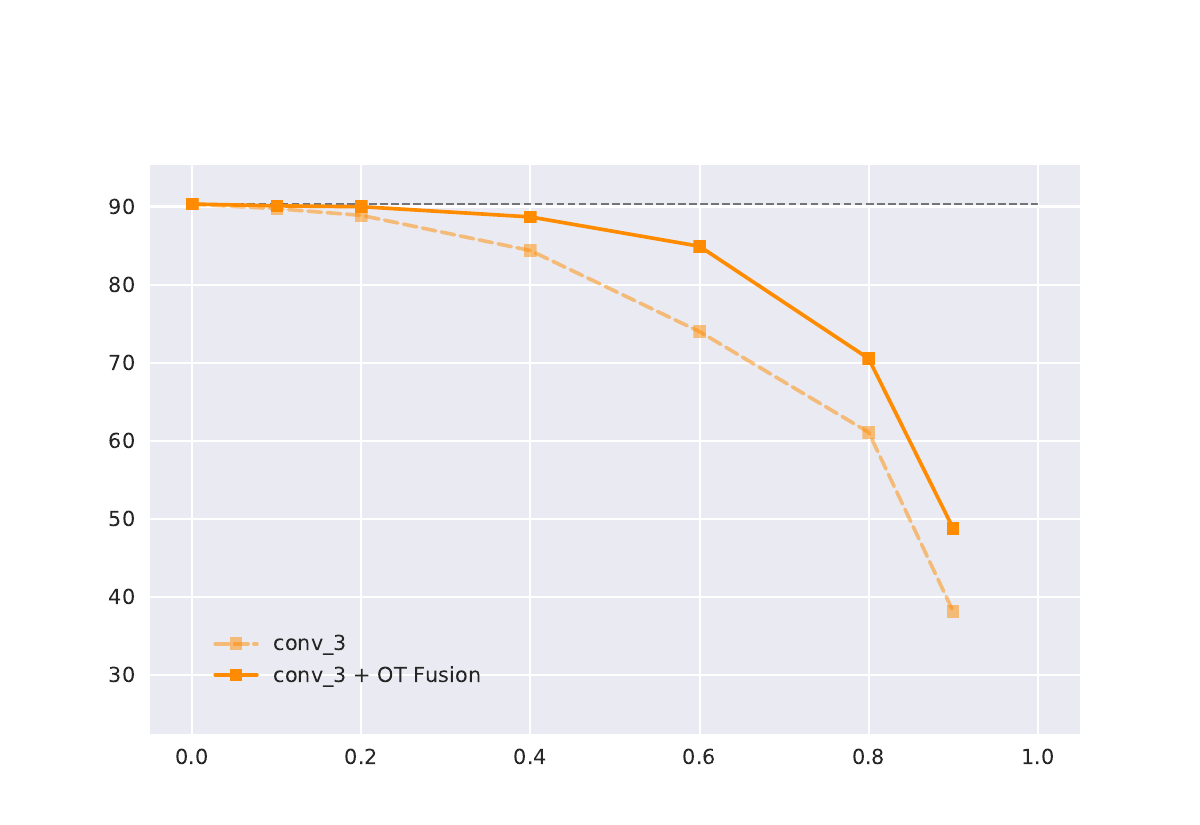}
	}
	\subfigure[conv\_4]{\includegraphics[width=0.4\textwidth]{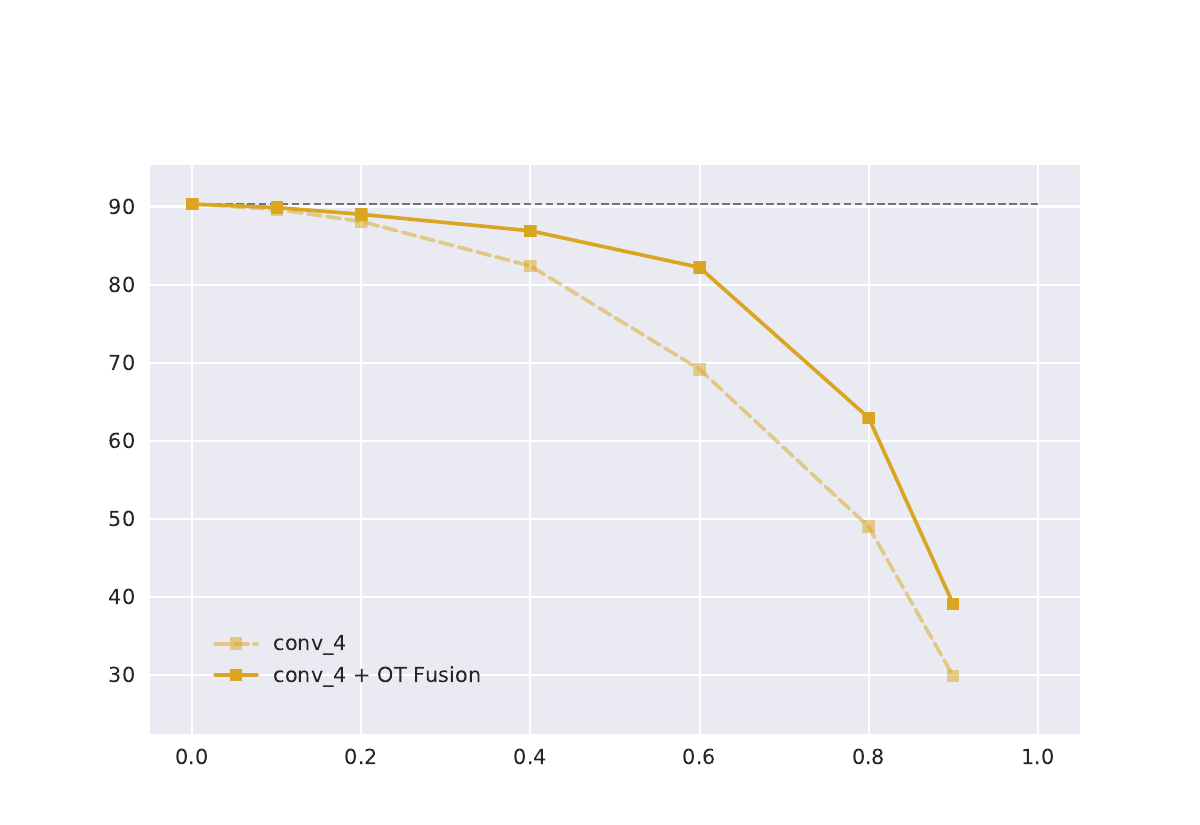}
	}\vspace{-3mm}
	\subfigure[conv\_5]{
		\includegraphics[width=0.4\textwidth]{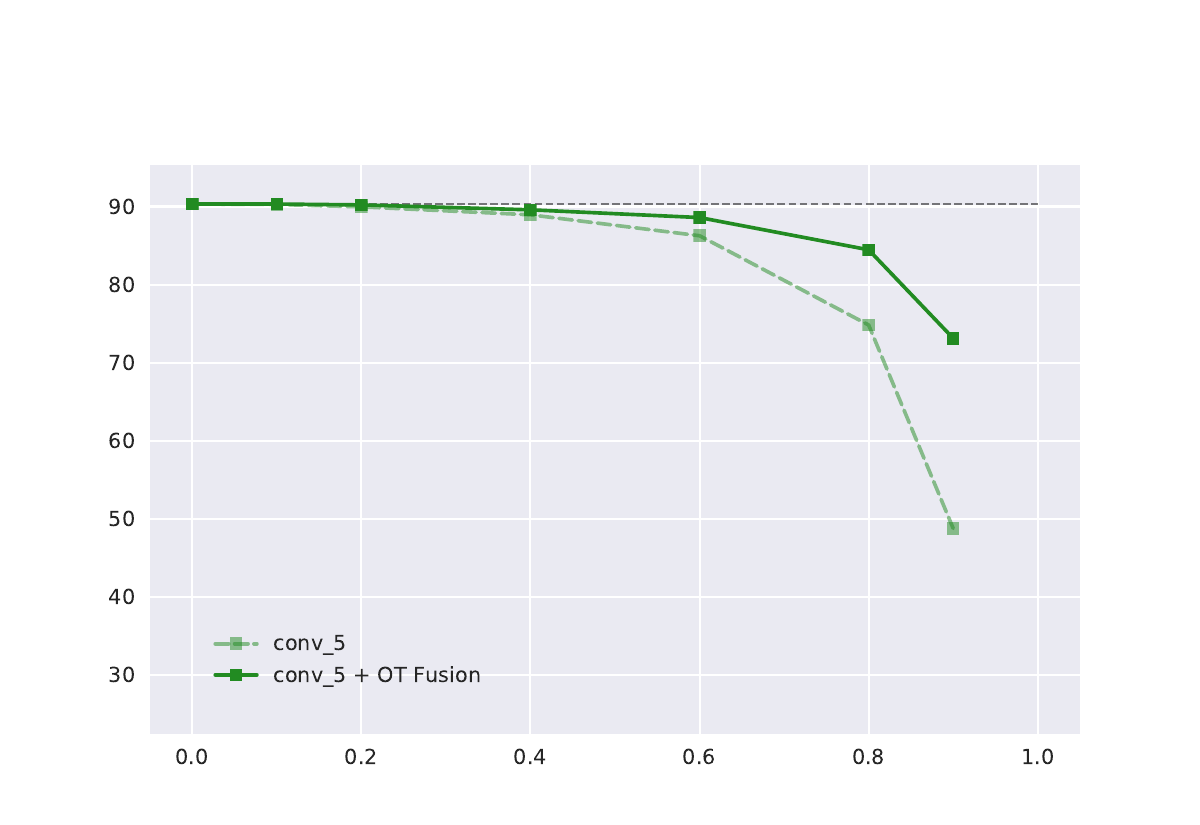}
	}
	\subfigure[conv\_6]{\includegraphics[width=0.4\textwidth]{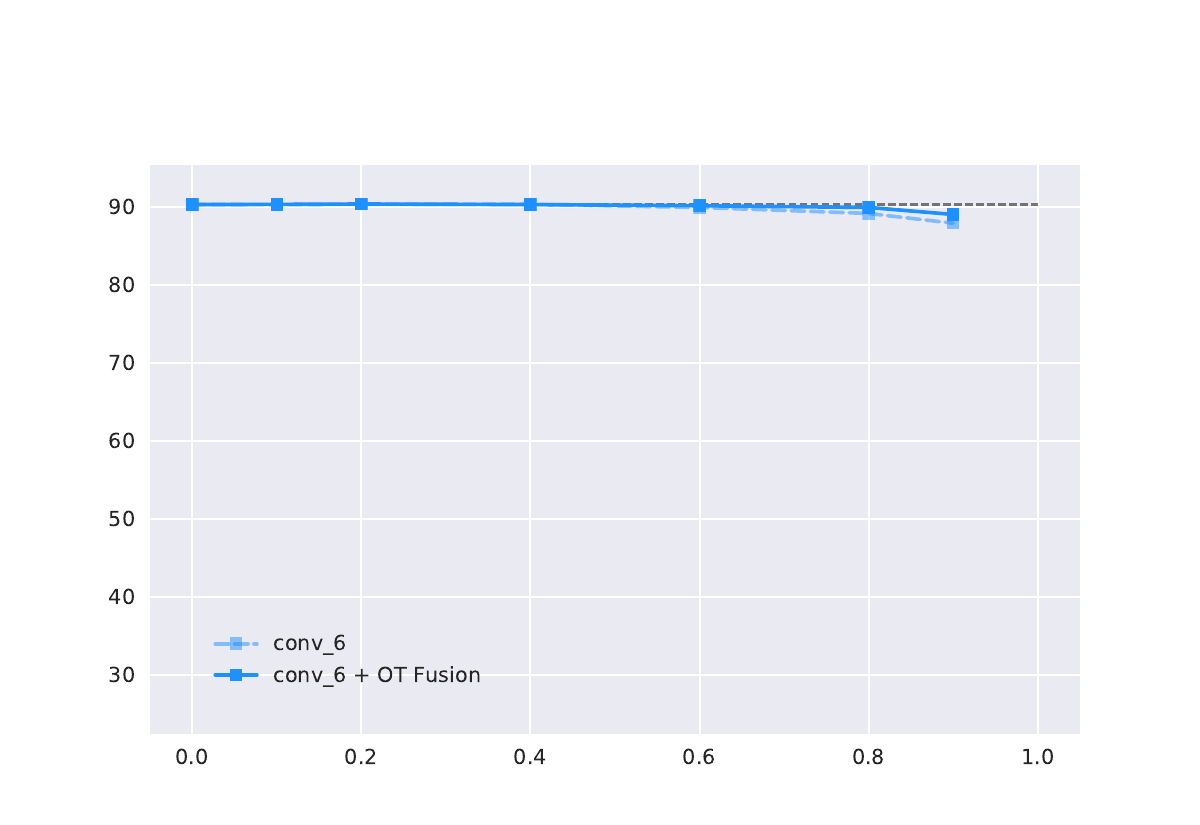}
	}\vspace{-3mm}
	\subfigure[conv\_7]{\includegraphics[width=0.4\textwidth]{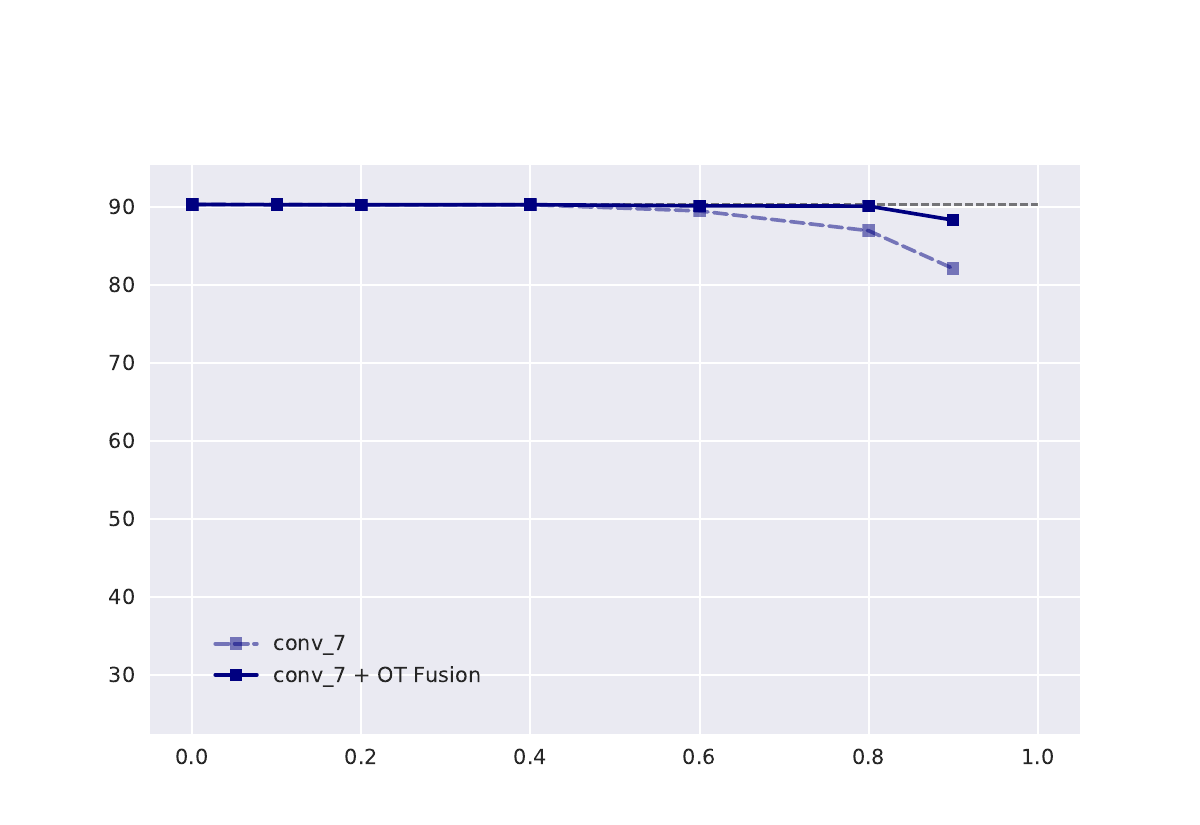}
	}
	\subfigure[conv\_8]{\includegraphics[width=0.4\textwidth]{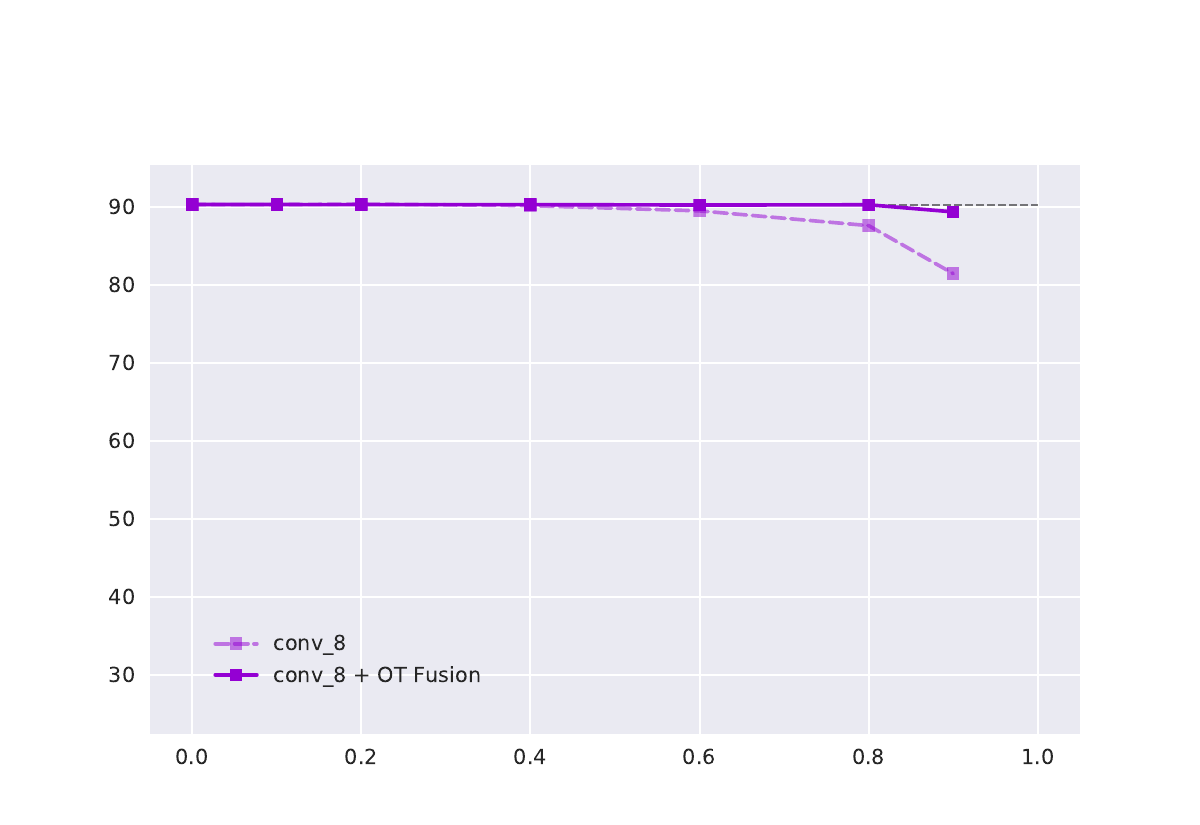}
	}\vspace{-3mm}
	\subfigure[all]{\includegraphics[width=0.4\textwidth]{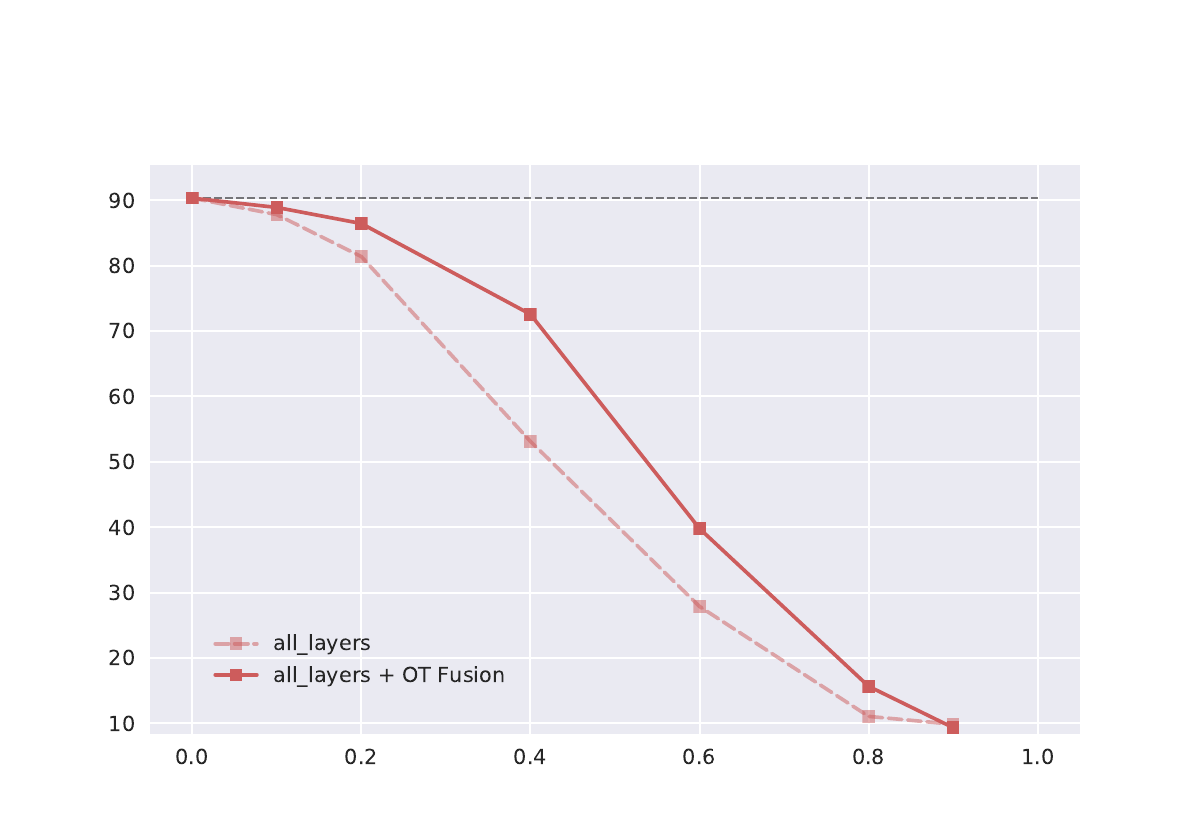}
	}
	\caption{Post-processing for structured pruning \textbf{with $\ell_1$ norm}, all figures: 
		Fusing the initial dense \textsc{VGG11} model into the pruned model helps test accuracy of the pruned model on \textsc{\textbf{CIFAR10}}.} %$m=400$, 
	\label{fig:l1_all}	
	\vspace{-1em}
\end{figure}

\begin{figure}[t]\vspace{-5mm}
	\centering    
	\subfigure[conv\_1]{
		\includegraphics[width=0.4\textwidth]{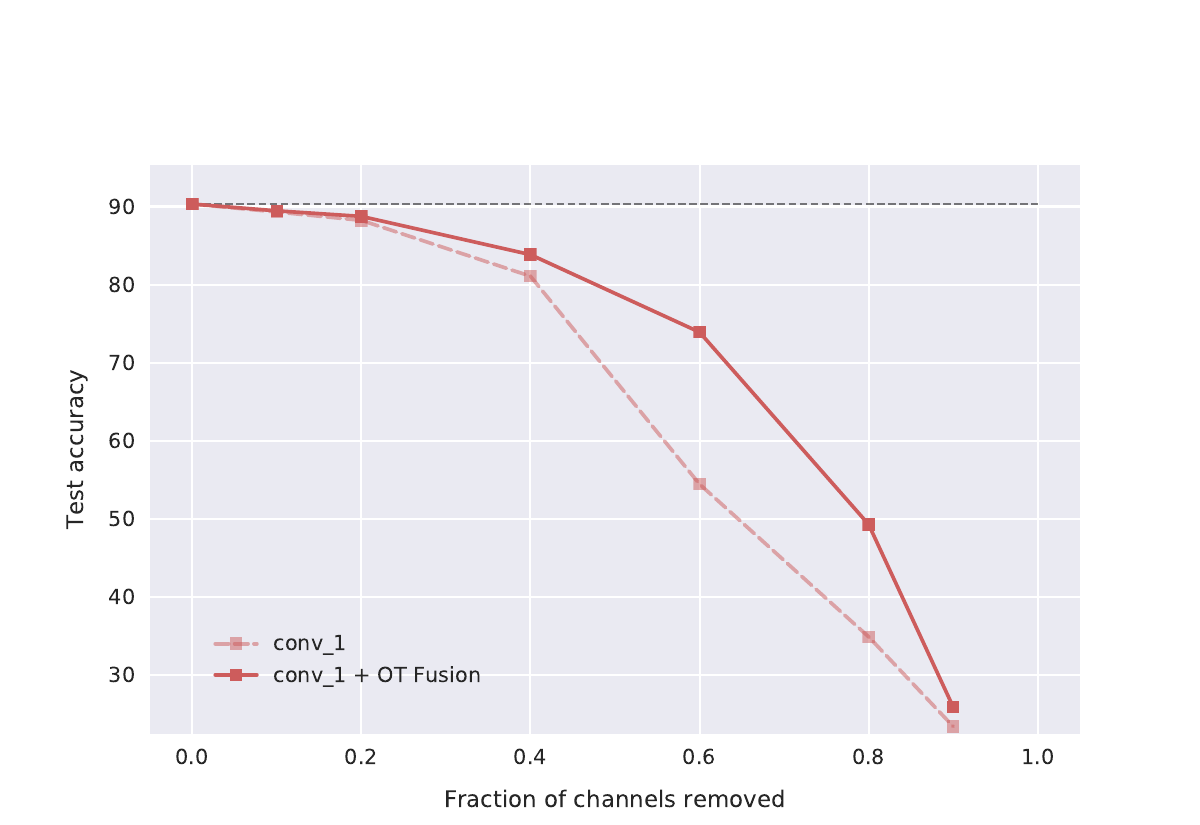}
	}
	\subfigure[conv\_2]{\includegraphics[width=0.4\textwidth]{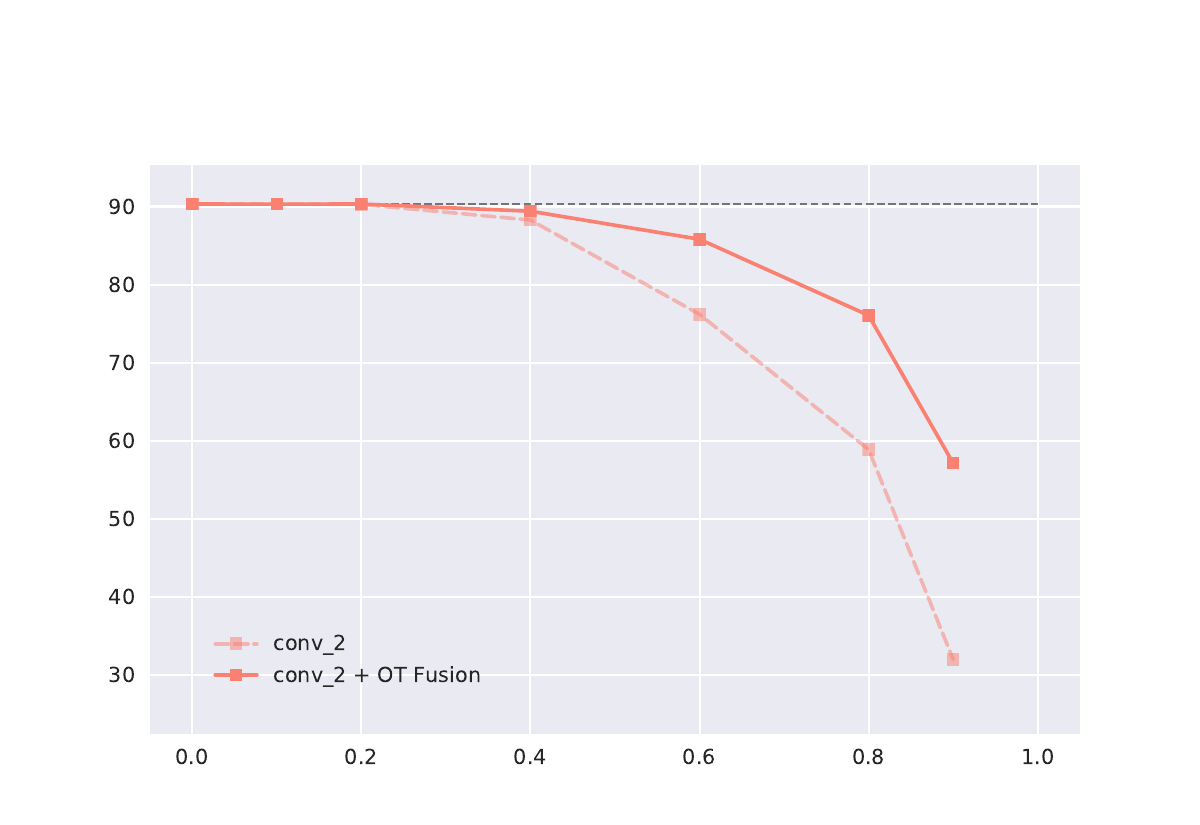}}
	\subfigure[conv\_3]{\includegraphics[width=0.4\textwidth]{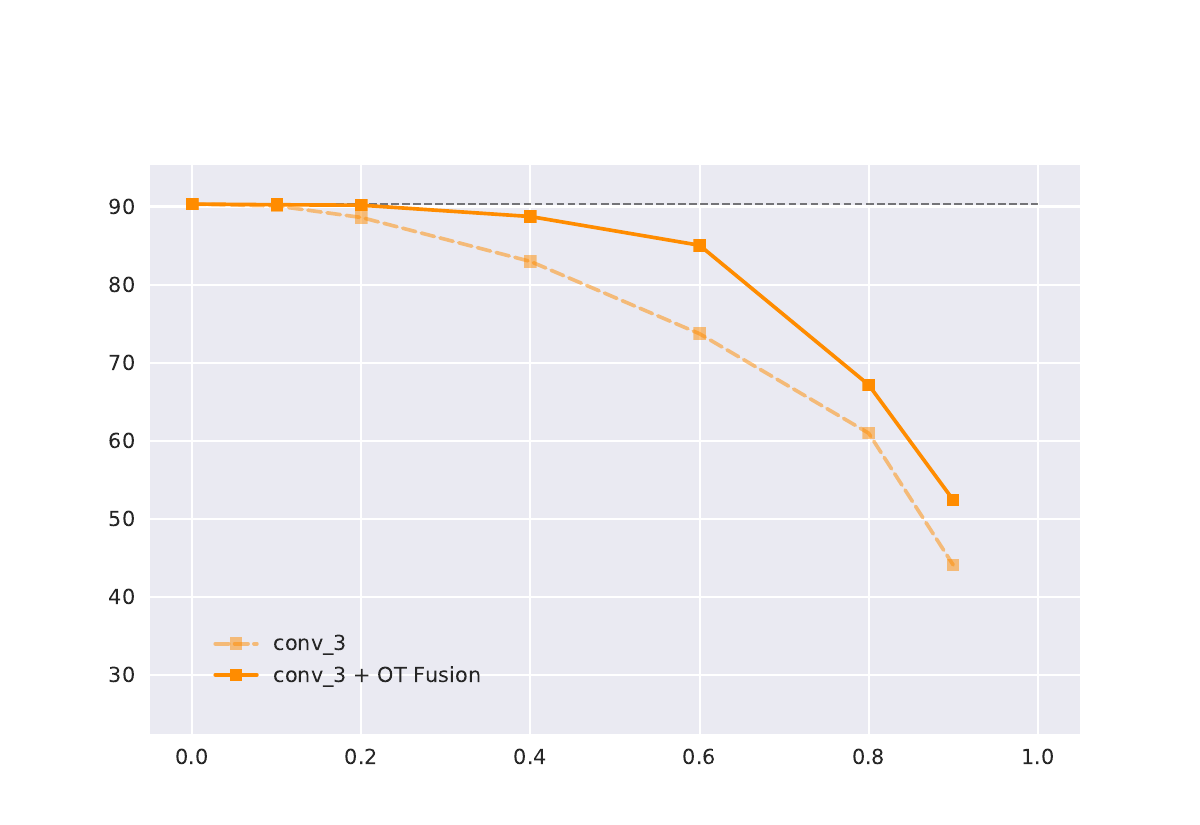}}
	\subfigure[conv\_4]{\includegraphics[width=0.4\textwidth]{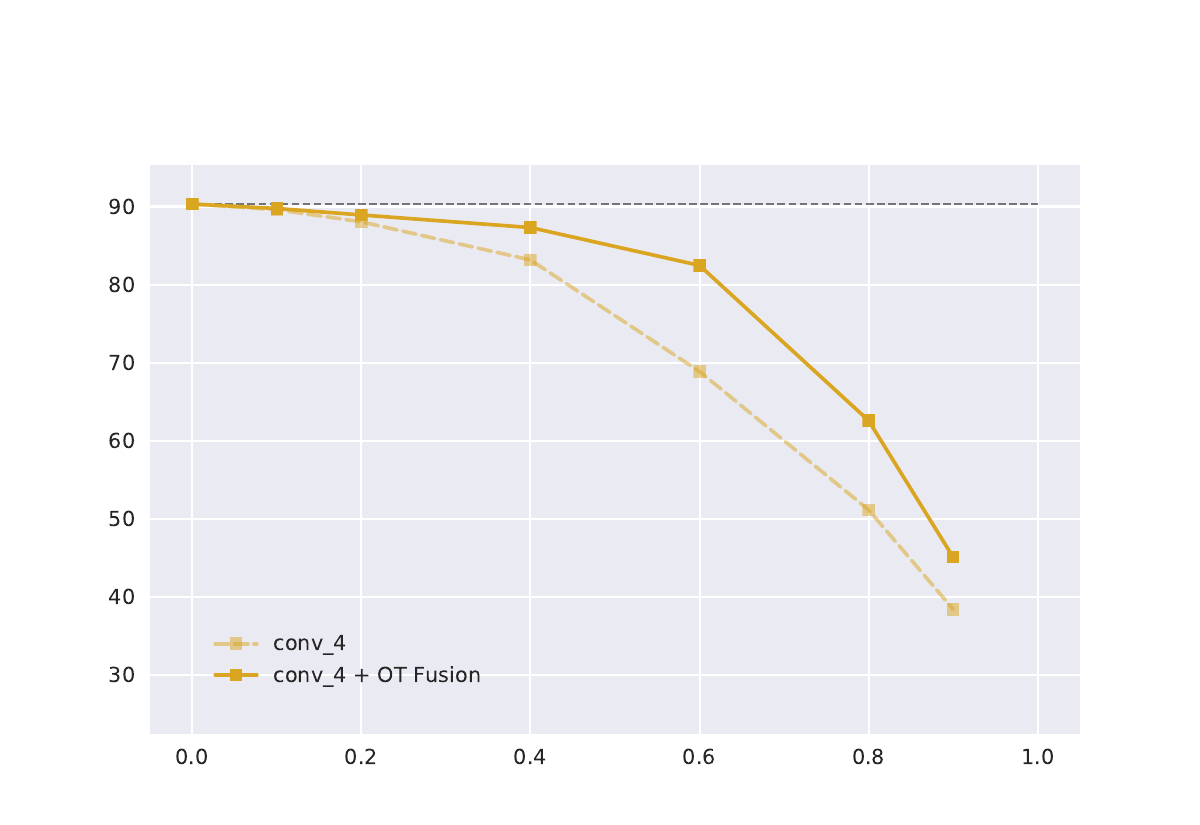}
	}
	\subfigure[conv\_5]{
		\includegraphics[width=0.4\textwidth]{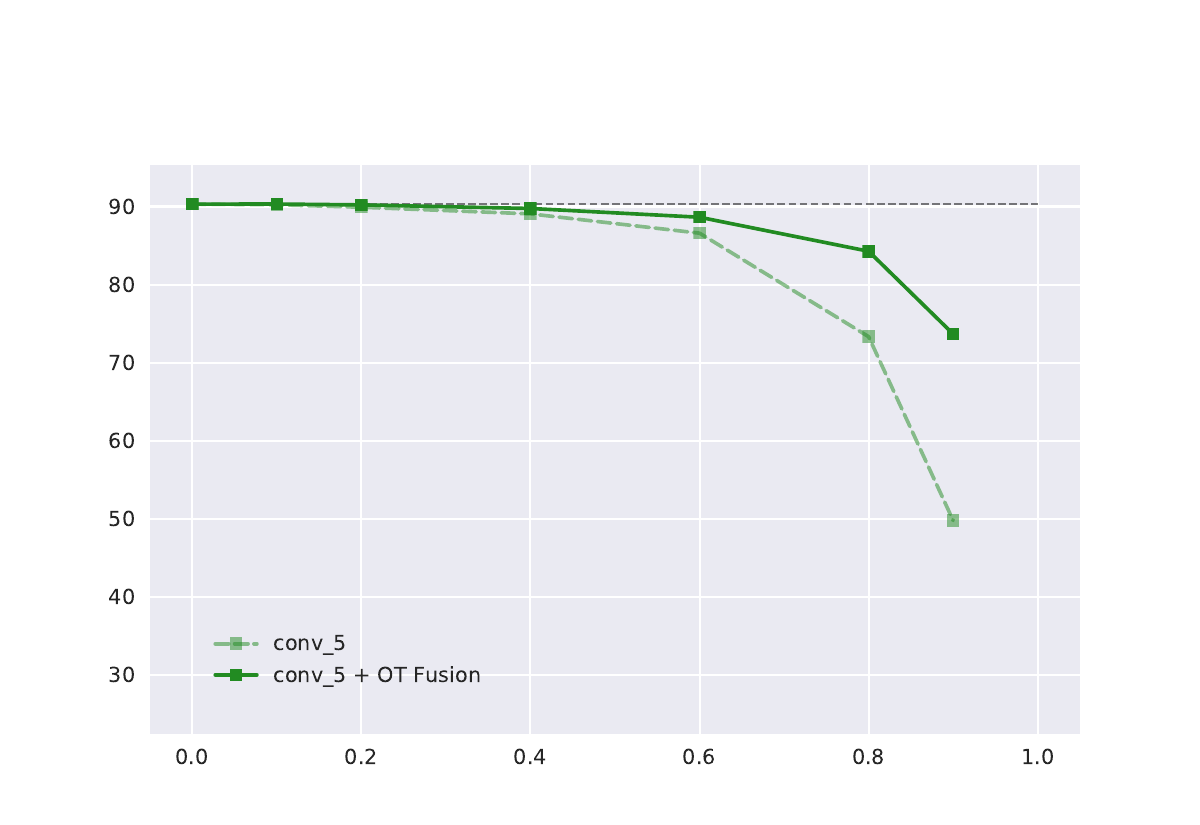}
	}
	\subfigure[conv\_6]{\includegraphics[width=0.4\textwidth]{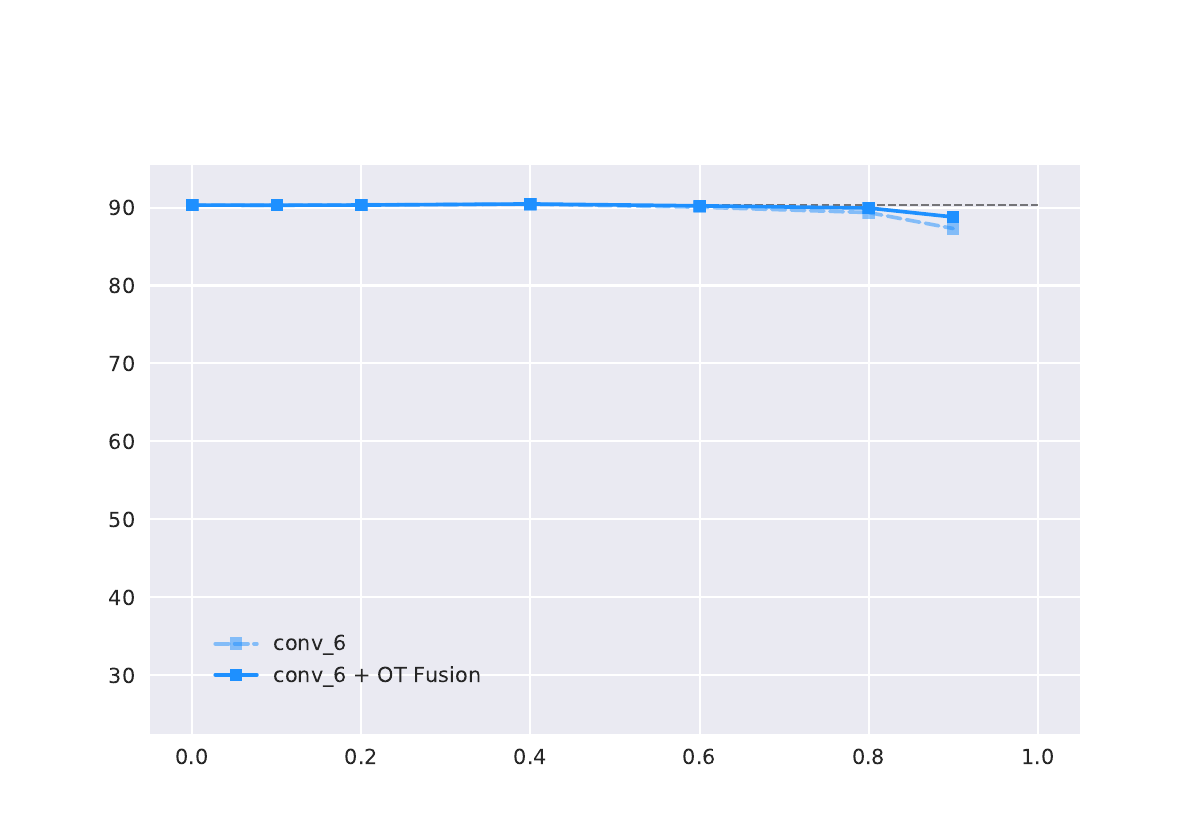}
	}
	\subfigure[conv\_7]{\includegraphics[width=0.4\textwidth]{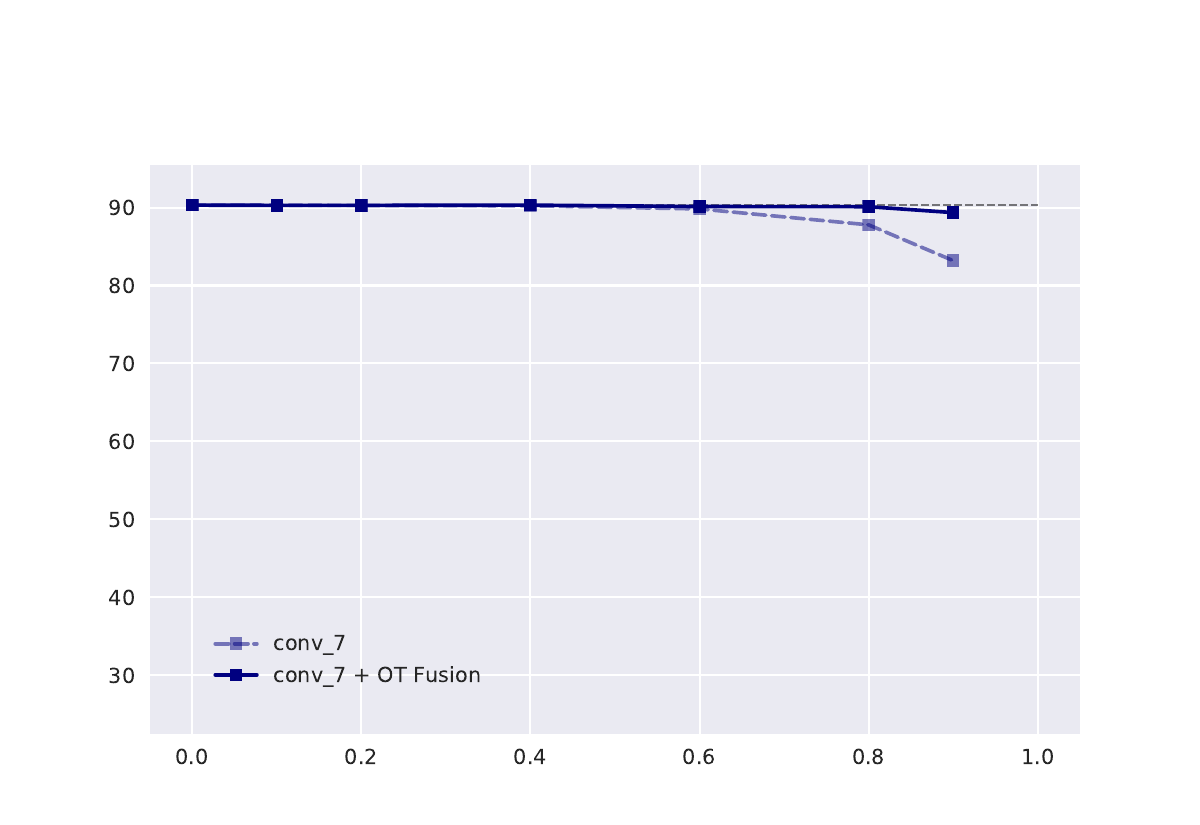}
	}
	\subfigure[conv\_8]{\includegraphics[width=0.4\textwidth]{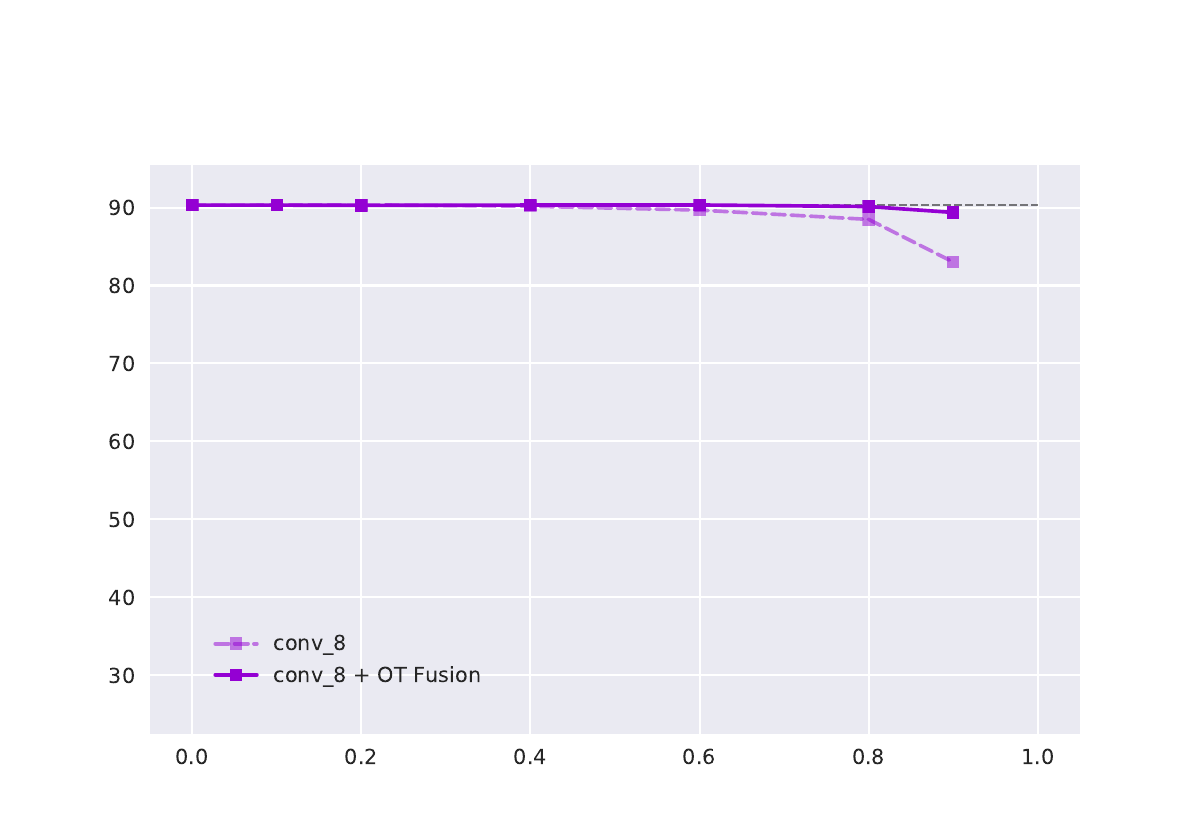}
	}
	\subfigure[all]{\includegraphics[width=0.4\textwidth]{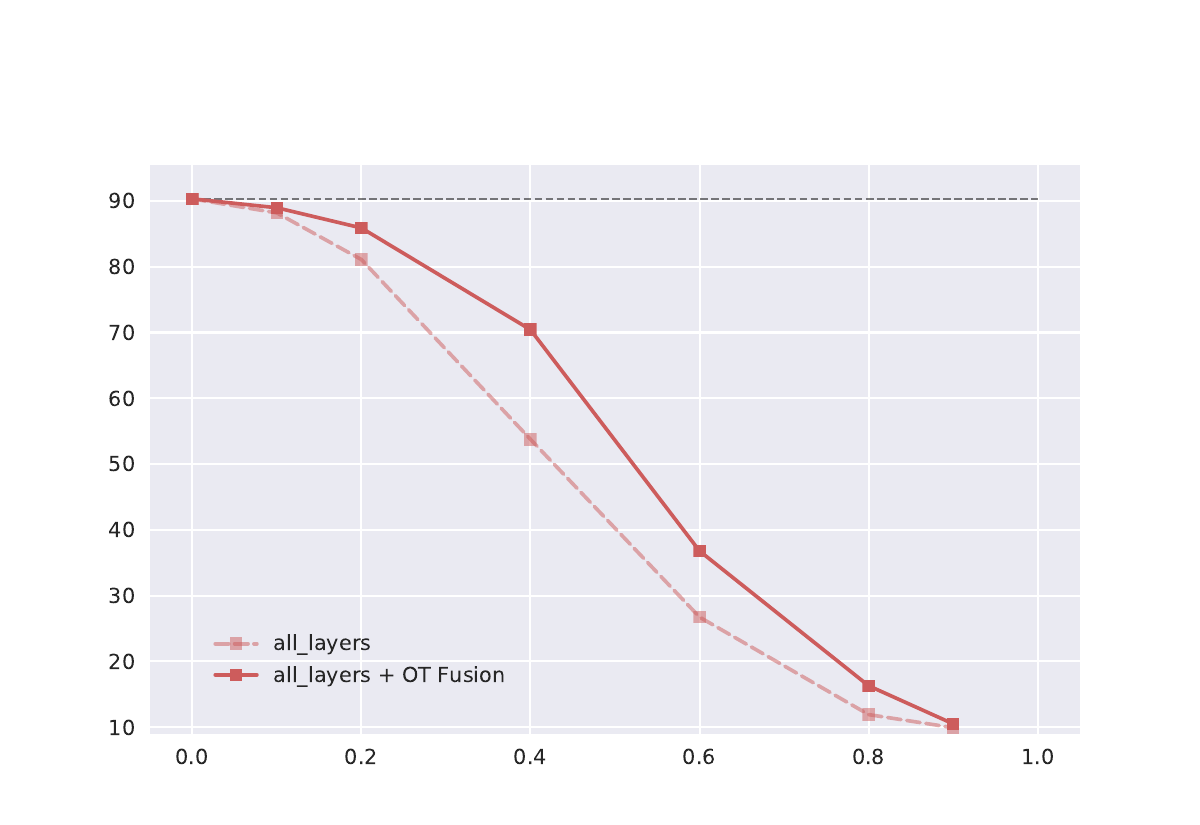}
	}
	\caption{Post-processing for structured pruning \textbf{with $\ell_2$ norm}, all figures: 
		Fusing the initial dense \textsc{VGG11} model into the pruned model helps test accuracy of the pruned model on \textsc{\textbf{CIFAR10}}.} %$m=400$, 
	\label{fig:l2_all}	
	\vspace{-1em}
\end{figure}

\begin{figure}[t]\vspace{-5mm}
	\centering    
	\subfigure[conv\_1]{
		\includegraphics[width=0.4\textwidth]{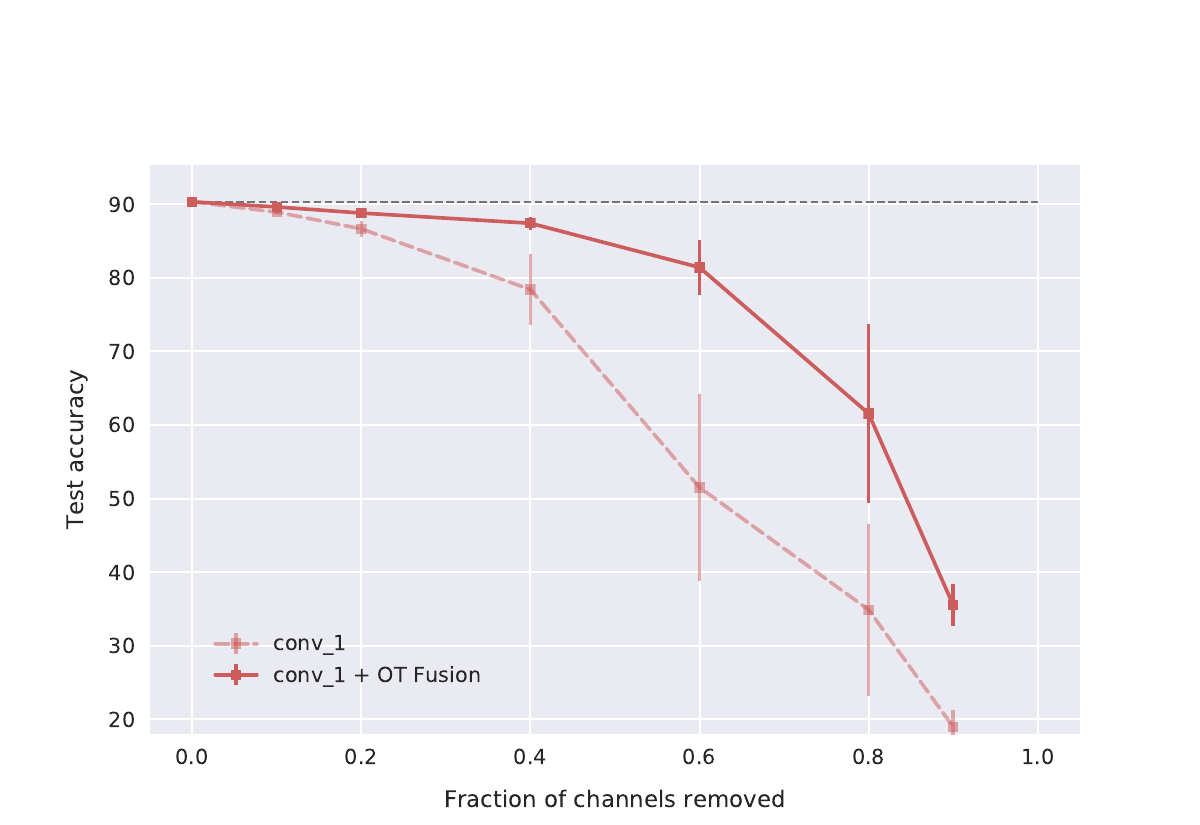}
	}
	\subfigure[conv\_2]{\includegraphics[width=0.4\textwidth]{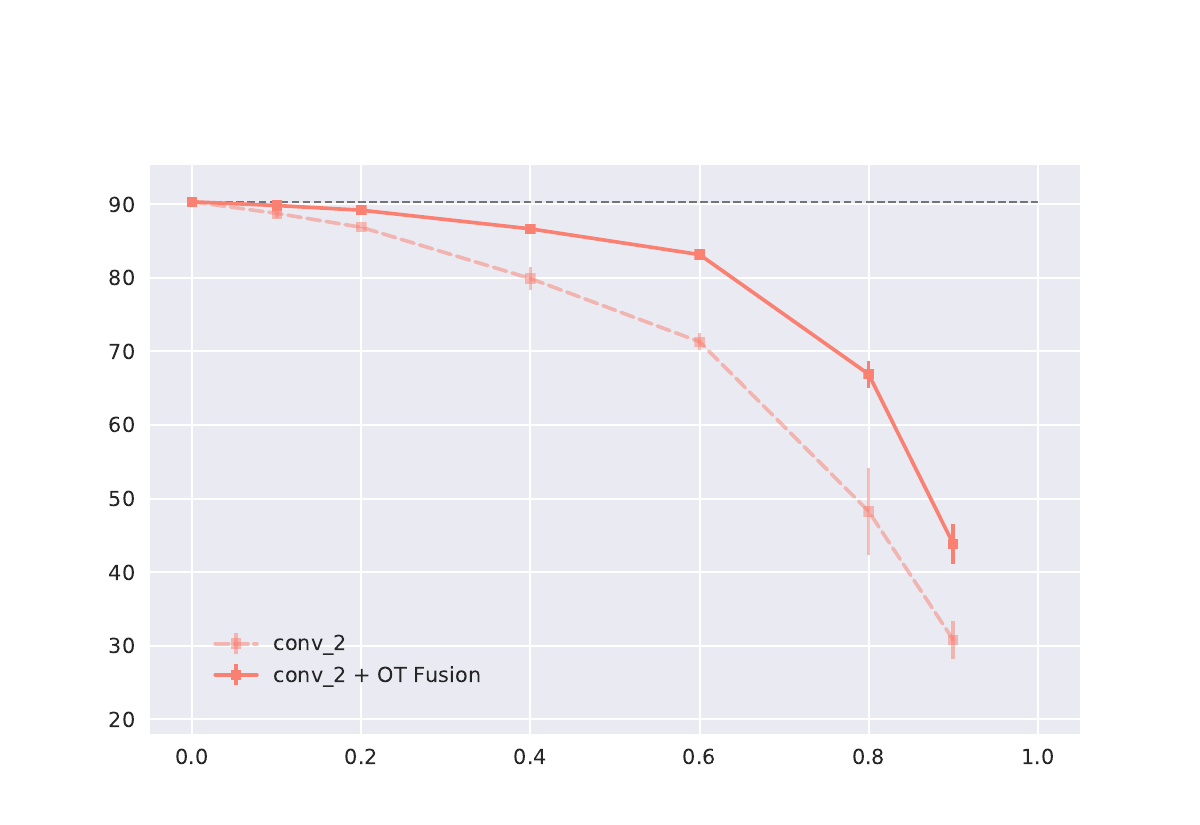}}\vspace{-3mm}
	\subfigure[conv\_3]{\includegraphics[width=0.4\textwidth]{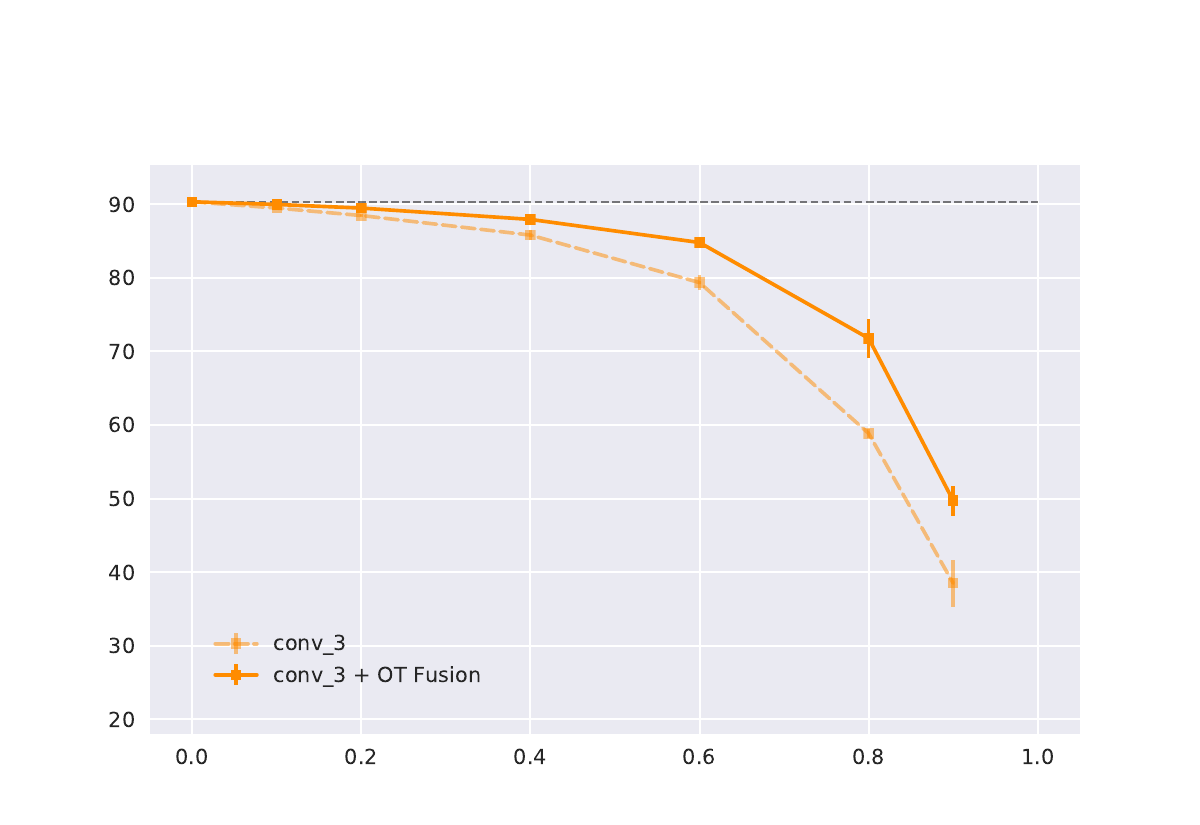}}
	\subfigure[conv\_4]{\includegraphics[width=0.4\textwidth]{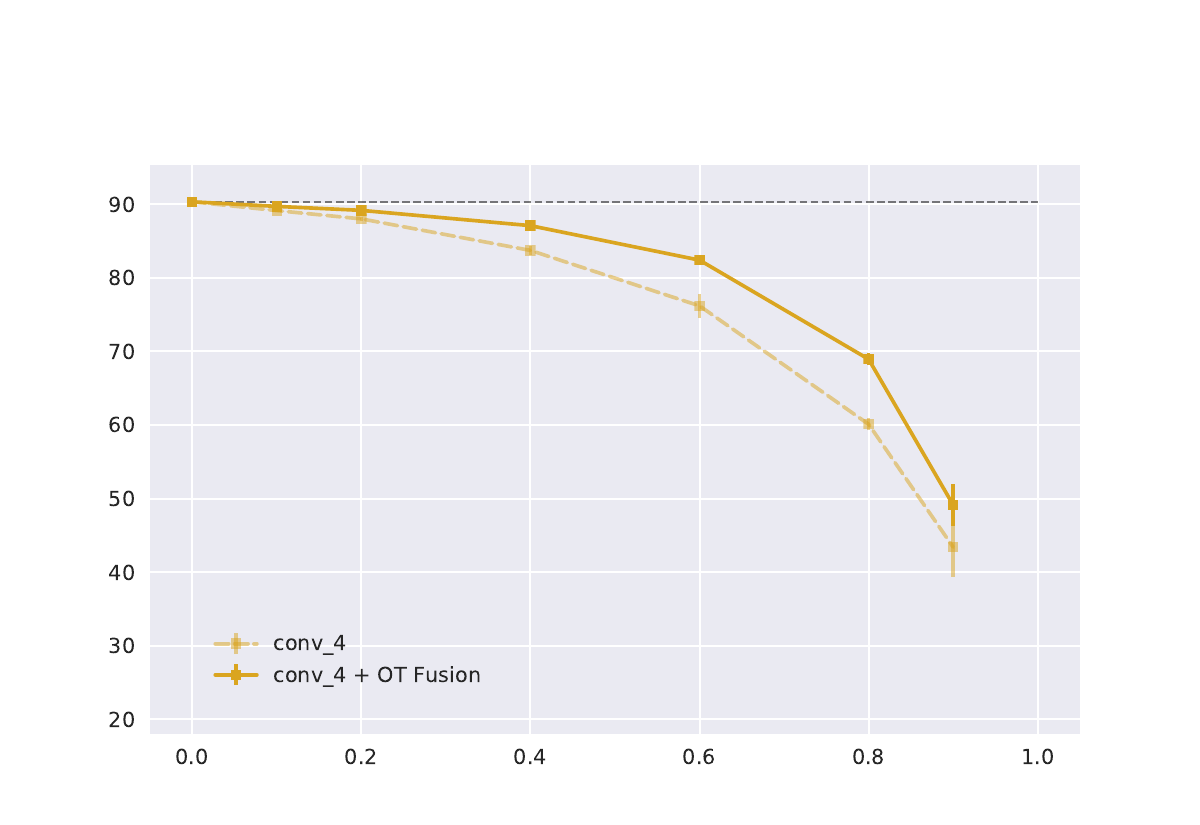}
	}\vspace{-3mm}
	\subfigure[conv\_5]{
		\includegraphics[width=0.4\textwidth]{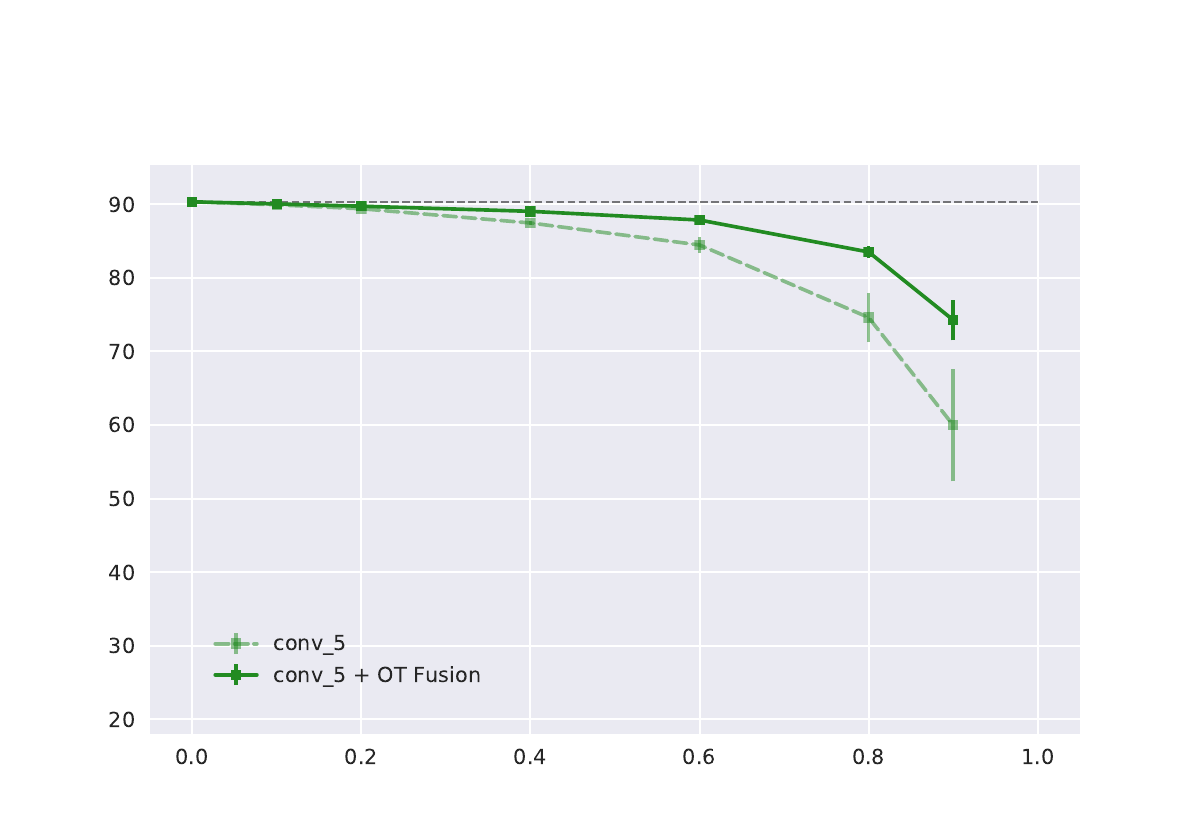}
	}
	\subfigure[conv\_6]{\includegraphics[width=0.4\textwidth]{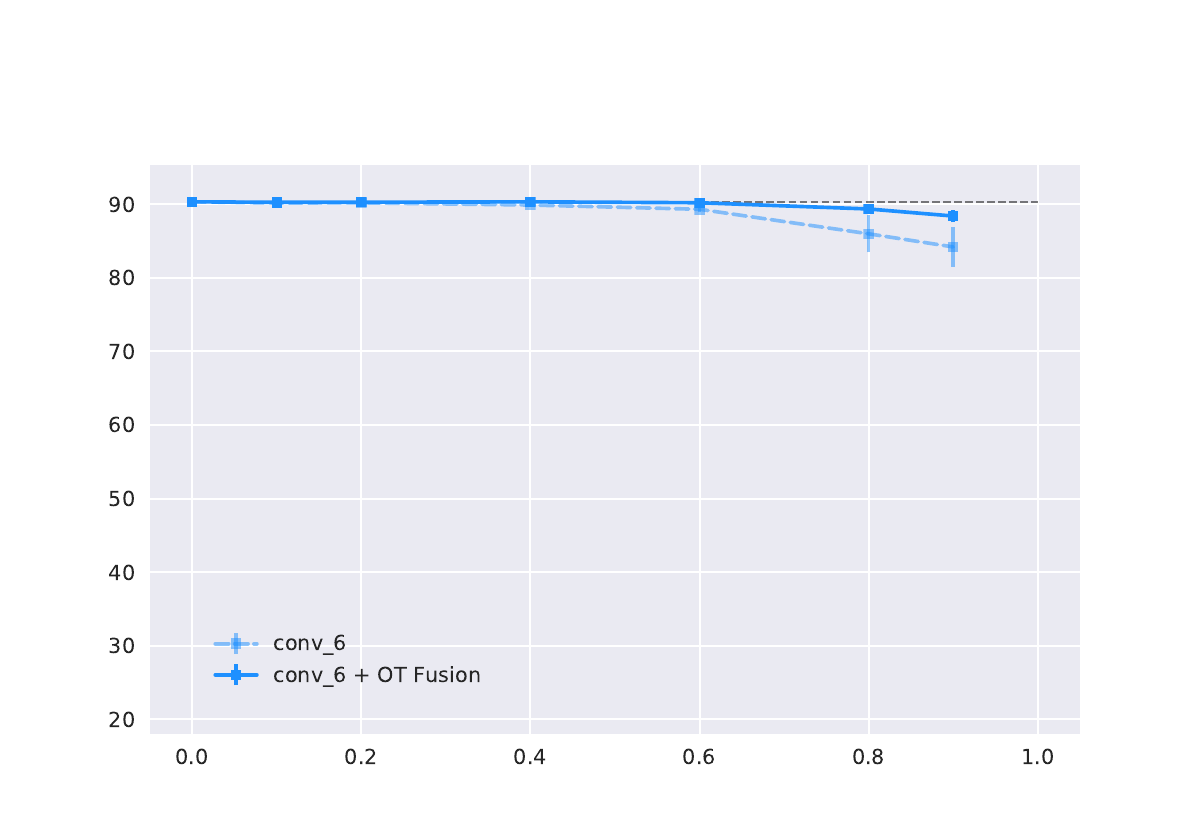}
	}\vspace{-3mm}
	\subfigure[conv\_7]{\includegraphics[width=0.4\textwidth]{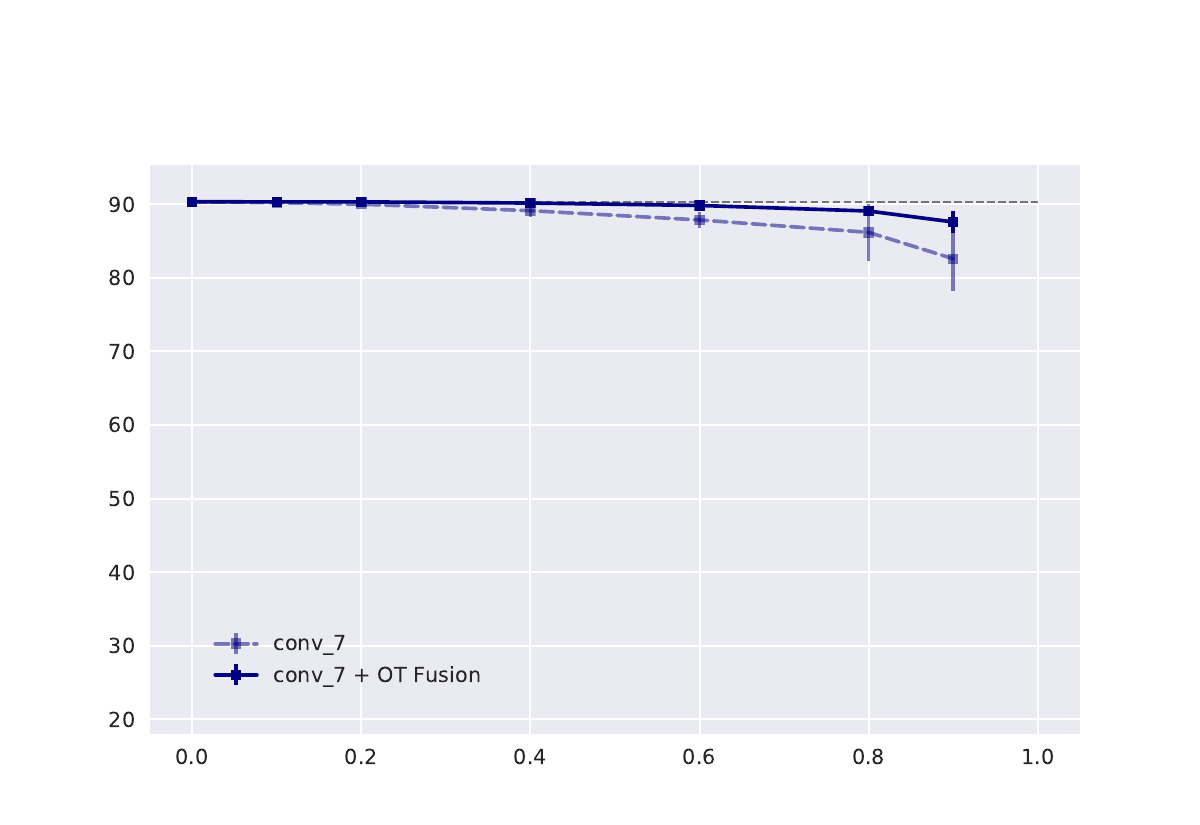}
	}
	\subfigure[conv\_8]{\includegraphics[width=0.4\textwidth]{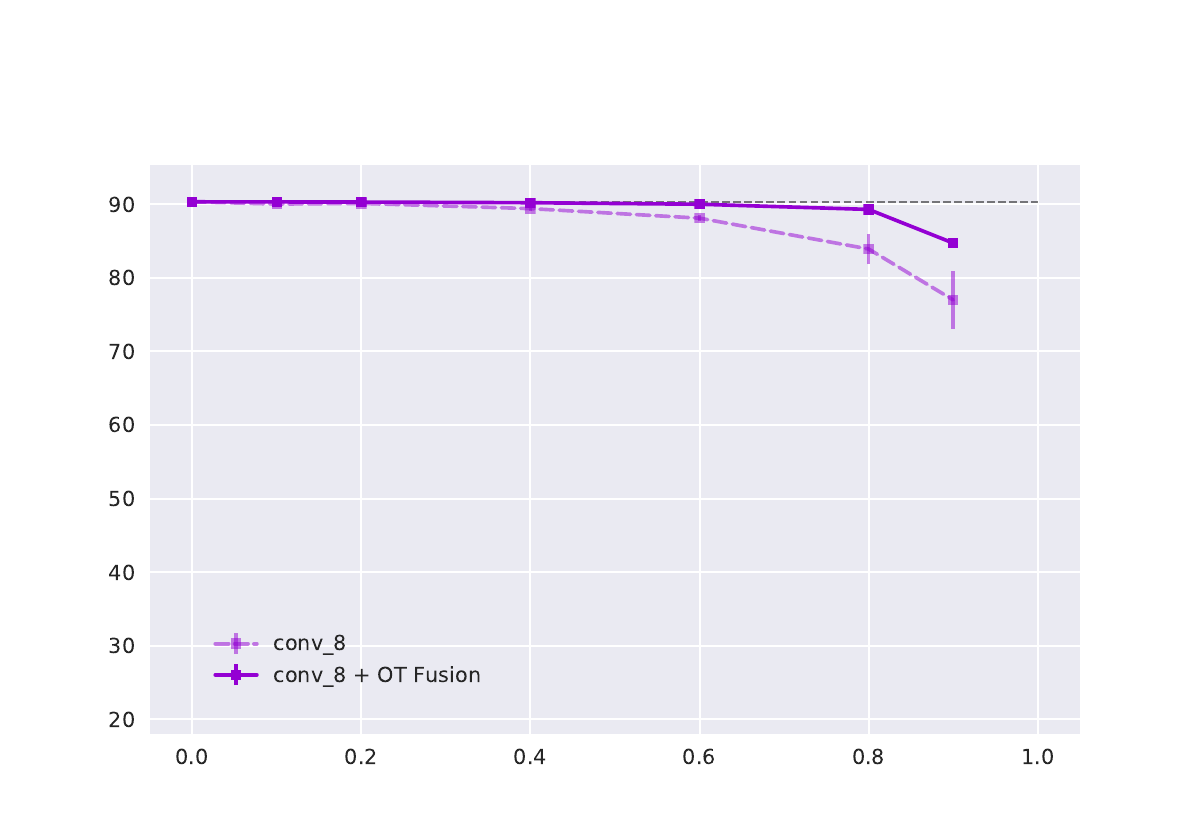}
	}\vspace{-3mm}
	\subfigure[all]{\includegraphics[width=0.4\textwidth]{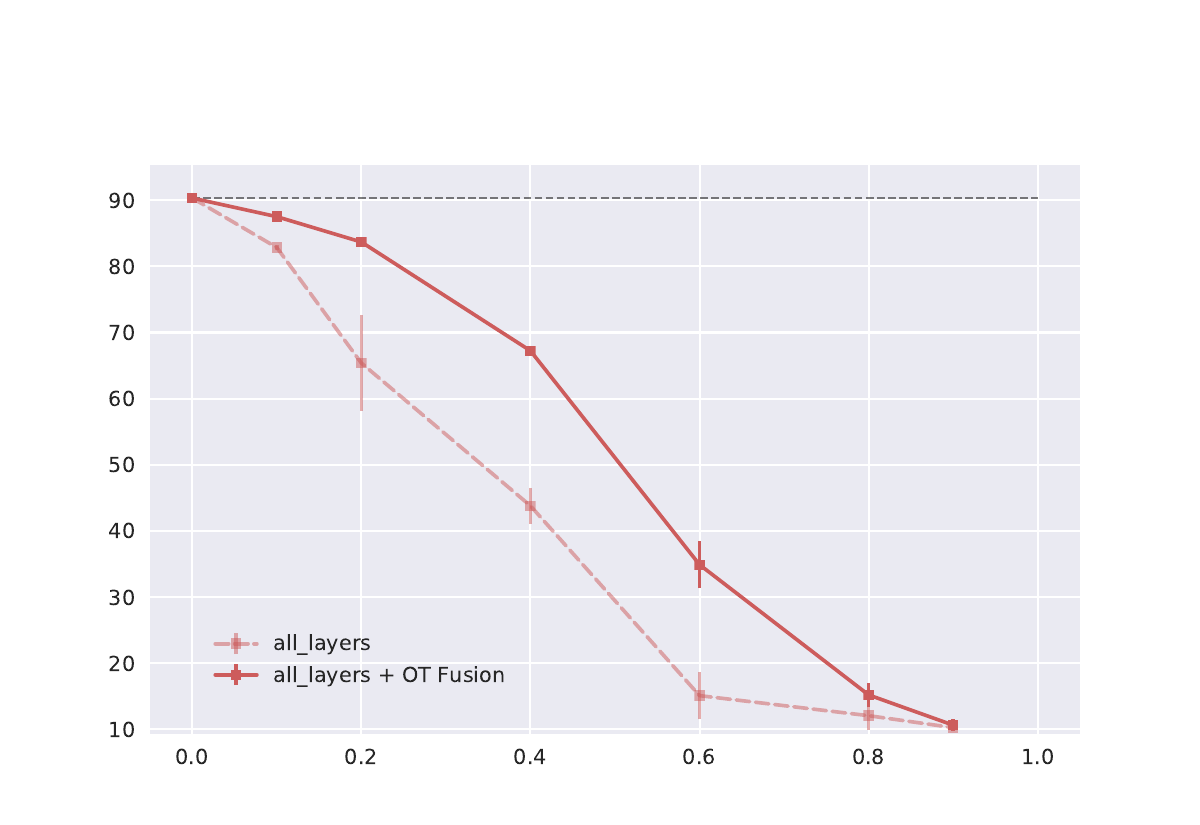}
	}
	\caption{Post-processing for structured pruning \textbf{with random}, all figures: 
		Fusing the initial dense \textsc{VGG11} model into the pruned model helps test accuracy of the pruned model on \textsc{\textbf{CIFAR10}}. Results are averaged over 3 seeds.} %$m=400$, 
	\label{fig:random_all}	
	\vspace{-1em}
\end{figure}

\paragraph{CIFAR10.} In this section, we present the detailed results for using OT fusion as a post-processing tool for structured pruning. We show the benefit gained by OT fusion when separately pruning all layers of \textsc{VGG11}, as well as pruning them all together. This is illustrated for the three cases: (a) when filters with smallest $\ell_1$ norms are removed, (b) when filters with smallest $\ell_2$ norms are removed, and (c) when filters are removed randomly, in Figures~\ref{fig:l1_all}, ~\ref{fig:l2_all}, and ~\ref{fig:random_all} respectively.

\begin{figure}[t]\vspace{-5mm}
	\centering    
	\subfigure[conv\_1]{
		\includegraphics[width=0.4\textwidth]{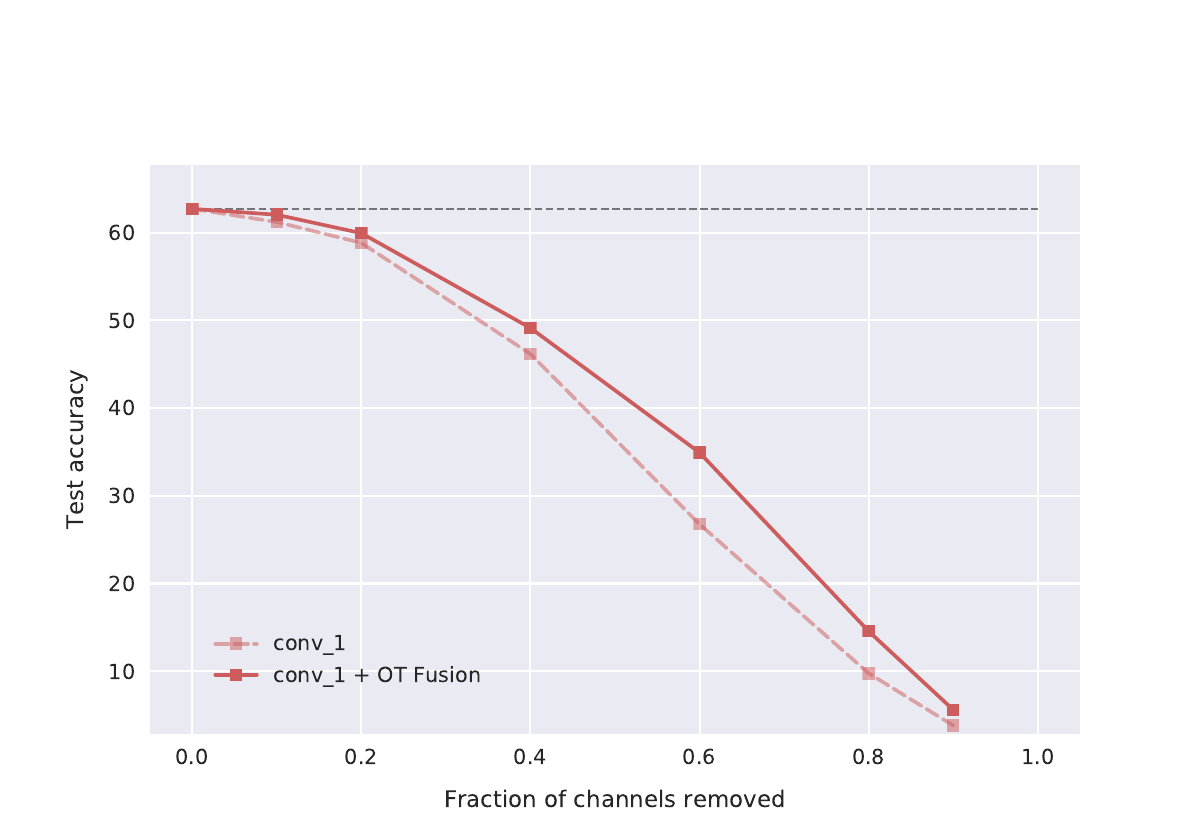}
	}
	\subfigure[conv\_2]{\includegraphics[width=0.4\textwidth]{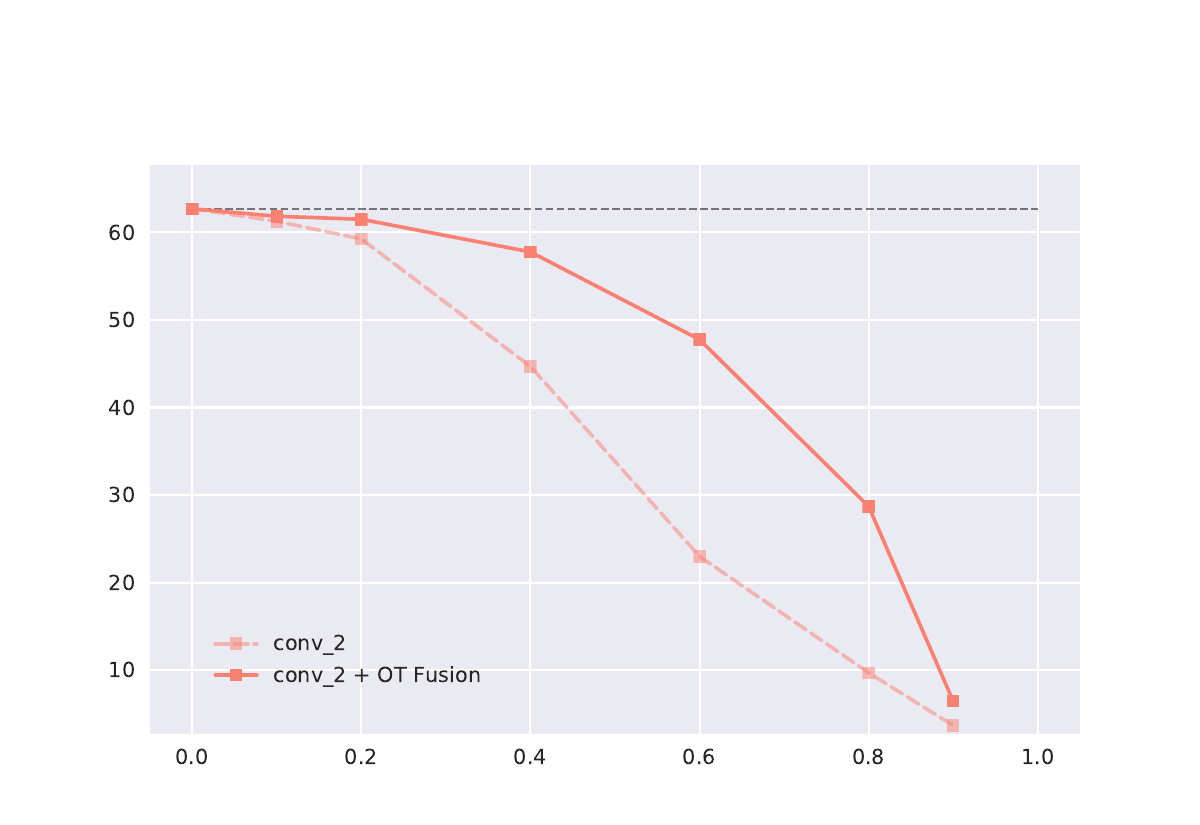}
	}\vspace{-3mm}
	\subfigure[conv\_3]{\includegraphics[width=0.4\textwidth]{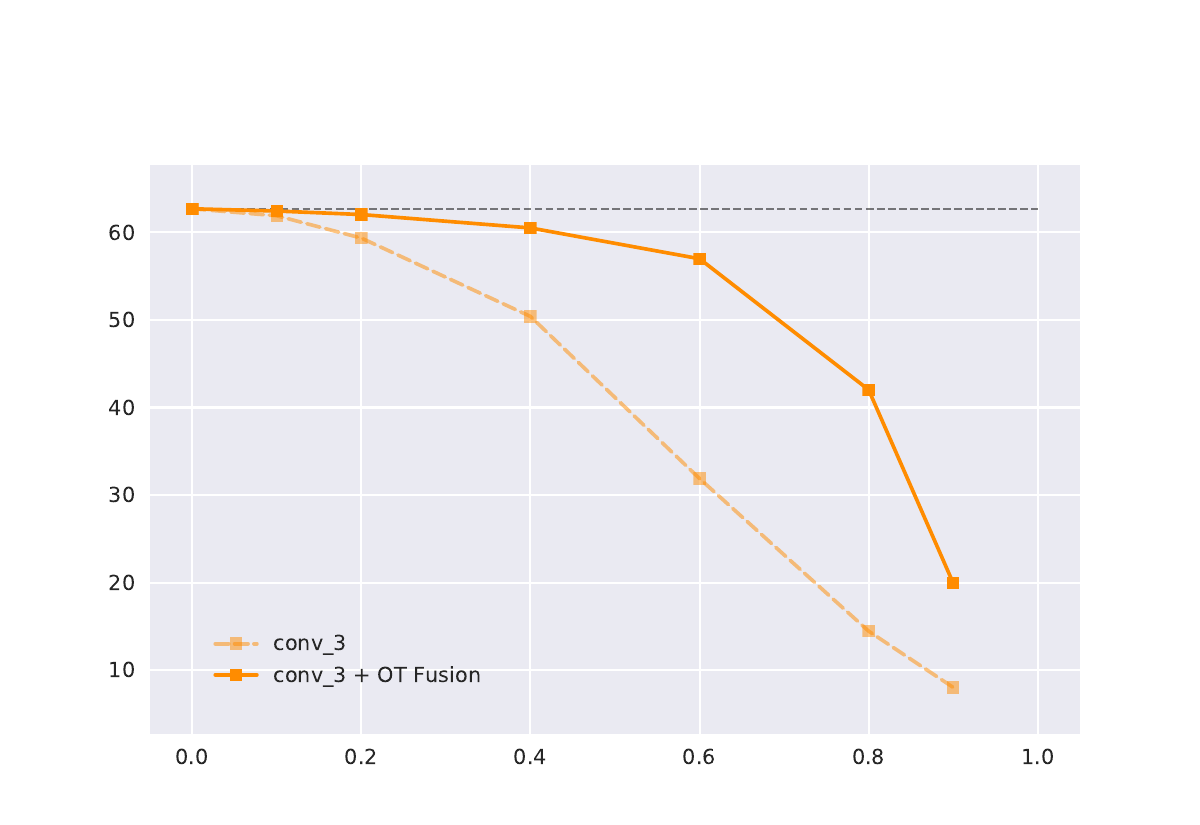}}
	\subfigure[conv\_4]{\includegraphics[width=0.4\textwidth]{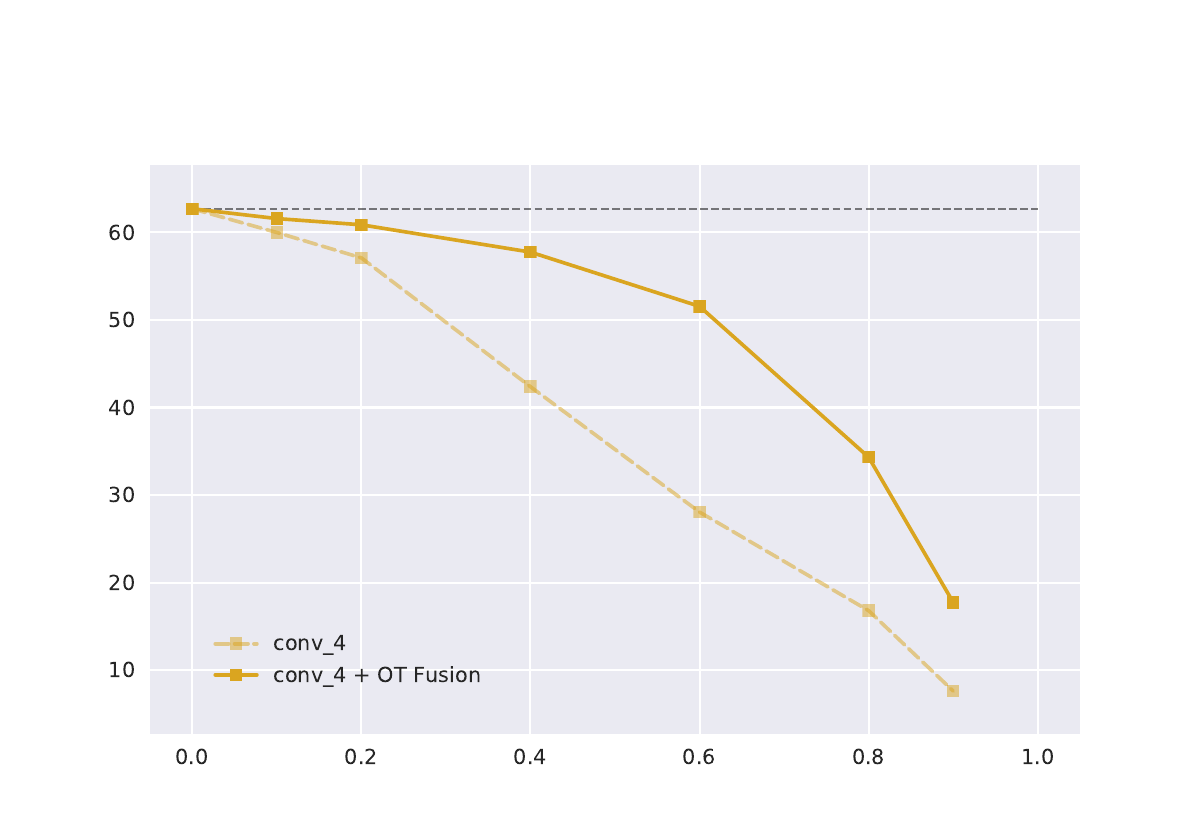}
	}\vspace{-3mm}
	\subfigure[conv\_5]{
		\includegraphics[width=0.4\textwidth]{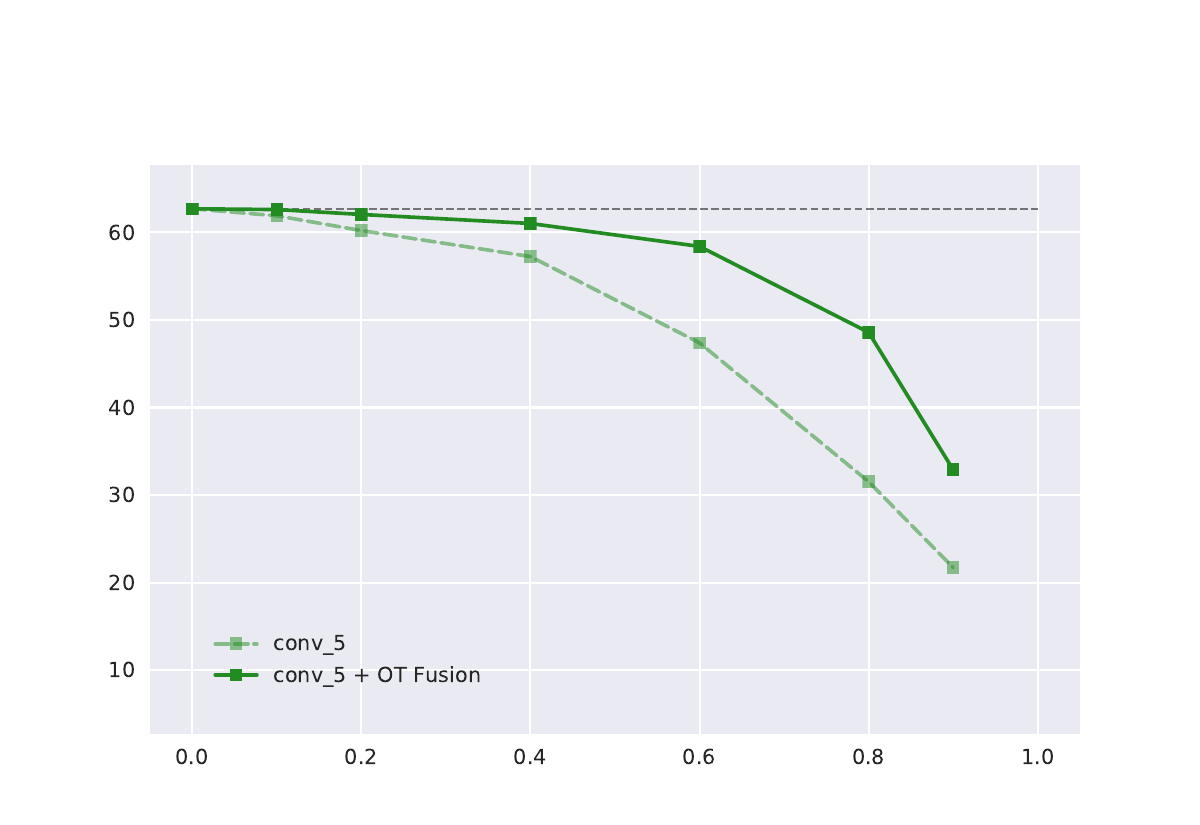}
	}
	\subfigure[conv\_6]{\includegraphics[width=0.4\textwidth]{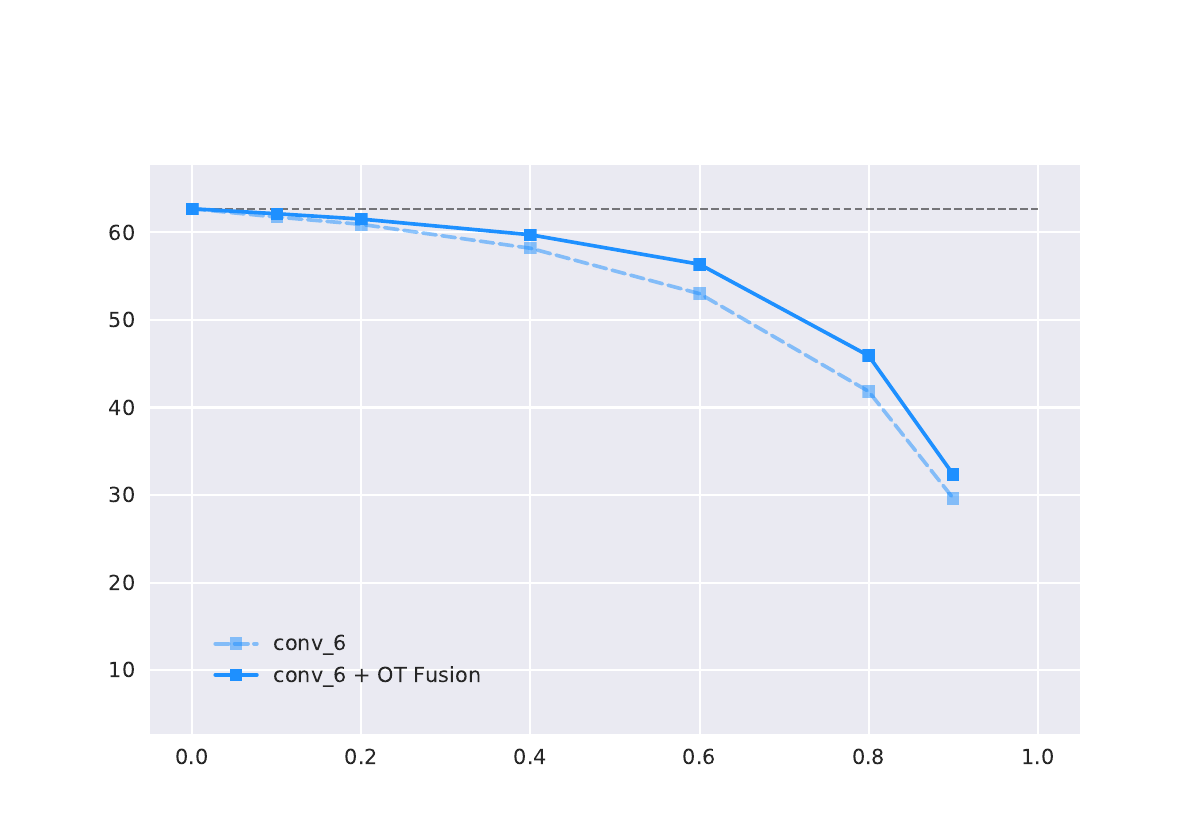}
	}\vspace{-3mm}
	\subfigure[conv\_7]{\includegraphics[width=0.4\textwidth]{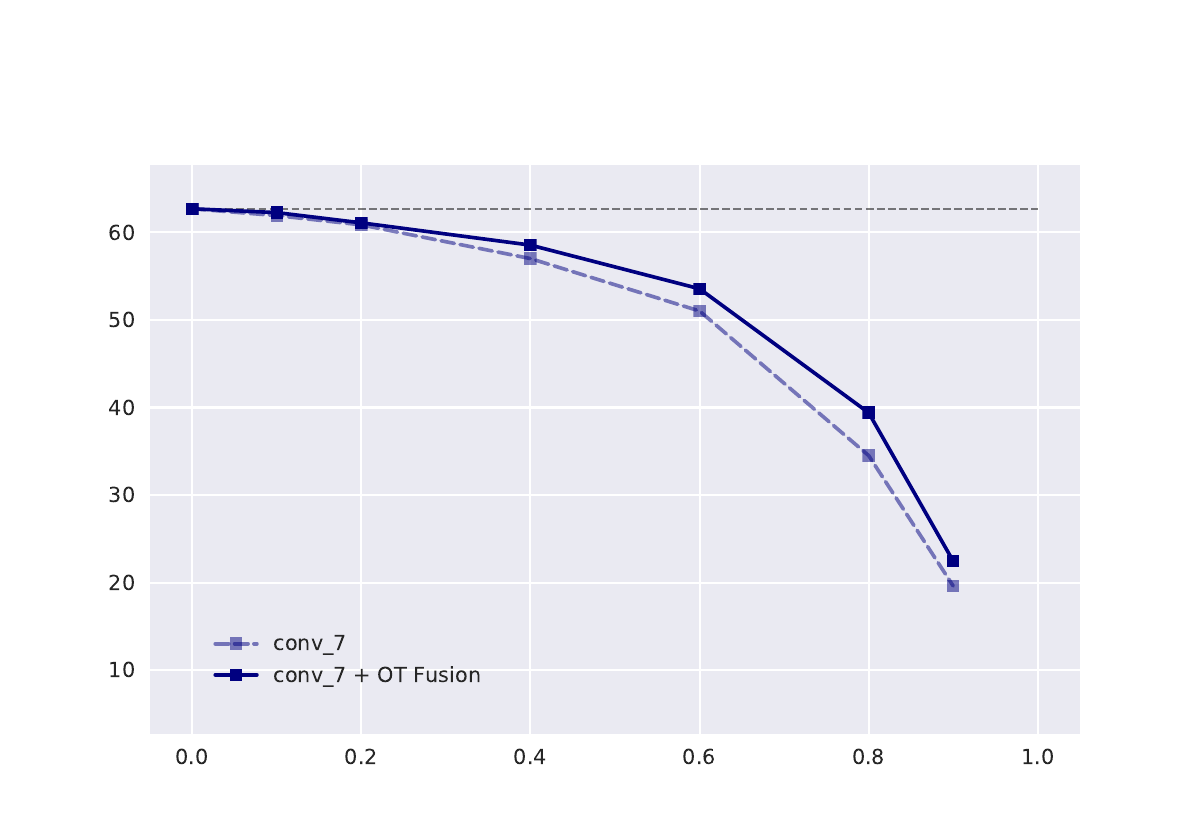}
	}
	\subfigure[conv\_8]{\includegraphics[width=0.4\textwidth]{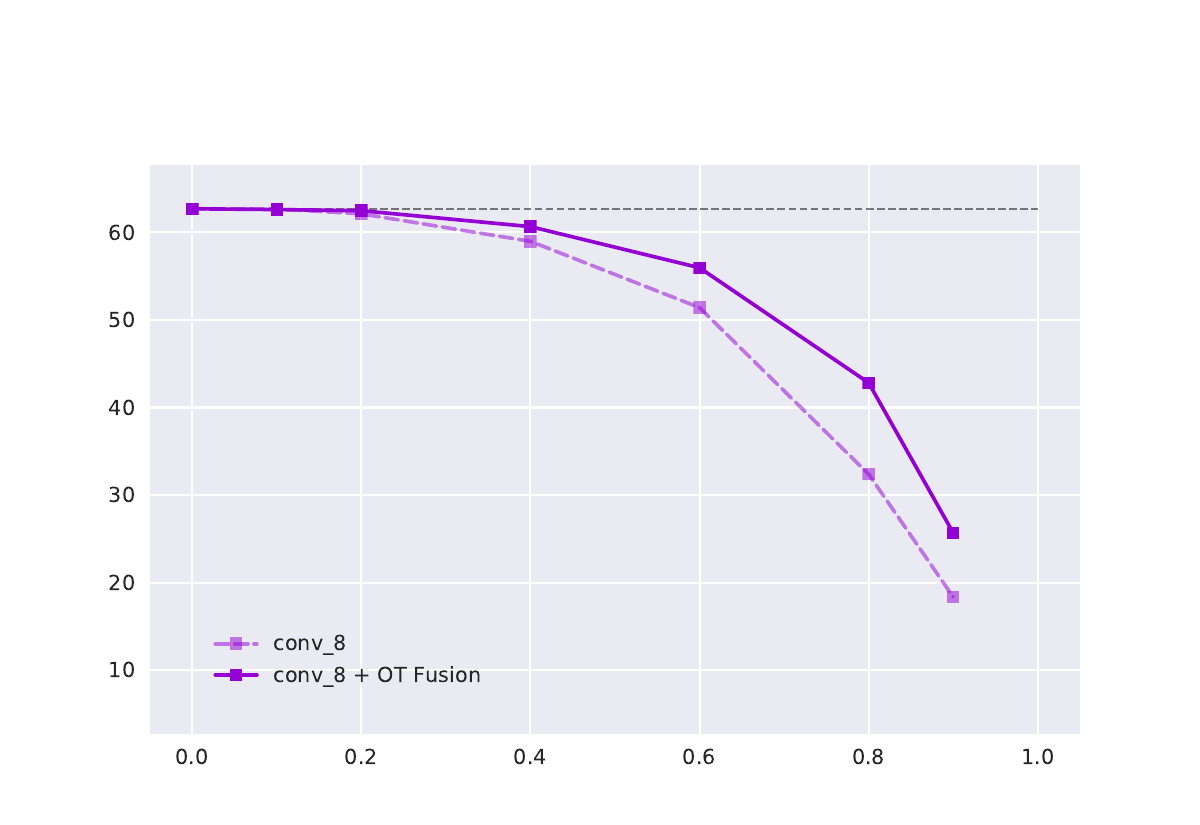}
	}\vspace{-3mm}
	\subfigure[all]{\includegraphics[width=0.4\textwidth]{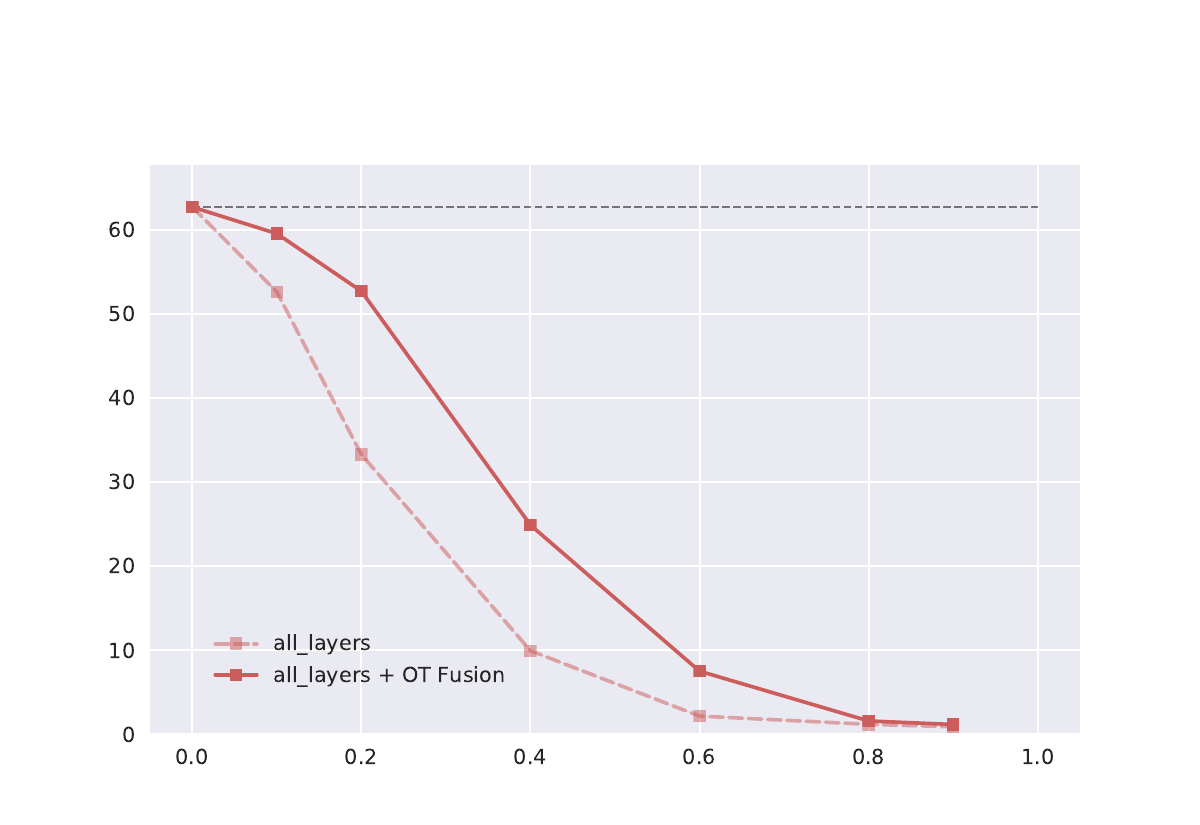}
	}
	\caption{Post-processing for structured pruning \textbf{with $\ell_1$ norm}, all figures: 
		Fusing the initial dense \textsc{VGG11} model into the pruned model helps test accuracy of the pruned model on \textsc{\textbf{CIFAR100}}.} %$m=400$, 
	\label{fig:l1_all_cifar100}	
	\vspace{-1em}
\end{figure}

\paragraph{CIFAR100.} Also in Figures~\ref{fig:l1_all_cifar100}, we show the results of a similar experiment when pruning a VGG11 model trained on CIFAR100. Here as well, OT fusion leads to a performance boost when used as for post-processing. For simplicity, we only include the results with $\ell_1$-pruner. 

\clearpage

\section{Additional discussion on the update rule in the algorithm}\label{app:free-support}

\subsection{Barycentric projection}
The original formulation by \cite{monge1781memoire} considers finding a mapping $f: \mS_A \mapsto \mS_B$, which maps points in the support of $\mu_A$ to points in the support of $\mu_B$. However, under this formulation, the optimal transport problem is not always feasible and \citet{kantorovich1942translocation} relaxed this by instead considering the optimization over the set of coupling matrices $\mT$ (i.e., doubly stochastic matrices). Hence, a simple way to obtain the optimal map $f$ from this coupling/transportation matrix $\mT$ is $f({\mS_A}^i) = \mZ_i$, where $\mZ = \mS_B \mT^{\top} \operatorname{diag}\left(\alpha^{-1}\right)$, c.f. \citet{Ferradans_2013}. Essentially, this maps each point in support of A to a weighted average of the points in support of B. This is referred to as the barycentric projection. 

For our algorithm in the weight-based alignment, we basically need the opposite map $f^\prime: \mS_B \mapsto \mS_A$ to get the weights for the model A aligned with respect to B. This can be done by simply exchanging the above supports and transposing the matrix $\mT$. Thus we have, $f^\prime(\mS_B^i) = \mZ^\prime_i$, where $\mZ^\prime = \mS_A \mT \operatorname{diag}\left(\beta^{-1}\right)$. In other words, we represent each point in the support of B with a weighted average of points in support of A. 

Lastly, the supports in this weight-based alignment are defined by the corresponding weight matrices of the layers. Thus, implying the update in Eq.~\eqref{eq:ot-upd}. Note that, when the underlying ground cost used is the squared Euclidean distance, this barycentric mapping is known to be optimal \cite{ambrosio}. 

\subsection{Free-support barycenters}
Next, we discuss the setup and a part of the derivation of  free-support barycenters proposed in \cite{cuturi_doucet14}. 

\paragraph{Problem formulation.} Assume that the ground metric $D_\GS$ is the Euclidean distance and $p=2$. Consider the supports $\mS_A$ and $\mS_B$ as family of $\na$ and $\nb$ points respectively in   $\mathbb{R}^{d}$. Therefore represent them by a matrix in  $\mathbb{R}^{d \times \na }$ and $\mathbb{R}^{d \times \nb}$ respectively. If we use the notation, 
\(\mathbf{s_A} \stackrel{\mathrm{def}}{=} \operatorname{diag}\left(\mS_A^{\top} \mS_A\right)\) and \(\mathbf{s_B} \stackrel{\mathrm{def}}{=} \operatorname{diag}\left(\mS_B^{\top} \mS_B\right),\) then we can write the pairwise squared-Euclidean distances as follows: 

\begin{equation}
\mC_{A B}=\mathbf{s_A} \1_{\nb}^{\top}+\1_{\na} \mathbf{s_B}^{\top}-2 \mS_A^{\top} \mS_B \in \mathbb{R}^{\na \times \nb}.
\end{equation}

Now the optimal transport objective mentioned in Section \ref{sec:backgroud_OT} can be written in a more compact form in the equation \eqref{eq:ot-neat} by using the matrix inner product notation. As a recall, $\langle \mU, \mV \rangle  = \operatorname{tr}(\mU^{\top}\mV)$, so we have:

\begin{equation}\label{eq:ot-neat}
\begin{aligned}\left\langle \mT, \mC_{A B}\right\rangle &=\left\langle \mT, \mathbf{s_A} \1_{d}^{\top}+\1_{d}^{\top} \mathbf{s_B}-2 \mS_A^{\top} \mS_B\right\rangle \\ &=\operatorname{tr}(\mT^{\top} \mathbf{s_A} \1_{d}^{\top})+\operatorname{tr}(\mT^{\top} \1_{d}^{\top} \mathbf{s_B})-2\left\langle \mT, \mS_A^{\top} \mS_B\right\rangle \\ &=\mathbf{s_A}^{\top} \alpha+\mathbf{s_B}^{\top} \beta -2\left\langle T, \mS_A^{\top} \mS_B\right\rangle. \end{aligned}
\end{equation}

Suppose we are given the transport map $\mT^{\star}$ which is optimal for the above Eq~\eqref{eq:ot-neat}, but we do not know the support $\mS_B$. One way to compute it is by minimizing the above with respect to $\mS_B$. Hence, let's discard the constant terms in \(\mathbf{s_A}\) and \(\alpha\). Recall $\mu_A=(\alpha, \mS_A)$ and $\mu_B=(\beta, \mS_B)$, so we have that minimizing \( \text{OT}\left(\mu_A, \mu_B, \mC_{A B}\right)\) with respect to the locations \(\mS_B\) is same as solving

\begin{equation}\label{eq:ot-free}
\min _{\mS_B \in \mathbb{R}^{d \times \nb}} \mathbf{s_B}^{\top} \beta - \left\langle \mT^{\star}, \mS_A^{\top} \mS_B\right\rangle
\end{equation}

\paragraph{Quadratic Approximation.} \citet{cuturi_doucet14} show that the above minimization problem in Eq.~\eqref{eq:ot-free} is non-convex in the locations $\mS_B$, the proof of which can be found in their work. Therefore,they resort to a local quadratic approximation mentioned in equation \eqref{eq:free-ot-min}, minimizing which yields the Newton update in equation \eqref{eq:free-ot-newton}.

\begin{equation}\label{eq:free-ot-min}
\begin{array}{r}{\mathbf{s_B}^{\top} \beta-\left\langle \mT^{\star}, \mS_A^{\top} \mS_B\right\rangle=\left\|\mS_B \operatorname{diag}\left(\beta^{1 / 2}\right)-\mS_A \mT^{\star} \operatorname{diag}\left(\beta^{-1 / 2}\right)\right\|^{2}} \\ {-\left\|\mS_A \mT^{\star} \operatorname{diag}\left(\beta^{-1 / 2}\right)\right\|^{2}}\end{array}
\end{equation}

\begin{equation}\label{eq:free-ot-newton}
\mS_B \leftarrow \mS_A \mT^{\star} \operatorname{diag}\left(\beta^{-1}\right)
\end{equation}

This can be interpreted as follows as follows: the matrix \(\mT^{\star} \operatorname{diag}\left(\beta^{-1}\right)\) has \(n_B\) columns in the simplex
\(\Sigma_{\na} \) and thus post-multiplying $\mS_A$ with this matrix means that we are performing convex combinations of the points in \(\mS_A\) with weights defined
by the optimal transport map \(\mT^{\star}\). 

\paragraph{Relation to Model Fusion.} Let's come back to our algorithm where we use the weight-based alignment. Here, the locations (or the supports) are defined by the corresponding weight matrices of the layers. This bears resemblance to the update in Eq. \eqref{eq:ot-upd} in Section \ref{sec:methodo}, where the equivalent of the unknown $\mS_B$ are the weights of A aligned with respect to B.

\section{Connection to the mean-field limit}
\begin{table}[h] \centering\ra{1.3}
	\centering
	\resizebox{0.8\textwidth}{!}{
		\setlength{\tabcolsep}{3pt}
		\large 
		\begin{tabular}{@{}c|cc|ccc|cc@{}}
			\toprule
			\multirow{2}{*}{\textsc{Hidden layers}} 	 &  \multicolumn{5}{c}{\% \textsc{Test accuracy}} &  \multicolumn{2}{c}{\textit{\% \textsc{Relative Gap}
			}} \\
			\cmidrule{2-8}
			& \multicolumn{1}{c}{\textsc{Model A}} & \multicolumn{1}{c}{\textsc{Model B }} & \multicolumn{1}{c}{\textsc{Prediction avg.}} & \multicolumn{1}{c}{\textsc{Vanilla avg.}} & \multicolumn{1}{c}{\textsc{OT avg. }} &  \multicolumn{1}{c}{\textsc{Vanilla Avg.}} & \multicolumn{1}{c}{\textsc{OT avg. }} \\
			\midrule
			
			$\lbrack40, 20, 10\rbrack$ & 96.69 & 96.91 & 97.50 & 34.82  & 82.91 & 64.07 & \textbf{14.44 }\\
			$\lbrack200, 100, 50\rbrack$ & 98.00 & 97.97 & 98.16 & 47.30  & 93.93 & 51.73 & \textbf{4.16 }\\
			$\lbrack400, 200, 100\rbrack$ & 98.13 & 98.09 & 98.21 & 73.51  & 96.70 & 25.09 & \textbf{1.45 }\\
			$\lbrack1000, 500, 250\rbrack$ & 98.08 & 98.21 & 98.20 & 78.21  & 97.35 & 20.36 & \textbf{0.87 }\\
			$\lbrack2000, 1000, 500\rbrack$ & 98.26 & 98.16 & 98.21 & 85.71  & 97.41 & 12.77 & \textbf{0.86 }\\
			\bottomrule
	\end{tabular}}
	\tiny \caption{Relative gap of OT avg. wrt the best individual model as the width of the hidden layers increases. }
	\label{tab:mean-field}
\end{table}

\textit{Effect of layer width: } Table \ref{tab:mean-field} illustrates that as the width of networks increases, the gap in performance of one-shot OT averaging compared to the best individual network decreases. This also suggests a very interesting  potential connection with the mean-field limit for neural networks \citep{mei2019meanfield}. 

Here, the authors show that as the size of the hidden layer goes to infinity, doing gradient descent on the network weights is equivalent to considering a probability density over the neurons in a layer, which evolves with Wasserstein gradient flow.  Then given two neural networks evolving under the dynamics of Wasserstein gradient flow, fusing them into one network by Wasserstein barycenter would be a natural consideration. 

We
empirically show that this limit is roughly achieved in practice when the width $\approx$1000. In particular, Table \ref{tab:mean-field} illustrates that as the width of networks increases, the gap in performance of one-shot averaging (with respect to the best individual network) on MNIST decreases.  
As a consequence, this further implies that in the setting of finite hidden layer sizes, it would help to choose the $\alpha$ and $\beta$ in a better way than just uniform. We aim to study this aspect in a future work.

\clearpage

\section{Teacher-Student Fusion}\label{app:fusion_size}

We present the results for a setting where we have trained teacher and student networks, and we would like to combine the knowledge of large teacher network into the smaller student network. This is essentially reverse of the client-server setting described in Section~\ref{sec:client_server}. We consider that all the hidden layers of the teacher model \ma $,$ are a constant $\rho \times$ wider than all the hidden layers of student model \mb. We experiment with two instances of this (a) on \mnist $\,$+ \mlp, with $\rho \in \{2, 10\}$ and (b) on \cifar$\,$ + \vgg, with $\rho \in \{2, 8\}$, and the results are presented in the Table \ref{tab:model_compression_app}. This leads to the mentioned model sizes (\# of parameters) for both these models. Our OT average uses activation-based alignment for both the settings described in the Table~\ref{tab:model_compression_app}. \\

Vanilla averaging can not be used due to different sizes of the networks. So, as a first baseline, we consider the performance of finetuning the model \mb.  We observe that across all the settings, OT avg. + finetuning outperforms this baseline as well as the original model \mb, resulting in the desired knowledge transfer from the teacher network. 

\begin{table}[h!] \centering\ra{1.3}
	\centering
	
	\resizebox{0.65\textwidth}{!}{
		\begin{tabular}{@{}cc|c|cc|cc@{}}
			\toprule
			
			\multirow{1}{*}{\textsc{Dataset + }} & \multirow{1}{*}{\textsc{$\#$ params}}  &\multirow{2}{*}{\textit{\ma}} & \multirow{2}{*}{\textsc{\mb}} & \multirow{1}{*}{\textsc{OT}} &  \multicolumn{2}{c}{\textsc{Finetuning}} \\
			
			\multirow{1}{*}{\textsc{Model}} & \multirow{1}{*}{ (\ma, \mb) }  &  &   &\multirow{1}{*}{\textsc{avg.}} &  \multirow{1}{*}{\textsc{\mb}}& \multirow{1}{*}{\textsc{OT avg.}}\\
			
			\midrule
			\multirow{1}{*}{\mnist $\,$+} & \multirow{1}{*}{(414 K, 182 K)} &	{98.11} &	97.84 & 95.67  & 98.06 & \textbf{98.22}\\
			\multirow{1}{*}{ \mlp} &	\multirow{1}{*}{(414 K, 32 K)} &	98.11 & {97.08} & 96.50	  & 97.31& \textbf{97.42} \\
			\midrule
			\multirow{1}{*}{\cifar $\,$+} & \multirow{1}{*}{(118 M, 32 M)} &	91.22 &	90.66 & 86.73  & 90.67 & \textbf{90.89}\\
			\multirow{1}{*}{ \vgg} &	\multirow{1}{*}{(118 M, 3 M )} &	91.22 & {89.38} & 88.40	  & 89.64 & \textbf{89.85}\\

			\bottomrule
	\end{tabular}}
	\caption{\textit{Compressing \ma $\,$ to smaller models. }The finetuning results of each method are at their best scores across different finetuning hyperparameters (like, learning rate schedules). OT avg. has the same number of parameters as \mb.}\label{tab:model_compression_app}
	
\end{table}

\begin{table}[h!]\centering\ra{1.3}
	\caption{Hyper-parameters corresponding to the results for model compression presented in Tables~\ref{tab:model_compression_app}. LR denotes the learning rate. For the \mnist $\,$ + \mlp  $\,$ setting a constant learning rate was employed, similar to its training procedure. While for the \cifar $\,+$ \vgg $\,$  setting, the learning rate schedule was also tuned, keeping in accordance with its training procedure as well.  The \# params column indicates the size of the resultant compressed (smaller) model. }\label{tab:hyper_params}
	\vspace{1em}
	
	\resizebox{\linewidth}{!}{		\begin{tabular}[h]{lcccccccccc}
			\toprule
			\multirow{2}{*}{\textit{Dataset + Model}} &  \multirow{2}{*}{$\#$ Student params} & \multicolumn{3}{c}{Finetune} & \multicolumn{3}{c}{LR schedule hyper-params} \\
			\cmidrule(l{3pt}r{3pt}){3-5}\cmidrule{6-8}
			& & type & LR & Epochs & Enabled & LR Decay Factor & LR Decay Epochs \\
			\midrule
			
			\multirow{4}{*}{\mnist $\,$+ \mlp} &  \multirow{1}{*}{182 K} &  \multirow{2}{*}{\mb} &	0.01 &  \multirow{4}{*}{60} &	\multirow{4}{*}{\xmark} & \multirow{4}{*}{---} & \multirow{4}{*}{---}\\
			& 	 \multirow{1}{*}{32 K} & & 0.002 & &	 & & \\
			\cmidrule{2-4}
			& \multirow{1}{*}{182 K} &  \multirow{2}{*}{OT Avg.} &	0.01 &	& & \\
			&	 \multirow{1}{*}{32 K} & & 0.001&  &	 & & \\
			\midrule
			\multirow{4}{*}{\cifar $\,$+ \vgg} &  \multirow{1}{*}{32 M} &  \multirow{2}{*}{\mb} &	\multirow{2}{*}{0.01} & \multirow{4}{*}{120}  & \multirow{4}{*}{\cmark}	 & \multirow{1}{*}{2.0} & \multirow{1}{*}{$[20, 40, 60, 80, 100]$} \\
			& 	 \multirow{1}{*}{3 M} & &  & &	 & \multirow{1}{*}{1.5} & \multirow{1}{*}{$[10, 30, 50, 70, 90, 110]$} \\
			\cmidrule{2-4} \cmidrule{7-8}
			& \multirow{1}{*}{32 M} &  \multirow{2}{*}{OT Avg.} &	\multirow{2}{*}{0.01} & & & \multirow{2}{*}{2.0} & \multirow{2}{*}{$[20, 40, 60, 80, 100]$} \\
			&	 \multirow{1}{*}{3 M} & & & &	&  &  \\
			
			\bottomrule
	\end{tabular}}
	
\end{table}

We show that even if the smaller model  \mb $\,$ were to be finetuned with many different hyper-parameters, it does not outperform the performance gained by finetuning the OT average. 

For \mnist $\,$+ \mlp, we did a sweep for the following set of finetuning learning rates  $\{0.01, 0.002, 0.001, 0.00067, 0.0005\}$. These correspond to scaling the training learning rate by a factor $\{1, 5, 10, 15, 20\}$ respectively. Both OT average and the model \mb $\,$ were finetuned for 60 epochs using these choices as a constant learning rate. 

Next, for \cifar$\,$ + \vgg, we additionally sweep for the hyper-parameters associated with the learning rate (LR) schedule during finetuning. Since unlike the \mnist $\,$ case, the original models here used a decaying learning rate schedule. The sweep was carried out for the following set of values: LR decay factor $=\{1.2, 1.5, 2\}$, LR decay epochs $=\{[20, 40, 60, 80, 100], [10, 30, 50, 70, 90, 110], [30, 60, 90]\}$. The learning rate (LR) itself was picked from $\{0.01, 0.005, 0.0033, 0.0025\}$ corresponding to scaling the training learning rate by a factor of $\{5, 10, 15, 20\}$ respectively, as done in Section \ref{sec:cifar_vgg_app} before. Table~\ref{tab:hyper_params} thus indicates the hyper-parameter choice corresponding to the best results presented in Table~\ref{tab:model_compression_app}.

From sweeping on the width-ratio $\rho=8$ for \cifar + \vgg $\,$ (i.e., when the smaller model has 3M params), we found that the learning rates $\{0.01, 0.005\}$ produced the best results for both the finetuning type/methods (namely, OT average and model \mb). Thus, while sweeping on the width-ratio $\rho=2$ for \cifar + \vgg $\,$ (i.e., when the smaller model has 32M params), we reduce the hyper-parameter space by restricting the learning rate from this set $\{0.01, 0.005\}$, while still sweeping the other sets of hyper-param values, in order to save on resource costs.

Besides these best results for each method, we find that even under the same hyper-parameter configuration, finetuning OT average leads to a better performance than finetuning the smaller model \mb $\,$ across multiple runs, for a majority of the hyper-parameter settings. Overall, we conclude that the OT average and then finetuning successfully allows us to transfer the performance from a bigger model into a smaller one.

\section{Results for distillation}\label{app:distill_more}

	In this section, we present the results in relation to distilling the knowledge of a bigger model \ma $\,$ into a smaller model \mb$\,$. Here, we consider that one already has a trained version of both the models and we are interested in boosting the performance of the smaller model. 
	
	We discuss two ways of approaching this problem. One is to consider the OT average of the individual models and then finetune. The other option is to use distillation \cite{hinton2015distilling}, where we augment the loss during finetuning with a term that essentially encourages the student's (smaller model) logit distribution (smoothed) to be close that of the teacher's (bigger model) logit distribution. The smoothing is done by raising the temperature $(T)$ in the final softmax. This loss term from distillation gets is weighted by a factor of $\gamma$ and a factor of $1-\gamma$ is given to the usual finetuning loss. 
	
	The main drawback of distillation is that it requires searching for the optimal values of these hyper-parameters: temperature $T$ and loss-weighting factor $\gamma$. As evident from the Tables~\ref{tab:distillation_hyper_2} and \ref{tab:distillation_hyper_10}, even for \mnist + \mlp, depending on the size of the smaller model, the optimal hyper-parameter values can be quite different. This can be prohibitive when dealing with larger models or datasets. 
	
	Nevertheless, we compare the performance of both these approaches in Tables~\ref{tab:distill_mnist_main_app_rep} and \ref{tab:distill_mnist_app_other}, for the setting of \mnist + \mlp$\,$ with width-ratio $\rho=2, 10$  (the ratio of hidden layers sizes of the larger to the smaller model) respectively. For distillation, we consider three possible initializations for the smaller (student) model: random, model \mb, and OT average of models \ma, \mb. 
	
	\begin{table}[h] \centering\ra{1.2}
		\vspace{0.4mm}
		\centering
		\resizebox{0.7\textwidth}{!}{
			\begin{tabular}{@{}c|ccc|cc|ccc@{}}
				\toprule
				\multirow{2}{*}{\textsc{\ma}} & \multirow{2}{*}{\textsc{\mb}} &   \multirow{1}{*}{\textsc{Prediction }} & \multirow{1}{*}{\textsc{OT}} &  \multicolumn{2}{c|}{\textsc{Finetuning}} &\multicolumn{3}{c}{\textsc{Distillation}} \\
				
				& &  \multirow{1}{*}{avg.} &  \multirow{1}{*}{avg.} & \multirow{1}{*}{\textsc{\mb}} & \multirow{1}{*}{\textsc{OT avg.}}  &\multirow{1}{*}{\textsc{Random}}  &  \multirow{1}{*}{\textsc{\mb}} & \multirow{1}{*}{\textsc{OT avg.}}\\
				
				\midrule
				
				\multirow{1}{*}{98.11} &	\multirow{1}{*}{97.84}	& 98.10 & \multirow{1}{*}{95.49}  &  	\multirow{1}{*}{98.04} & 	\multirow{1}{*}{98.19} & 98.18 & 98.22  & \textbf{98.30}\\
				\cmidrule{1-6}
				\multicolumn{6}{c|}{\footnotesize Mean across distillation temperatures}& 98.13  & 98.17 & \textbf{98.26}\\
				
				\bottomrule
			\end{tabular}
		}
		\caption{\textit{Compressing the bigger teacher model \ma $\,$  to half its size ($\rho=2$).  }
			The distillation scores for each student network initialization are taken for its best hyperparameter values. Both finetuning and distillation were run for 60 epochs using SGD with the same hyperparameters. Each entry has been averaged across 4 seeds. 
		}
		\label{tab:distill_mnist_main_app_rep}
	\end{table}

	\begin{table}[h] \centering\ra{1.2}
	
	\centering
	\resizebox{0.7\textwidth}{!}{
		\begin{tabular}{@{}c|ccc|cc|ccc@{}}
			\toprule
			\multirow{2}{*}{\textsc{\ma}} & \multirow{2}{*}{\textsc{\mb}} &  \multirow{1}{*}{\textsc{Prediction }}  & \multirow{1}{*}{\textsc{OT}} &  \multicolumn{2}{c|}{\textsc{Finetuning}} &\multicolumn{3}{c}{\textsc{Distillation}} \\
			
			& &  \multirow{1}{*}{avg. }  & \multirow{1}{*}{\textsc{avg.}} & \multirow{1}{*}{\textsc{\mb}}  & \multirow{1}{*}{\textsc{OT avg.}} &\multirow{1}{*}{\textsc{Random}} &  \multirow{1}{*}{\textsc{\mb}}  & \multirow{1}{*}{\textsc{OT avg.}} \\
			
			\midrule
			
			\multirow{1}{*}{98.11} &	\multirow{1}{*}{97.08} &	\multirow{1}{*}{98.13} & \multirow{1}{*}{96.50}  &   	\multirow{1}{*}{97.19} &	\multirow{1}{*}{97.35}  & 97.39  & 97.67  & \textbf{97.68}\\
			\cmidrule{1-6}
			\multicolumn{6}{c|}{\footnotesize Mean across distillation temperatures}& 97.21 & 97.55 & \textbf{97.59}  \\
			
			\bottomrule
		\end{tabular}
	}
	\caption{\textit{Compressing the bigger teacher model \ma $\,$  to one-tenth of its size ($\rho=10$). }The student model for distillation is initialized in 3 possible ways: random, OT avg., and model \mb. 
		In the first row, the distillation scores are taken at its best hyperparameter values. Both finetuning and distillation were run for 60 epochs. Each entry in the table has been averaged across four seeds. 
	}
	\label{tab:distill_mnist_app_other}
\end{table}

		The choice of distillation hyper-parameters tried was: temperature $T=\{20, 10, 8, 4, 1\}$ and loss-weighting factor $\gamma=\{0.05, 0.1, 0.5, 0.7, 0.95, 0.99\}$. Tables~\ref{tab:distill_mnist_main_app_rep} and \ref{tab:distill_mnist_app_other} report the best scores across these hyper-parameter choices. Also, each each of the reported scores in the tables have been averaged across 4 seeds. The optimization parameters are same for both finetuning and distillation to ensure fair comparison. Namely, the learning rate $=0.01$ and momentum $=0.5$, and both were optimized with SGD for 60 epochs. 
		
		 In terms of the results, we observe that initializing with OT average does the best in comparison to random and model \mb-based initializations. Further, we see that when the results for distillation with the other initializations are averaged across the distillation temperatures, the gain in performance pales in comparison to simply OT average and finetuning (c.f. Table~\ref{tab:distill_mnist_main_app_rep}).

		\begin{table}[h] \centering\ra{1.2}
			\centering
			\resizebox{0.5\textwidth}{!}{
				\begin{tabular}{@{}cccc@{}}
					\toprule
					\multirow{1}{*}{\textsc{Temperature}} & \multicolumn{3}{c}{\textsc{Distillation initializations}} \\
					\multirow{1}{*}{\textsc{$T$}} & \multirow{1}{*}{\textsc{Random ($\gamma$)}} & \multirow{1}{*}{\textsc{\mb$\,$ ($\gamma$)}} &  \multirow{1}{*}{\textsc{OT avg. ($\gamma$)}} \\					
					\midrule
					20 &  98.13 (0.05) & 98.20 (0.10)  & \textbf{98.26} (0.05)\\
					10 &  98.15 (0.05) & 98.19 (0.05)  & \textbf{98.28} (0.05)\\
					8  & 98.18 (0.05) &  98.22 (0.05) & \textbf{98.28} (0.05)\\
					4  &  98.11 (0.10)  & 98.21 (0.10)  & \cellcolor{orange!40}{\textbf{98.30} (0.05)} \\
					1  &  98.06 (0.05) & 98.04 (0.05)  & \textbf{98.17} (0.05)\\				
					\midrule
					\textit{Mean} & 98.13  & 98.17 & \textbf{98.26}\\
					\bottomrule
				\end{tabular}
			}
			\caption{\textit{Distillation results for the setting of $\rho=2$:}  Best results for various distillation initializations are shown for all the tried temperatures ($T$) values. The corresponding choice of the loss-weighing factor ($\gamma$) for these best scores is indicated next to them in brackets. Each of the scores has been averaged over four seeds. The cell in orange indicates the top result across all hyper-parameter settings and methods.
			}
			\label{tab:distillation_hyper_2}
		\end{table}
	
		\begin{table}[h] \centering\ra{1.2}
			\centering
			\resizebox{0.5\textwidth}{!}{
				\begin{tabular}{@{}cccc@{}}
					\toprule
					\multirow{1}{*}{\textsc{Temperature}} & \multicolumn{3}{c}{\textsc{Distillation initializations}} \\
					\multirow{1}{*}{\textsc{$T$}} & \multirow{1}{*}{\textsc{Random ($\gamma$)}} & \multirow{1}{*}{\textsc{\mb$\,$ ($\gamma$)}} &  \multirow{1}{*}{\textsc{OT avg. ($\gamma$)}} \\					
					\midrule
					20 &  97.25 (0.50) & 97.61 (0.70)  & \cellcolor{orange!40}{\textbf{97.68} (0.70)}\\
					10 &  97.32 (0.70) & \textbf{97.67 }(0.70)  & 97.65 (0.70)\\
					8  & 97.38 (0.50) &  \textbf{97.67} (0.70) & 97.65 (0.70)\\
					4  &  97.39 (0.70)  & 97.53 (0.70)  & \textbf{97.57} (0.99) \\
					1  &  96.73 (0.05) & 97.28 (0.95)  & \textbf{97.40} (0.95)\\				
					\midrule
					\textit{Mean} & 97.21  & 97.55 & \textbf{97.59}\\
					\bottomrule
				\end{tabular}
			}
			\caption{\textit{Distillation results for the setting of $\rho=10$:}  Best results for various distillation initializations are shown for all the tried temperatures ($T$) values. The corresponding choice of the loss-weighing factor ($\gamma$) for these best scores is indicated next to them in brackets. Each of the scores have been averaged over four seeds. The cell in orange indicates the top result across all hyper-parameter settings and methods.
			}
			\label{tab:distillation_hyper_10}
		\end{table}

			 Lastly, in Tables~\ref{tab:distillation_hyper_2} and \ref{tab:distillation_hyper_10}, we show the detailed results for each of the distillation initializations for each of the temperature values tried. These correspond respectively to the summarized results presented in Tables~\ref{tab:distill_mnist_main_app_rep} and \ref{tab:distill_mnist_app_other}. We observe that distillation from OT average performs the best for most of the hyper-parameter settings, as well as in terms of the overall top performance for both the width-ratio $\rho$ settings. 

			To conclude, when distillation is out of the question due to resource constraints, OT average + finetuning can go a long way. While in cases where distillation is permissible, it can be advantageous to initialize with OT average when fusing a big model into a smaller one.

\end{document}